\documentclass[preprint,3p]{elsarticle}

\usepackage{amssymb}
\usepackage{hyperref}
\usepackage{xcolor}
\usepackage{natbib}
\usepackage{algorithm}
\usepackage{algpseudocode}
\usepackage{amsmath}

\usepackage{multirow}
\usepackage{tabularx,siunitx}
\usepackage{threeparttable}
\usepackage{subcaption}
\usepackage{float}
\usepackage{caption}
\usepackage{amsthm}

\usepackage{enumitem}

\usepackage{lineno}
\journal{Expert Systems with Applications}

\makeatletter
\renewcommand{\fnum@figure}{Fig. \thefigure}
\makeatother

\begin{document}

\begin{frontmatter}



\title{Enhancing Graph U-Nets\\for Mesh-Agnostic Spatio-Temporal Flow Prediction}


\author[inst1]{Sunwoong Yang}
\ead{sunwoongy@kaist.ac.kr}
\affiliation[inst1]{organization={Cho Chun Shik Graduate School of Mobility, Korea Advanced Institute of Science and Technology},
            city={Daejeon},
            postcode={34051}, 
            country={Republic of Korea}}

\author[inst2]{Ricardo Vinuesa}
\ead{rvinuesa@mech.kth.se}
\affiliation[inst2]{organization={FLOW, Engineering Mechanics, KTH Royal Institute of Technology},
            city={Stockholm},
            postcode={SE-100 44}, 
            country={Sweden}}

\author[inst1,inst3]{Namwoo Kang\corref{cor1}}

\affiliation[inst3]{organization={Narnia Labs},
            city={Daejeon},
            postcode={34051}, 
            country={Republic of Korea}}

\cortext[cor1]{Corresponding author. \texttt{nwkang@kaist.ac.kr}}


\begin{abstract}
This study aims to overcome the limitations of conventional deep-learning approaches based on convolutional neural networks in complex geometries and unstructured meshes by exploring the potential of Graph U-Nets for unsteady flow-field prediction. We present a comprehensive investigation of Graph U-Nets, originally developed for classification tasks, now tailored for mesh-agnostic spatio-temporal forecasting of fluid dynamics. Our focus is on enhancing their performance through systematic hyperparameter tuning and architectural modifications. We propose novel approaches to improve mesh-agnostic spatio-temporal prediction of transient flow fields using Graph U-Nets, enabling accurate prediction on diverse mesh configurations. Key enhancements to the Graph U-Net architecture, including the Gaussian-mixture-model convolutional operator and noise injection approaches, provide increased flexibility in modeling node dynamics: the former reduces prediction error by 95\% compared to conventional convolutional operators, while the latter improves long-term prediction robustness, resulting in an error reduction of 86\%. We demonstrate the effectiveness of these enhancements in both transductive and inductive learning settings, showcasing the adaptability of Graph U-Nets to various flow conditions and mesh structures. This work contributes to the field of reduced-order modeling for computational fluid dynamics by establishing Graph U-Nets as a viable and flexible alternative to convolutional neural networks, capable of accurately and efficiently predicting complex fluid flow phenomena across diverse scenarios.

\end{abstract}




\end{frontmatter}


\section{Introduction}
\label{sec:intro}

Instantaneous transient flow prediction is essential for the realization of digital twins, which serve as the cornerstone for real-time forecasting and decision making in a variety of engineering fields. More specifically, the integration of accurate unsteady flow prediction models is critical to the fidelity and effectiveness of digital twins, enabling them to mirror their physical counterparts in real-time. However, conventional high-fidelity computational fluid dynamics (CFD) analysis, which is one of the most useful approaches for understanding fluid-flow behavior, has the limitation of requiring significant computational resources for practical applications \cite{hu2022aerodynamic, qu2022unsteady}.

In this context, artificial intelligence (AI) approaches have emerged as promising alternatives, playing a critical role in predicting real-time flow fields within digital-twin scenarios \cite{yang2024data,vinuesa2024perspectives,eivazi2022physics,barwey2023multiscale,barnett2023neural,jeon2024residual}. Leveraging the power of AI can significantly reduce the computational burden while improving the accuracy of transient flow simulations and automating the entire analysis process. In addition, AI-driven approaches offer the potential for real-time prediction, which is essential for digital twins that can adapt to dynamic flow conditions. These advances pave the way for more sophisticated and responsive digital twins, ensuring that their digital counterparts can operate in real-time with unprecedented accuracy and efficiency, revolutionizing predictive maintenance and operational effectiveness across CFD-related industries.

Particularly, convolutional neural networks (CNNs) have led to significant advances in the prediction of both steady and unsteady flow fields \cite{kang2022physics, maulik2021reduced, hasegawa2020machine, guastoni2021convolutional, hu2022mesh, eivazi2022towards}. While CNNs have introduced more efficient and physically meaningful flow-prediction algorithms due to their spatial locality property, they are inherently designed for structured grids and therefore face significant limitations when confronted with unstructured meshes. This limitation is exacerbated as unstructured meshes become more prevalent in real-world industrial sites due to their flexibility in handling complex geometries and dynamic mesh scenarios. Consequently, the reliance on CNNs, which are not applicable to unstructured meshes, highlights a significant limitation of their potential within digital twins for real-world industrial scenarios. This gap highlights the need for advanced deep-learning models that can handle the various unstructured grid systems for generality and versatility in practical applications.

In that regard, graph neural networks (GNNs) can be a powerful solution, effectively bridging the gap left by CNNs through their inherent mesh scalability \cite{barwey2023multiscale,lino2022multi,xiang2024solving,ogoke2021graph,wang2024identification,xiang2024solving,gao2023dynamic,pichi2024graph, chen2024adaptive}. They can handle different structures of mesh datasets by learning directly from the graph consisting of nodes and edges without the spatial topology restrictions (structured grid) inherent in CNNs \cite{roznowicz2024large}. Therefore, GNNs are perfectly positioned to address the challenges associated with both structured and unstructured meshes commonly encountered in CFD simulations. The mesh flexibility of GNNs enables them to seamlessly adapt to various mesh configurations during training and inference, making them highly compatible with diverse CFD datasets. By directly operating on graph-structured data, GNNs can effectively capture the intricate spatial relationships and dependencies present in the mesh, allowing them to model the complex dynamics and evolution of unsteady flow fields with high fidelity.

Among the various approaches based on GNNs, Graph U-Nets, which originated from the U-Net architecture, have been introduced for their performance in node/graph classification tasks \cite{gao2019graph}. U-Nets have been widely adopted for their efficiency in feature extraction and localization across different resolution scales, enabling precise segmentation and reconstruction tasks \cite{anand2023fusion, ronneberger2015u, deng2022vortex, jurado2022deep, feng2021urnet, cremades2024identifying, tang2020deep, song2024two}. This ability of the U-Nets to capture multi-scale features is particularly well-suited for flow-prediction tasks where flow structures with a wide range of length scales exist. By integrating the U-Net framework with GNNs, Graph U-Nets inherit these strengths, enabling exceptional performance even in untrained mesh scenarios \cite{gao2019graph}. However, their comprehensive validation has not been conducted in regression tasks with high-dimensional spatio-temporal output space, which are the most common in practical flow problems. Therefore, this work attempts to explore the potential of this combination, U-Nets and GNNs, which would ensure the multi-scale nature of U-Nets even in the unstructured-data situation, allowing them to capture both local and global flow patterns and make them highly effective in predicting transient flow fields with complex spatio-temporal dependencies.

This study focuses on leveraging the Graph U-Net architecture to predict vortex-shedding flow over a circular cylinder, which is one of the most famous benchmark flow problems for spatio-temporal prediction \cite{patel1978karman}. We propose several novel approaches to augment the synergy between the strengths of GNNs in capturing mesh-based dependencies and the spatio-temporal dynamics of transient flows. The proposed methods introduce several key enhancements to the traditional Graph U-Net architecture, including the Gaussian mixture model (GMM) convolutional operator \cite{monti2017geometric} for increased flexibility in modeling node dynamics, and noise injection approaches \cite{pfaff2020learning} to mitigate error accumulation over long time-stepping. Throughout this study, extensive parametric studies are conducted to investigate the impact of various architectural choices and hyperparameters on the performance of the Graph U-Net model in order to optimize its performance specifically for fluid dynamics applications.

\clearpage
The main contributions of this study can be summarized as follows:

\begin{enumerate}
    \item A comprehensive exploration of Graph U-Nets in unsteady flow-field prediction, bridging the gap between structured grid-based CNN approaches and the need for flexible, mesh-agnostic real-time prediction models in CFD domain.
    \item Systematic investigation and optimization of Graph U-Net architecture for fluid dynamics applications, including: 
    \begin{enumerate}[label=\alph*)]
        \item Implementation of the Gaussian mixture model (GMM) convolutional operator, demonstrating increased flexibility and accuracy in modeling node dynamics. 
        \item Analysis of pooling strategies, revealing their impact on prediction accuracy and generalization capabilities. 
        \item Implementation of noise injection techniques to significantly enhance long-term prediction stability. 
        \item Comparative assessment of normalization methods, uncovering their crucial role in model performance across different mesh scenarios.
    \end{enumerate}
    \item In-depth investigation of both transductive and inductive learning performance, demonstrating the ability of Graph U-Nets to generalize to untrained node regions and entirely new mesh configurations.
    \item Novel insights into inductive learning for Graph U-Nets, including:
    \begin{enumerate}[label=\alph*)]
        \item Successful extension of the training scenario to encompass meshes with varying vortex shedding periods.
        \item The counter-intuitive finding that omitting pooling operations leads to superior performance in inductive settings.
        \item Discovery of the significant impact of normalization techniques on inductive learning, with performance highly dependent on whether the mesh structure remains static or changes dynamically over time.
    \end{enumerate}
    \item Demonstration of substantial improvements in prediction accuracy and long-term stability through carefully tuned Graph U-Net models, establishing their viability for practical fluid dynamics applications.
    \item Development of a comprehensive framework for adapting and optimizing Graph U-Nets, originally designed primarily for classification tasks, to perform effectively in regression tasks for spatio-temporal flow prediction. This work demonstrates the versatility of Graph U-Nets beyond their conventional classification applications, unlocking new potential for their use in complex flow dynamics predictions.
\end{enumerate}

By addressing these key aspects and introducing novel enhancements to the Graph U-Net architecture, our study presents a comprehensive and tailored approach for mesh-agnostic spatio-temporal forecasting of unsteady flow fields. We believe that this work not only advances the application of graph-based models in computational fluid dynamics but also provides valuable insights for researchers and practitioners seeking to leverage Graph U-Nets as an efficient, accurate, and adaptable method for flow field prediction across a variety of engineering domains.

\clearpage
\section{Methodologies}
\label{sec:method}

\subsection{Graph convolutional operators}
\label{sec:GCN}
In this section, we first review the basic aspects of GCN operator and then move on to the GMM operator, which can be considered as an enhanced GCN. 

\subsubsection{Graph convolutional network (GCN) operator}
\label{sec:GCN_oper}
It is important to note that GCNs are inspired by the foundational principles of CNNs, specifically their ability to perform feature extraction through the convolution operation \cite{kipf2016semi, wu2024hyperspectral}. While CNNs apply convolutional filters over regular, grid-like data structures (e.g., images), GCNs adapt this concept to non-Euclidean graph-structured data where the layout is irregular. Both methods aggregate local neighborhood information to capture spatial hierarchies, but GCNs extend this idea to nodes and their connections, enabling feature extraction from graphs in a manner analogous to how CNNs operate on images. The conventional GCN operation in node-wise formulation is as follows:

\begin{equation}\
\label{eq:GCN}
\mathbf{x}^{\prime}_i = \mathbf{\Theta}^{\top} \sum_{j \in \mathcal{N}(i) \cup \{ i \}} \frac{e_{j,i}}{\sqrt{\hat{d}_j \hat{d}_i}} \mathbf{x}_j,
\end{equation}
where $\mathbf{x}_i'$ represents the updated features for node $i$, $\mathcal{N}(i) \cup \{ i \}$ denotes the set of neighbors for node $i$, including $i$ itself to account for self-loops (which means $A + I$ when adjacency matrix is defined as $A$ and $I$ is the identity matrix), $e_{j,i}$ is the edge weight from node $j$ to node $i$, $\hat{d}_i$ is the degree of node $i$ adjusted for edge weights ($\hat{d}_i = 1 + \sum_{j \in \mathcal{N}(i)} e_{j,i}$), and $\mathbf{\Theta}$ is the learnable weight matrix. This formulation effectively captures the aggregation of neighbor features ($\mathbf{x}_j$) weighted by their respective edge weights ($e_{j,i}$) and normalized by the nodes' degrees ($\sqrt{\hat{d}_j \hat{d}_i}$), facilitating learning on graphs.

In the original Graph U-Net paper \cite{gao2019graph}, an improved version of the GCN layer was proposed by modifying the computation of the adjacency matrix. That is, instead of using $A + I$ as in conventional GCN, it is suggested to use $A + 2I$. This modification assigns higher weights to self-loops in the graph, effectively increasing the importance of a node's own features during the aggregation process.

Note that the inclusion of edge weights ($e_{j,i}$) in the GCN operation is optional and can be decided by the users based on their specific problem and available data. Edge weights can provide additional information about the distance or importance of connections between nodes, allowing the GCN to capture more nuanced relationships in the graph. However, in cases where edge weights are not available or not relevant, they can be set to 1 or omitted altogether. In Section \ref{sec:Conv}, GCN approaches with and without edge-weight consideration will be compared.

\subsubsection{Gaussian mixture model (GMM) operator}
\label{sec:GMM_oper}
The above-mentioned GCN approach exhibits limitations when it comes to considering edge features, which play a crucial role in capturing the complex interactions and propagation of information in certain domains, particularly in CFD. In the context of fluid-flow problems, edge features often represent the physical distances or connections between nodes, carrying valuable information about the spatial relationships and dependencies within the mesh. To address this limitation, developing graph-based models that effectively incorporate edge features and leverage complex physical correlations between nodes is crucial. In this context, the GMM convolutional operator can come into play \cite{monti2017geometric}. This operator has the ability to emulate more nuanced relationships between nodes by providing additional information about edge features, specifically the distance between neighboring nodes in this study. This adaptability and improved modeling capability makes it particularly useful for tasks that require a deep understanding of graph geometries, such as predicting transient flows where the connectivity between nodes is diverse and complex.

However, also GCN can consider the edge attributes by simply setting $e_{j,i}$ in Eq. \ref{eq:GCN} as the Euclidean distance between two nodes: $e_{j,i}=\sqrt{(x_i-x_j)^2+(y_i-y_j)^2}$. On the other hand, GMM does not use the Euclidean distance naively: it employs a mixture of Gaussians as a learnable weight function (also known as kernel) to flexibly learn how to weight edge attributes for the specific task during training. More specifically, it trains a mean vector and diagonal covariance matrix to describe the weights of the edge attributes. By incorporating this concept, the GMM convolutional operator can assign different levels of importance to edges based on the relative positions of neighboring nodes, enabling it to capture the local geometry and topology of the graph more effectively. This capability makes GMM particularly suitable for capturing spatial dependencies and complex flow patterns in graph-structured CFD data. Furthermore, GMM introduces a hyperparameter that determines the number of weighting functions (kernels) to be used in the convolutional process: the effects of this hyperparameter will be investigated in Section \ref{sec:Conv}. In summary, the node-wise formulation of GMM can be written as:

\begin{equation}\
\label{eq:GMM}
\mathbf{x}^{\prime}_i = \frac{1}{|\mathcal{N}(i)|}
\sum_{j \in \mathcal{N}(i)} \frac{1}{K} \sum_{k=1}^K      \mathbf{w}_k(\mathbf{e}_{i,j}) \odot \mathbf{\Theta}_k \mathbf{x}_j,
\end{equation}
where $K$ is the number of kernels, $\mathbf{e}_{i,j}$ is the edge attribute between nodes $i$ and $j$ (Euclidean distance in this study), and 

\begin{equation}\
\label{eq:GMMweight}
\mathbf{w}_k(\mathbf{e}) = \exp \left( -\frac{1}{2} {\left(
        \mathbf{e} - \mathbf{\mu}_k \right)}^{\top} \Sigma_k^{-1}
        \left( \mathbf{e} - \mathbf{\mu}_k \right) \right)
\end{equation}
denotes a kernel based on a trainable mean vector $\mathbf{\mu}_k$ and a diagonal covariance matrix $\mathbf{\Sigma}_k$ (there is a hyperparameter that determines the size of $\mathbf{\mu}_k$ and $\mathbf{\Sigma}_k$, which is pseudo-coordinate dimensionality \cite{monti2017geometric} --- it is set as 1 throughout this study). By learning $\mathbf{\mu}_k$ and $\mathbf{\Sigma}_k$ of edge vectors, which can be considered as additional degrees of freedom compared to conventional GCN, GMM assigns variable importance to connections based on their relative positions. This approach allows for a richer, more detailed understanding of graph topologies, enhancing the model's ability to discern and predict intricate patterns within the data.

\subsection{Graph U-Nets}
\label{sec:Gun}

The U-Net architecture, known for its effectiveness in image-segmentation tasks, features a symmetric encoder-decoder structure \cite{ronneberger2015u}. Specifically, the encoder reduces the dimensionality of the graph by pooling and thus captures the context, while the decoder restores the detail and spatial dimension by unpooling. Crucially, U-Net uses skip-connections between the encoder and decoder, allowing information to flow directly through the U-shaped structure. These connections help preserve spatial information lost during pooling in the encoder, enhancing the network's ability to accurately learn images by combining deep contextual and spatial information. Therefore, it can be concluded that its strength lies in the multi-resolution capability to combine low-level feature maps with higher-level ones, which ensures precise localization. 

\begin{figure*}[htb!]
    \centering
        \includegraphics[width=.95\textwidth]{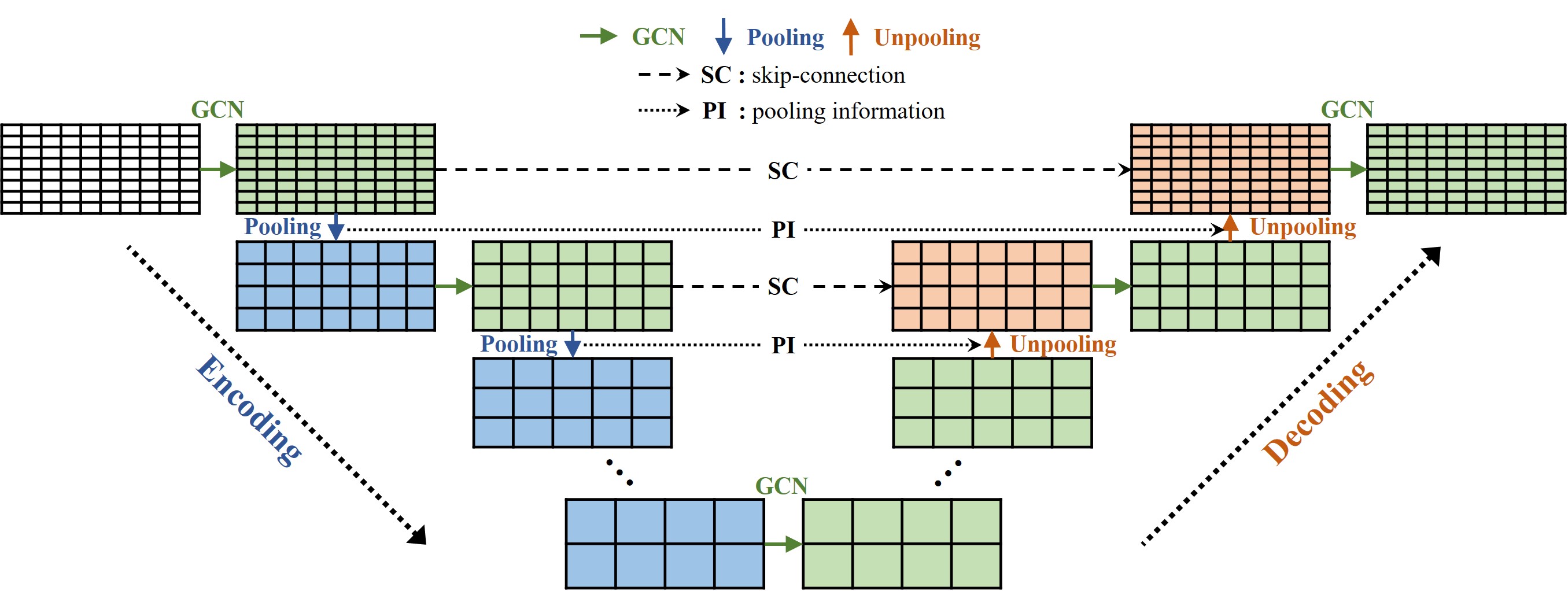}        
    \caption{Graph U-Net architecture, where SC and PI denote skip-connection and pooling information respectively.}
    \label{fig:Unet}
\end{figure*}

Graph U-Nets extend this concept to graph data, preserving the benefits of U-Nets while providing the flexibility to be applied in non-Euclidean spaces. As shown in Fig. \ref{fig:Unet}, its overall architecture is similar to the previous U-Net; however, the main difference between U-Net and Graph U-Net can be seen as instead of CNN operators in the encoder and decoder blocks of U-Net, Graph U-Nets use GCN operators \cite{gao2019graph}. Also, it should be noted that the pooling procedure in graph data is totally different from that in structured datasets. In the context of CNNs, pooling serves several crucial purposes, such as reducing the spatial dimensions of feature maps to decrease computational complexity and memory requirements \cite{zafar2022comparison}, enlarging the receptive field of the network to capture larger-scale features and contextual information \cite{gao2019graph}, and providing a form of spatial invariance to make the network more robust to small translations and distortions in the input data. However, conventional pooling operations used in CNNs, such as max pooling or average pooling, are designed for structured, grid-like data and cannot be directly applied to graph-structured data. This is because nodes in a graph do not have a fixed ordering or spatial arrangement, making it challenging to define a consistent unpooling operation \cite{gao2019graph}: in GNNs, there is no information about what the dimensions of the next graph should be during unpooling. Consequently, Graph U-Net requires specialized pooling/unpooling techniques that can effectively handle the unique characteristics of graph data while retaining the benefits of pooling operations in CNNs.

\subsubsection{Pooling for graph datasets}
\label{sec:Gun_pool}

CNNs can pool the data using max pooling or average pooling to get the reduced dimensional data, since the dataset used here has a regular grid-like structure. However, GNNs cannot naively pool the irregular unstructured data with such a simple principle because each node in the graph does not have the same number of edges, so we cannot apply the same pooling techniques as CNNs. To bridge the gap between CNNs and GNNs in terms of the pooling operation, \citet{gao2019graph} proposed the new pooling algorithm for the graph datasets: the graph pooling (gPool) technique described in Algorithm \ref{alg:gpool}. Its aim is to adaptively select a subset of nodes to form a coarsened graph based on a trainable projection vector. Specifically, the algorithm first projects the input feature matrix $X^{\ell}$ onto the projection vector $p$ (line 1 in Algorithm \ref{alg:gpool}). Next, the nodes are ranked according to their projection scores, and indices of the top $k$ nodes are selected to form the coarsened graph (line 2). Note that the hyperparameter $k$ should be predetermined, and it can be defined in two ways: a specific integer value or the ratio, such that the ratio of 0.3 means that the top 30\% of the nodes of the original graph are preserved (in this study, the ratio is used to define $k$). Then, an activation function $\sigma$ is applied to obtain a scalar projection score for each node (line 3). To mitigate the issue of isolated nodes during pooling, the adjacency matrix of the coarsened graph is updated using the second-power of the original adjacency matrix, effectively considering the two-hop neighborhood of each node\footnote{In this study, the two-hop neighborhood is not considered. The original purpose of using the second-power adjacency matrix is to compensate for the significantly coarsened graph during pooling. However, in the domain of CFD, nodes in the graph (mesh) are usually highly connected to each other, so this approach rather tends to increase the number of edges after pooling --- this result is somewhat ironic when we recall that the purpose of pooling is to coarsen the graph to capture features at a larger scale.} (line 4-{}-5) \cite{gao2019graph}. Finally, the selected node features (line 6) are gated using their corresponding projection scores to obtain the feature matrix of the coarsened graph (line 7). This process allows the gPool layer to learn a hierarchical representation of the input graph while preserving its essential structure and features, and also allows the training of the projection vector $p$ by back-propagation \cite{lecun2002efficient}. The details of the gPool can be found in Ref. \cite{gao2019graph}.

\begin{algorithm}[htb!]
\caption{Graph pooling (gPool) layer}\label{alg:gpool}
\begin{algorithmic}[1]
\renewcommand{\algorithmicrequire}{\textbf{Input:}}
\renewcommand{\algorithmicensure}{\textbf{Output:}}
\Require Graph $G^{\ell}$ with adjacency matrix $A^{\ell} \in \mathbb{R}^{N \times N}$ and feature matrix $X^{\ell} \in \mathbb{R}^{N \times F}$, pooled nodes $k$, trainable projection vector $p \in \mathbb{R}^{F}$, and activation function $\sigma$
\Ensure Coarsened graph $G^{\ell+1}$ with adjacency matrix $A^{\ell+1} \in \mathbb{R}^{k \times k}$ and feature matrix $X^{\ell+1} \in \mathbb{R}^{k \times F}$

\State $y \gets X^{\ell} \cdot p / |p|$ \Comment{Project features onto $p$}
\State $idx \gets \operatorname{rank}(y, k)$ \Comment{Get indices of top $k$ nodes}
\State $\tilde{y} \gets \sigma(y(idx))$ \Comment{Apply activation to selected nodes}
\State $A^2 \gets A^{\ell}A^{\ell}$ \Comment{Compute the second-power of adjacency matrix}
\State $A^{\ell+1} \gets A^2(idx, idx)$ \Comment{Update adjacency matrix of coarsened graph}
\State $\tilde{X}^{\ell} \gets X^{\ell}(idx, :)$ \Comment{Select features of top $k$ nodes}
\State $X^{\ell+1} \gets \tilde{X}^{\ell} \odot (\tilde{y} \cdot \mathbf{1}_F^T)$ \Comment{Apply gating to selected features}

\State \Return $A^{\ell+1}, X^{\ell+1}$
\end{algorithmic}
\end{algorithm}

\subsubsection{Unpooling for graph datasets}
\label{sec:Gun_unpool}

The Graph Unpooling (gUnpool) layer is a critical component of the Graph U-Net's decoder, enabling the restoration of the original graph structure from the coarsened graph obtained through the gPool layer during encoding. The gUnpool layer is designed to perform the inverse operation of the gPool layer, effectively upsampling the graph to its original size while restoring the essential information lost during the encoding process \cite{gao2019graph}. However, since the graph does not have a regular grid-like structure, it is impossible to recover the original dimension of the data before pooling without using any additional information. In this respect, the gUnpool layer works by using the information obtained during the gPool layer at the same encoding-decoding level (see Fig. \ref{fig:Unet}: pooling information is transferred horizontally from the encoder to the decoder): this is the core concept of the gUnpool operation.


The overall process of gUnpool layer can be found in Algorithm \ref{alg:gUnpool}: it demonstrates the scenario of upsampling the already coarsened graph $G^{\ell+1}$ back to the original graph before pooling, $G^{\ell}$ (note that in Algorithm \ref{alg:gpool}, $G^{\ell}$ was coarsened to $G^{\ell+1}$ by the gPool layer). First, the information of the original graph $G^{\ell}$ should be passed from the gPool layer: number of nodes ($N_{saved}$), selected node indices after pooling ($idx_{saved}$), and adjacency matrix ($A^{\ell}_{saved}$). Then, it initializes the output feature matrix $X^{\ell}$ based on $N_{saved}$, ensuring that upsampled graph has the same node dimension as the graph before pooing (line 1 in Algorithm \ref{alg:gUnpool}). By utilizing the saved node indices $idx$, the gUnpool layer can accurately place the nodes and their associated features back into their original positions in the restored graph (line 2). Finally, the adjacency matrix of the recovered graph is updated to that of the graph before pooling (line 3). The use of the saved variables during the gPool, $N_{saved}$, $idx_{saved}$, and $A^{\ell}_{saved}$, highlights the fact that gPool layer not only generates a coarsened graph for efficient representation learning, but also provides the necessary mapping information for the gUnpool layer to reconstruct the original graph structure accurately. This seamless integration of the pooling and unpooling layers allows the decoder to incrementally reconstruct the flow field from the reduced graph, preserving the intricate multi-scale dependencies learned during encoding. This synergistic encoder-decoder architecture is a key point of the Graph U-Net model, enabling it to learn hierarchical representations of the input graph and effectively capture both local and global patterns within the resolutions.

\begin{algorithm}[htb!]
\caption{Graph Unpooling (gUnpool) Layer}\label{alg:gUnpool}
\begin{algorithmic}[1]
\renewcommand{\algorithmicrequire}{\textbf{Input:}}
\renewcommand{\algorithmicensure}{\textbf{Output:}}
\Require Graph to be upsampled $G^{\ell+1}$ with feature matrix $X^{\ell+1} \in \mathbb{R}^{k \times F}$ \& information of the original graph $G^{\ell}$: node size $N_{saved}$, node indices $idx_{saved}$, and adjacency matrix $A^{\ell}_{saved}$ saved at the corresponding gPool layer
\Ensure Upsampled graph $G^{\ell}$ with feature matrix $X^{\ell} \in \mathbb{R}^{N \times F}$ and adjacency matrix $A^{\ell}$

\State $X^{\ell} \gets \mathbf{0}_{N_{saved} \times F}$ \Comment{Initialize output feature matrix with zeros}
\State $X^{\ell}(idx_{saved}, :) \gets X^{\ell+1}$ \Comment{Fill the new features with the previous features}
\State $A^{\ell} \gets A^{\ell}_{saved}$
\Comment{Update the adjacency matrix}

\State \Return $A^{\ell}$, $X^{\ell}$
\end{algorithmic}
\end{algorithm}

\subsection{Normalization methods}

Inductive-learning performance, which measures the ability of the model to generalize to completely new graphs not seen during training, poses significant challenges for GNNs when training different physical properties and mesh structures at the same time. The heterogeneity of node features across different graphs can hinder model generalization and stability, necessitating the use of robust and adaptive feature scaling techniques. This section demonstrates the pivotal role of normalization methods, namely layer normalization (LN) and graph normalization (GN), in enhancing the training dynamics and predictive performance of Graph U-Nets in inductive settings.

\subsubsection{Layer normalization (LN)}
\label{sec:LN}
LN, as implemented in Algorithm \ref{alg:layernorm}, effectively addresses the challenge of internal covariate shift by normalizing node features ($x$-velocity in this study) for each individual node \cite{ba2016layer}. This technique operates independently on each node, standardizing features without considering inter-node relationships or batch-level characteristics. Such localized normalization provided by LN is crucial for preserving the unique characteristics of each node, thus enhancing model performance by ensuring consistency across diverse training environments. Also, since it is applied on a single graph, it is not dependent on the mini-batch size \cite{ba2016layer}. Note that all Graph U-Nets in this study except Section \ref{sec:induct_norm} adopt this LN approach.

\begin{algorithm}[htb!]
\caption{Layer normalization (LN)}\label{alg:layernorm}
\begin{algorithmic}[1]
\renewcommand{\algorithmicrequire}{\textbf{Input:}}
\renewcommand{\algorithmicensure}{\textbf{Output:}}
\Require Node features $X \in \mathbb{R}^{N \times F}$ for $N$ nodes each with $F$ features
\Ensure Normalized features $X' \in \mathbb{R}^{N \times F}$

\State Compute the mean $\mu_i$ and variance $\sigma^2_i$ across all features:
\[
\mu_i = \frac{1}{F} \sum_{j=1}^F X_{ij}, \quad \sigma^2_i = \frac{1}{F} \sum_{j=1}^F (X_{ij} - \mu_i)^2
\] \Comment{$i$ is the index for each node}
\State Normalize the features:
\[
\hat{X}_{ij} = \frac{X_{ij} - \mu_i}{\sqrt{\sigma^2_i + \epsilon}} \quad \text{for all } i,j
\] \Comment{$j$ is the index for each feature and $\epsilon$ is the hyperparameter for numerical stability}
\State Scale and shift the normalized features:
\[
X'_{ij} = \gamma_j \hat{X}_{ij} + \beta_j \quad \text{for all } i,j
\] \Comment{$\gamma_j$ and $\beta_j$ are learnable parameters for each feature dimension}

\State \Return $X'$ \Comment{Return the layer-normalized feature matrix}

\end{algorithmic}
\end{algorithm}

\subsubsection{Graph normalization (GN)}
\label{sec:GN}
GN, detailed in Algorithm \ref{alg:graphnorm}, is specifically designed to handle the variability and complexity of graph data \cite{cai2021graphnorm}. Unlike LN, which normalizes node features across the feature dimension for each node independently, GN normalizes node features across the node dimension for each feature (see the difference between line 1 in Algorithms \ref{alg:layernorm} and \ref{alg:graphnorm}). In particular, GN computes the statistics (mean and variance) of all nodes in each graph so that the mean and variance have the same dimensionality as the number of features. Also, there is an additional learnable parameter $\alpha_j$, which automatically determines how much the information in the mean will be weighted during the shift operation in terms of feature $j$ (lines 1-{}-2 in Algorithm \ref{alg:graphnorm}) \cite{cai2021graphnorm}. \citet{cai2021graphnorm} pointed out that this learnable shift allows GNNs to overcome the expressive degradation of the conventional normalization method, InstanceNorm \cite{ulyanov2016instance}. In fact, they showed that the GN approach successfully achieved better performance than other normalization techniques in graph-based classification benchmark problems.

\begin{algorithm}[htb!]
\caption{Graph normalization (GN)}\label{alg:graphnorm}
\begin{algorithmic}[1]
\renewcommand{\algorithmicrequire}{\textbf{Input:}}
\renewcommand{\algorithmicensure}{\textbf{Output:}}
\Require Node features $X \in \mathbb{R}^{N \times F}$ for $N$ nodes each with $F$ features
\Ensure Normalized features $X' \in \mathbb{R}^{N \times F}$

\State Compute the mean $\mu_j$ and variance $\sigma^2_j$ across all nodes:
\[
\mu_j = \frac{1}{N} \sum_{i=1}^N X_{ij}, \quad \sigma^2_j = \frac{1}{N} \sum_{i=1}^N (X_{ij} - \alpha_j \cdot \mu_j)^2
\] \Comment{$j$ is the index for each feature and $\alpha_j$ is the learnable parameter for each feature dimension}

\State Normalize the features:
\[
\hat{X}_{ij} = \frac{X_{ij} - \alpha_j \cdot \mu_j}{\sqrt{\sigma^2_j + \epsilon}} \quad \text{for all } i, j
\] \Comment{$i$ is the index for each node and $\epsilon$ is the hyperparameter for numerical stability}

\State Scale and shift the normalized features:
\[
X'_{ij} = \gamma_j \hat{X}_{ij} + \beta_j \quad \text{for all } i, j
\]
\Comment{$\gamma_j$ and $\beta_j$ are learnable parameters for each feature dimension}

\State \Return $X'$ \Comment{Return the graph-normalized feature matrix}

\end{algorithmic}
\end{algorithm}

\section{Details of experiments}
\label{sec:exp}

\begin{figure*}[htb!]
    \centering
        \includegraphics[width=.7\textwidth]{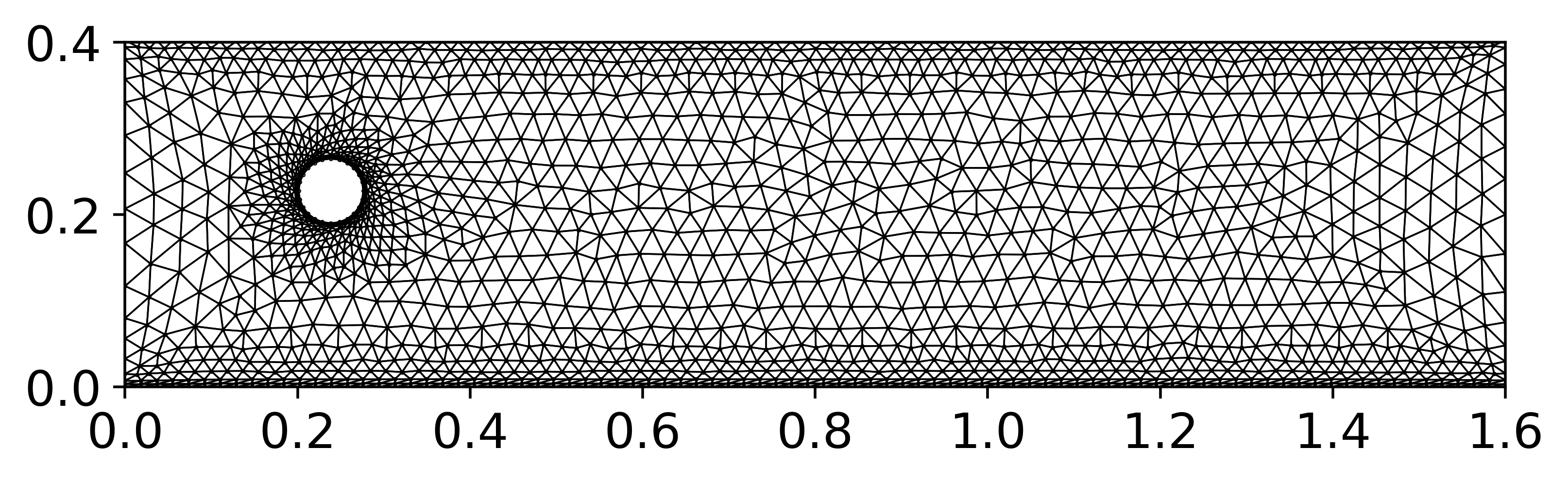}        
    \caption{Mesh used for training, containing 1,946 nodes, 11,208 edges, and 3,658 volume cells. The flow is from left to right and the details of the flow condition can be found in Table \ref{tab:induct} (named as baseline graph).}
    \label{fig:mesh_trained}
\end{figure*}
\begin{figure*}[htb!]
    \centering
        \includegraphics[width=.95\textwidth]{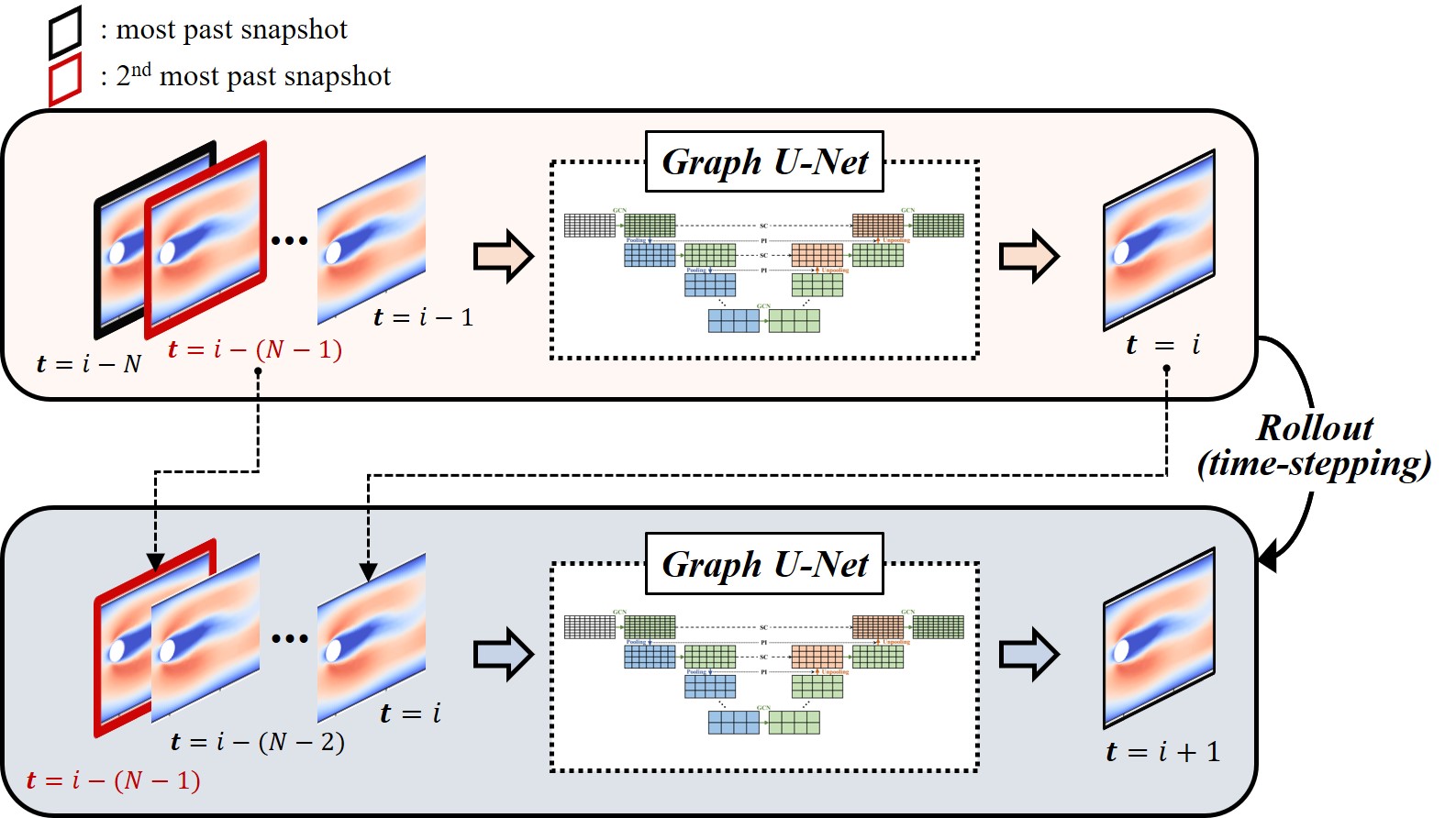}        
    \caption{Flowchart depicting how the rollout process is applied using the Graph U-Net: by using past $N$ snapshots as input, Graph U-Net predicts very next snapshot.}
    \label{fig:rollout}
\end{figure*}

In this section, we elaborate on the training process of the Graph U-Net models, designed for the prediction of spatio-temporal flow: more specifically, it is applied to the prediction of vortex shedding behind a two-dimensional (2D) circular cylinder. The training dataset is taken from the work of Google DeepMind \cite{pfaff2020learning}. The corresponding dataset consists of different mesh scenarios --- each mesh scenario has different size and location of the cylinder, and also the flow condition varies. For the CFD simulation of these meshes, the incompressible Navier-Stokes flow solver using COMSOL \cite{multiphysics2015comsol} was used to predict vortex shedding for a 6-second duration with a 0.01-second time step, yielding a total of 600 snapshots\footnote{Snapshot refers to the flow field at a specific time step, the spatial distribution of flow variables at that particular instant.} for each mesh scenario. And we first focused on a single mesh scenario consisting of 150 consecutive snapshots to form the training dataset --- by training Graph U-Nets using only these 150 past snapshots, we aim to predict the next future snapshots. The selected scenario features a mesh with 1,946 nodes, 11,208 edges, and 3,658 volume cells as in Fig. \ref{fig:mesh_trained}. A single period of vortex shedding consists of 29 snapshots, with the maximum inlet velocity\footnote{Since the $x$-velocity profile of the inlet is assumed to be parabolic, the midpoint of the profile, which is the maximum value of the inlet profile, is presented.} of 1.78 m/s and the cylinder diameter of 0.074m (the details of the flow fields are again summarized in Table \ref{tab:induct} under the name ``Baseline''). To perform the rollout (also known as time-stepping in CFD), the model's input consists of a sequence of 20 past snapshots: more specifically, the input data has 20 input channels. Note that the number of past snapshots used as the input of the model is smaller than the vortex-shedding period of the trained mesh. The output of the model is the flow field of the immediately future snapshot. Then, the loss function is computed as the (mean-squared error) MSE between this predicted future snapshot and the ground-truth future snapshot. Loss values are calculated over the entire range of training snapshots, and backpropagation is performed using the sum of all their values. After training, the rollout process for the future snapshots can be carried out --- see Fig. \ref{fig:rollout} for the details. Here, the $N$ past snapshots (recall that here $N=20$) are entered into the Graph U-Net as an input and then the very next snapshot is predicted as an output. This procedure is repeated in an auto-regressive manner, where the output of the previous rollout (snapshot at $t=i$ in Fig. \ref{fig:rollout}) is utilized as the input of the next rollout. This auto-regressive rollout allows the model to predict all subsequent snapshots during the inference phase, even though it was trained to predict only the very next snapshot during training. The nature of this rollout will be discussed again in Section \ref{sec:mot}. 

Due to constraints related to GPU memory limitations, our training exclusively focused on the $x$-velocity component of the flow field. However, it is important to note that our model is capable of handling multi-channel inputs, enabling the prediction of the complete flow field (including all velocity components and pressure) by leveraging additional computational resources.

The architecture of our Graph U-Net model incorporates a total of 4 graph convolutional network (GCN) layers as proposed by \citet{gao2019graph}. To effectively reduce the dimensionality of the dataset during encoding, gPool layer is employed subsequent to each GCN layer in the encoder process as in Fig. \ref{fig:Unet}. This pooling operation is applied with a ratio of 0.5, meaning that the number of nodes in the graph is reduced by 50\% after each pooling step. The original input graph, comprising 1,946 nodes and 11,208 edges with 20 channels (or snapshots), undergoes a series of transformations through the encoder part of the Graph U-Net (Fig. \ref{fig:Unet}). First, after the initial GCN layer coupled with LN and ELU activation function \cite{clevert2015fast} (all convolutional networks in this study are followed by normalization and activation function), the channel dimension of the graph changes, while the number of nodes remains unaltered. Since a total of 4 GCN layers are used, the number of channels of the graph data is set to decrease as 20, 15, 10, 5, and 1, so that the 20-channel input data is reduced to the single channel when leaving the encoder. And each GCN layer is followed by pooling: each time pooling is applied, the number of nodes is reduced to 50\%. By progressively reducing input dimensionality through pooling, the hierarchical mesh representation efficiently encodes multi-scale dependencies from the macroscopic down to the microscopic, highly beneficial for intricate fluid flows exhibiting cross-scale interactions. Thus, this multi-scale hierarchical pooling strategy is expected to enable the capture of dependencies across scales, a crucial aspect in fluid dynamics where phenomena manifest at multiple scales. At the end of the encoding phase, a GCN layer is applied again: this GCN layer serves as the bridge between the encoder and decoder, allowing for a seamless transition from the encoding process to the decoding process of the flow field.

Then, the decoder follows a reverse sequence to reconstruct the original high-dimensional flow field. The output of the bridging GCN layer undergoes an unpooling operation, which restores the number of nodes to what it was before the pooling performed in the encoder. Again, it is important to note that for this gUnpool, the information saved during the gpool should be received. After pooling, the decoder benefits from skip-connections that pass features from the corresponding level in the encoder to the decoder (Fig. \ref{fig:Unet}). These skip connections allow the decoder to effectively leverage the features from the encoding process, facilitating more accurate reconstructions. A GCN layer is then applied to the unpooled graph, learning to integrate the features from the previous decoder level with the features passed through the skip-connection. This unpooling-GCN block is repeated three more times with single channel (since the final output of the decoder should be a single channel snapshot), progressively increasing the number of nodes until the original graph size of 1,946 nodes is restored. At each of these steps, the skip-connections provide additional contextual information, enabling the GCN layers to effectively reconstruct the flow field features specific to that level of resolution.

The training process is conducted over a total of 7,500 epochs, using the Adam optimizer with an initial learning rate of $10^{-3}$. The learning rate is adaptively adjusted during training to facilitate convergence and optimize model performance. Additionally, we employ the early-stopping and checkpoint-saving mechanisms to prevent overfitting and retain the best-performing model weights.

To evaluate the accuracy of the trained models, the MSE of the unseen future snapshots is utilized. It is assessed using the averaged MSE values of all future snapshots predicted by the trained model as follows:

\begin{equation}\
\label{eq:error}
MSE=\cfrac{1}{S}\sum\limits_{s=1}^S\left(\cfrac{1}{N}\sum\limits_{n=1}^N(y_{n,s} - \hat{y}_{n,s})\right),
\end{equation}
where $\hat{y}_{n,s}$ and $y_{n,s}$ are the predicted value and ground truth at node $n$ for future snapshot $s$, respectively. $N$ is the total number of nodes, and $S$ is the number of future snapshots to be evaluated. This study adopts $S=100$ until Section \ref{sec:pool_ratio}, which indicates that the trained model is used to forecast the next 100 snapshots from the trained snapshot range --- while from Section \ref{sec:noise}, $S=200$ is used to evaluate longer rollout performance with improved Graph U-Nets.

\section{Enhanced Graph U-Nets for spatio-temporal flow prediction}
\label{sec:Improve}

\subsection{Effects of convolutional operators}
\label{sec:Conv}

In the original Graph U-Net paper \cite{gao2019graph}, the authors proposed the use of GCN operator for aggregating information from neighboring nodes. While GCN has demonstrated success in various graph-based tasks \cite{hamilton2017inductive, schlichtkrull2018modeling, ogoke2021graph, belbute2020combining}, it has an inherent limitation in its inability to consider complicated edge features as noted in Section \ref{sec:GMM_oper}. In the context of CFD problems, such as unsteady flow-field prediction, edge features play a crucial role in capturing the intricate interactions between nodes and the propagation of flow characteristics across the graph. In this regard, this section compares the Graph U-Nets with different convolutional operators: GCN and GMM. Since GMM has more degrees of freedom in terms of modeling the edge features, we try to investigate whether GMM indeed exhibits better performance than GCN in spatio-temporal flow prediction.

To investigate the impact of different GCN variants on the performance of the Graph U-Net model, we conduct experiments with three types of GCN operators: GCN without self-loop improvement (GCN), GCN with self-loop improvement (iGCN), and GCN with self-loop improvement and edge-weight consideration (iwGCN). The self-loop improvement refers to the addition of self-loops to the adjacency matrix ($\hat A = A + 2I$), as proposed by \citet{gao2019graph}, which helps to stabilize the training process and improve the Graph U-Net's performance. The edge-weight consideration indicates that $e_{j,i}$ in Eq. \ref{eq:GCN} is defined as $\sqrt{(x_i-x_j)^2+(y_i-y_j)^2}$, allowing for a more nuanced representation of the graph structure (see Section \ref{sec:GCN_oper} for further details). Note that when edge-weight consideration is not applied (such as GCN and iGCN operators), $e_{j,i}$ at the first GCN operator in encoder is defined as 1.

Furthermore, to address the limitations of GCN and explore alternative convolutional operators, we propose the use of the GMM convolutional operator as an alternative to GCN in the Graph U-Net architecture. It has the ability to account for edge features in a more sophisticated way than GCN, so it is expected to better provide accurate and physically meaningful representation of the flow characteristics. By incorporating the GMM operator into the Graph U-Net and comparing its performance with different GCN variants, we aim to provide a comprehensive analysis of the impact of convolutional operators on the model's ability to learn and predict the complex dynamics in unsteady flow fields. To explore the potential of GMM, its three variants with varying numbers of kernels are tested: GMMs with 1, 2, and 3 kernels (they are referred to as GMM-1, GMM-2, and GMM-3, respectively).

\begin{figure*}[htb!]
    \centering
    
    \begin{subfigure}[h]{0.48\textwidth}
        \centering
        \includegraphics[width=\textwidth]{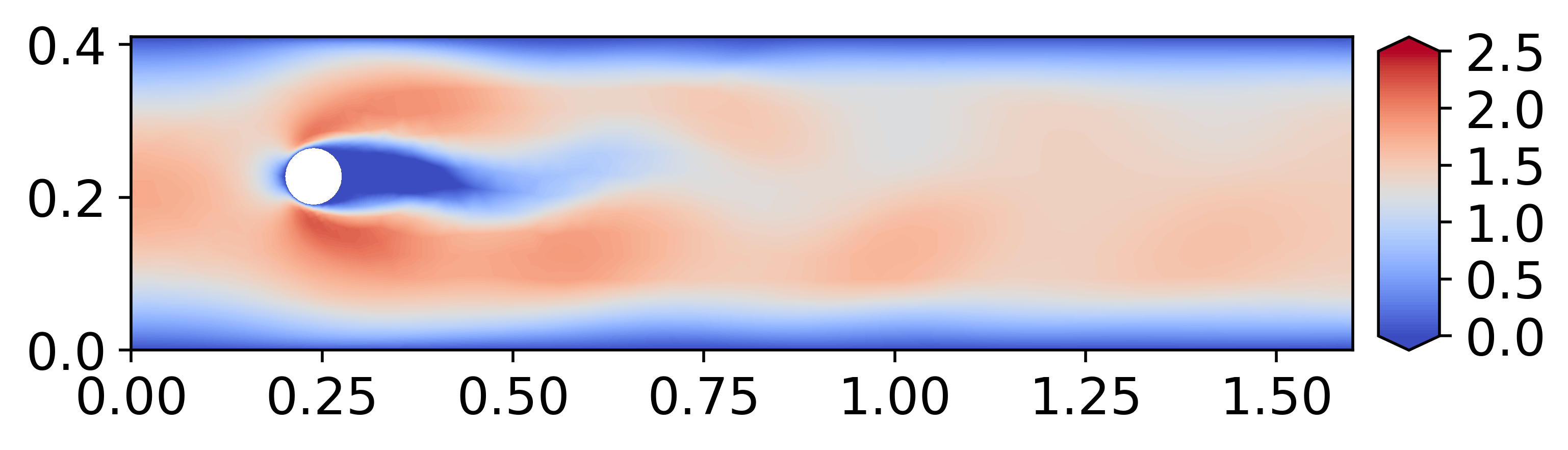}
        \caption{Ground truth}
    \end{subfigure}

    \vfill
    
    \begin{subfigure}[h]{0.48\textwidth}
        \centering
        \includegraphics[width=\textwidth]{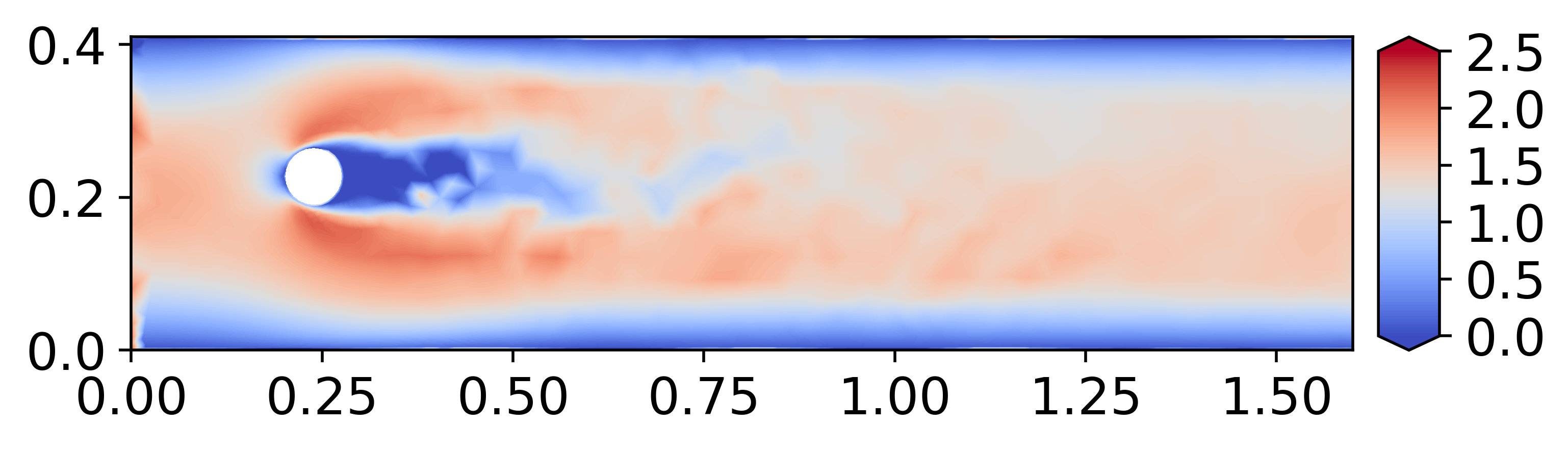}
        \caption{Prediction of iwGCN}
    \end{subfigure}
    \hfill
    \begin{subfigure}[h]{0.48\textwidth}
        \centering
        \includegraphics[width=\textwidth]{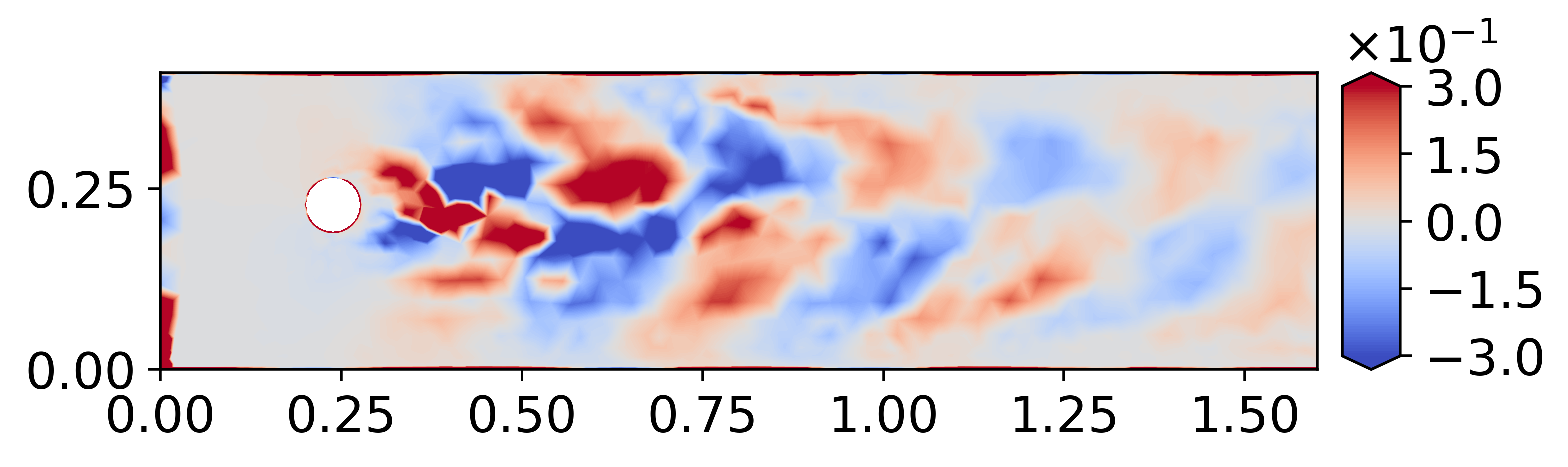}
        \caption{Error of iwGCN}
    \end{subfigure}

    \vfill
    
    \begin{subfigure}[h]{0.48\textwidth}
        \centering
        \includegraphics[width=\textwidth]{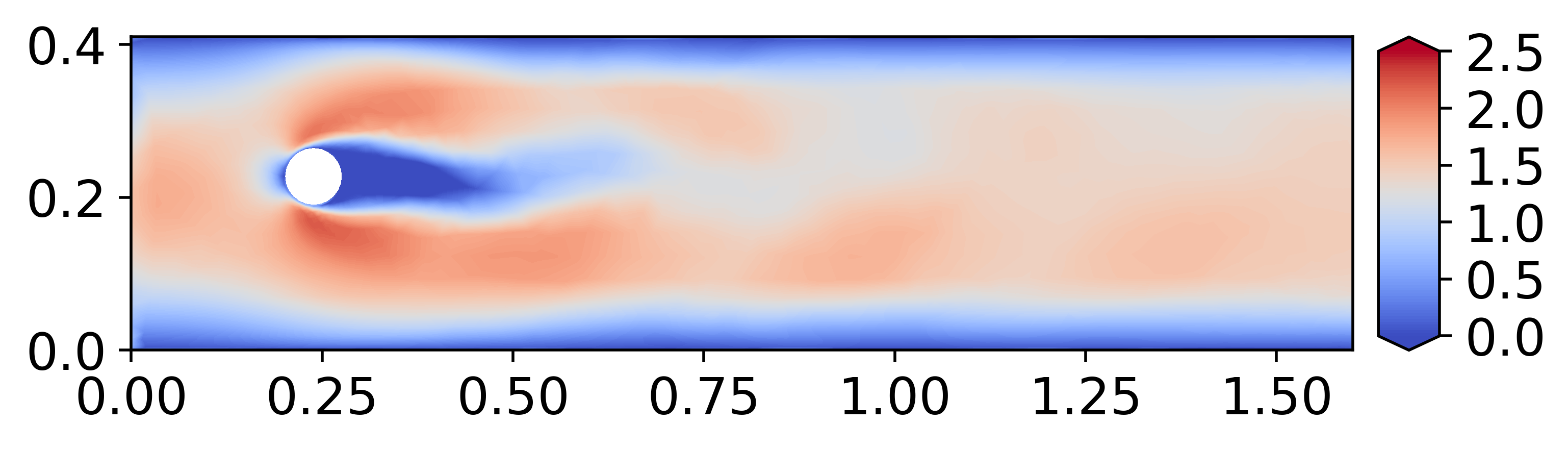}
        \caption{Prediction of GMM-1k}
    \end{subfigure}
    \hfill
    \begin{subfigure}[h]{0.48\textwidth}
        \centering
        \includegraphics[width=\textwidth]{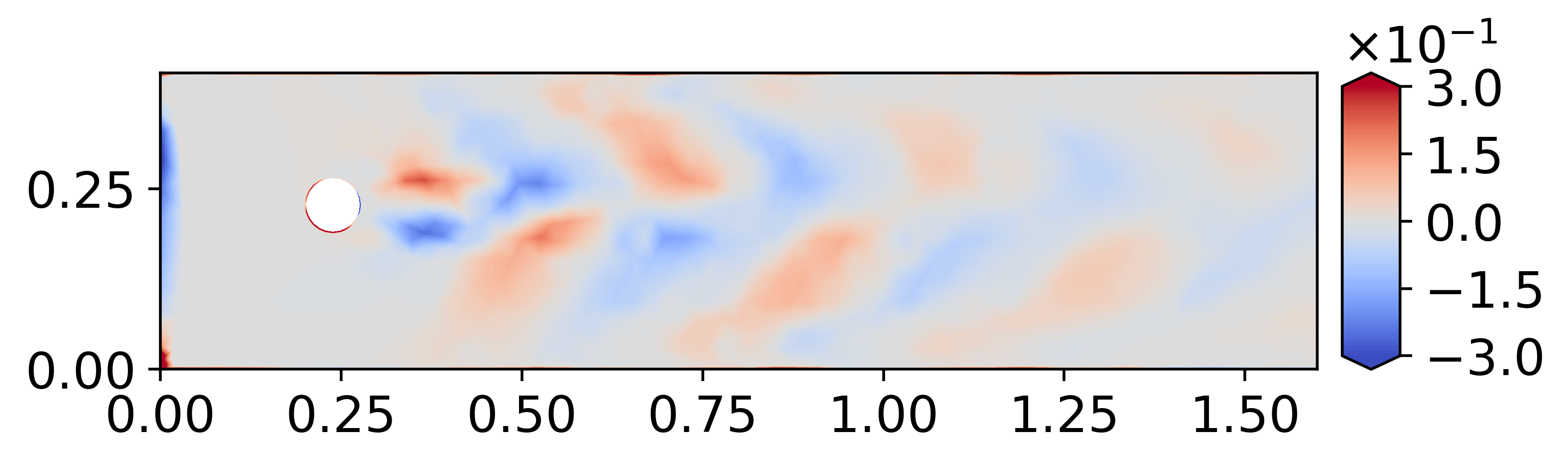}
        \caption{Error of GMM-1k}
    \end{subfigure}

    \caption{Visualization of the $x$-velocity fields 50 snapshots after the trained snapshot range: iwGCN and GMM-1k operators are compared.}
    \label{fig:GCN_flow}
\end{figure*}

The results of GCNs and GMMs can be found in Table \ref{tab:GCN}, where the MSE of all six models with different convolutaional types are presented. As previously documented \cite{gao2019graph}, the self-loop improvement helps to improve the model's performance: see iGCN and iwGCN. Also, when comparing iGCN and iwGCN, there is a clear improvement when edge weights are considered, indicating the limited effect of incorporating edge weights in the GCN operator. Despite the self-loop improvement and edge-weight consideration in the vanilla GCN, the GCN variants still underperform compared to the GMM operators, regardless of the number of kernels used, even though they have similar training times. In fact, GMM-1k shows an MSE of $0.005\times10^{-3}$, which is 95\% better than iwGCN. This is due to the GMM operator's ability to account for more complex physics in the edge features by using the mixture of Gaussians as the weight function. However, the results of the GMM operators show that simply increasing the number of kernels to exploit edge features in a more complicated way does not necessarily lead to improved performance: GMM-2k and GMM-3k have higher MSE than GMM-1k. This observation suggests that by increasing the number of kernels, the model can gain more flexibility and expressiveness theoretically, but its best selection would be highly problem-dependent due to overfitting issue. To visually compare the performance of GCNs and GMMs, iwGCN and GMM-1k are selected and Fig. \ref{fig:GCN_flow} shows their flow fields 50 snapshots after the trained snapshot range. As can be clearly seen, GMM-1k shows much better accuracy than iwGCN. Since the single-kernel GMM performs best, the GMM-1k operator is used in the following experiments. 

\renewcommand{\arraystretch}{1.1}
\begin{table}[htb!]
\centering
\begin{threeparttable}

\begin{tabular*}{0.7\textwidth}{@{\extracolsep{\fill}}c|cccccc}
\hline
\multirow{2}{*}{} & \multicolumn{6}{c}{Convolutional operator} \\ \cline{2-7}
 & GCN\tnote{1} & iGCN\tnote{2} & iwGCN\tnote{3} & GMM-1k\tnote{4} & GMM-2k\tnote{5} & GMM-3k\tnote{6} \\ \hline
MSE & 0.160 & 0.129 & 0.095 & \textbf{0.005} & 0.049 & 0.022 \\

Time & 0.84 & 0.82 & 0.84 & \textbf{0.81} & 0.83 & 0.84 \\ \hline
\end{tabular*}

\begin{tablenotes}\scriptsize
\item[1] Vanilla GCN \hfill \textsuperscript{4} GMM with 1 kernel 
\item[2] Improved GCN \hfill \textsuperscript{5} GMM with 2 kernels
\item[3] Improved and weighted GCN\hfill \textsuperscript{6} GMM with 3 kernels
\end{tablenotes}

\caption{MSE ($\times 10^{-3}$) over 100 rollout predictions with varying convolutional operators and their training time (hours) based on NVIDIA 3080 GPU: three GCN-based variants with different options and three GMM-based variants with different number of kernels.}
\label{tab:GCN}
\end{threeparttable}
\end{table}

The key point in this section is that simply using the traditional Graph U-Net architecture with the GCN operator (as first proposed by \citet{gao2019graph}) may not adequately address the complexities involved in unsteady-flow prediction. Our results also emphasize the critical need to select an appropriate operator that can more effectively represent the spatio-temporal features in the flow.

\subsection{Effects of pooling ratio}
\label{sec:pool_ratio}

In this section, we investigate the effects of the pooling ratio on the Graph U-Net model's performance. The pooling ratio determines the proportion of nodes that are selected during the pooling operation, effectively determining the degree of graph coarsening by controlling the number of nodes in the pooled graph. To study its impact on the performance of Graph U-Nets, we design various Graph U-Net models with pooling ratios ranging from 0.1 to 0.9, with increments of 0.1. Additionally, a variant of the model without any pooling/unpooling operation is included. Table \ref{tab:pool_ratio} presents the results of these experiments, showing the error values for each pooling ratio.

\begin{table}[htb!]
\centering
\begin{threeparttable}
\scriptsize 
\hfill \textsuperscript{1} Without pooling/unpooling
\small
\begin{tabular*}{0.8\columnwidth}{@{\extracolsep{\fill}}c|cccccccccc}
\hline
\multirow{2}{*}{} & \multicolumn{10}{c}{Pooling ratio} \\ 
& 0.1 & 0.2 & 0.3 & 0.4 & 0.5 & 0.6 & 0.7 & 0.8 & 0.9 & \textit{None}\tnote{1} \\ \hline
MSE & 5.1 & 5.1 & 2.9 & 4.0 & 7.4 & \textbf{2.7} & 3.7 & 28.9 & 34.0 & 94.4  \\ 
Time & 0.77 & 0.76 & 0.77 & 0.81 & 0.81 & 0.85 & 0.89 & 0.94 & 1.02 & \textbf{0.57} \\ \hline
\end{tabular*}
\caption{MSE ($\times 10^{-3}$) over 100 rollout after the trained snapshot range and training time (hours) based on NVIDIA 3080 GPU as pooling ratio varies. Note that pooling ratio \textit{None} indicates that the pooling/unpooling is not performed.}
\label{tab:pool_ratio}
\end{threeparttable}
\end{table}

The results demonstrate that the choice of pooling ratio has a significant impact on the Graph U-Net's performance. The model without the pooling/unpooling operation has an MSE of $94.4\times10^{-3}$, which is substantially higher than those of the models with pooling. It shows that graph coarsening is beneficial for capturing multi-scale features, highlighting the importance of incorporating pooling operations in the Graph U-Net architecture. Among the various pooling ratios, we observe that the MSE value with a pooling ratio of 0.6 is the best, indicating that this ratio provides an optimal balance between graph coarsening and preservation of spatial information in our case. Increasing the pooling ratio beyond 0.7 dramatically degrades the performance of the model, as insufficient coarsening can result in the failure to capture critical physical information within the dense mesh.

\begin{figure*}[htb!]
    \centering
    
    \begin{subfigure}[h]{0.48\textwidth}
        \centering
        \includegraphics[width=\textwidth]{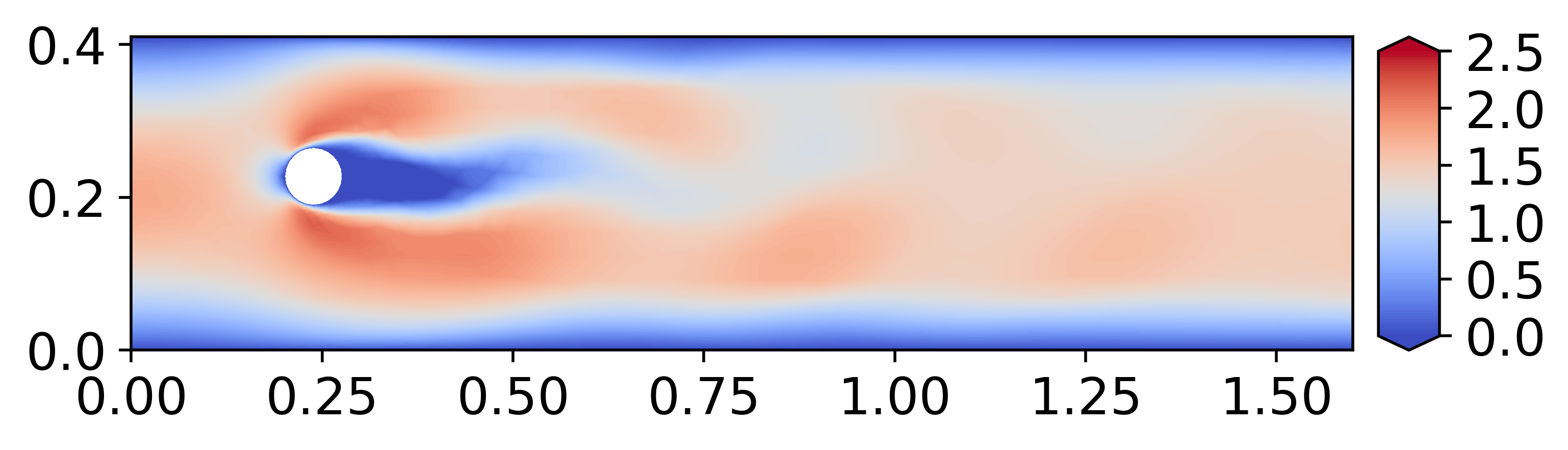}
        \caption{Ground truth}
    \end{subfigure}

    \vfill
    
    \begin{subfigure}[h]{0.48\textwidth}
        \centering
        \includegraphics[width=\textwidth]{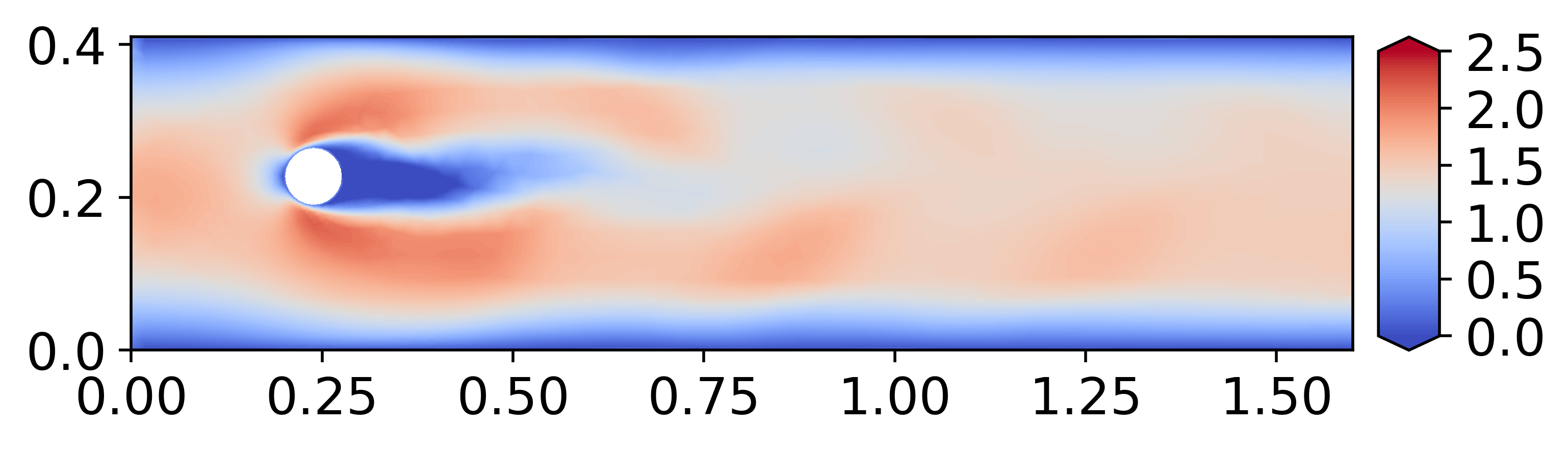}
        \caption{Prediction of the model with pooling ratio 0.6}\label{fig:pooling_flow_b}
    \end{subfigure}
    \hfill
    \begin{subfigure}[h]{0.48\textwidth}
        \centering
        \includegraphics[width=\textwidth]{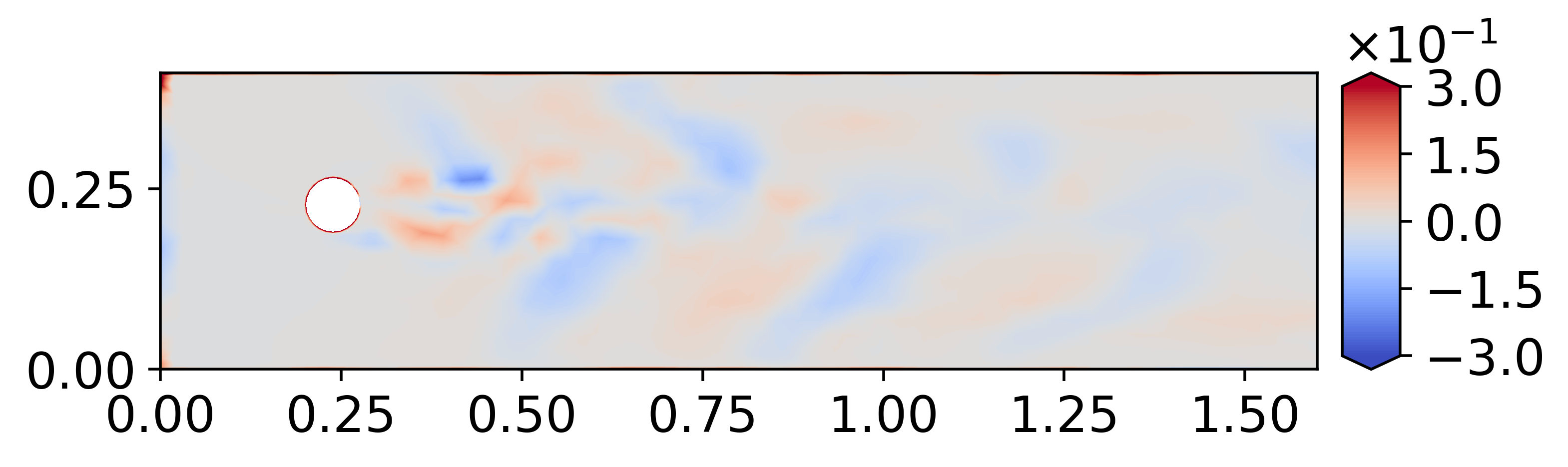}
        \caption{Error of the model with pooling ratio 0.6}
    \end{subfigure}

    \vfill
    
    \begin{subfigure}[h]{0.48\textwidth}
        \centering
        \includegraphics[width=\textwidth]{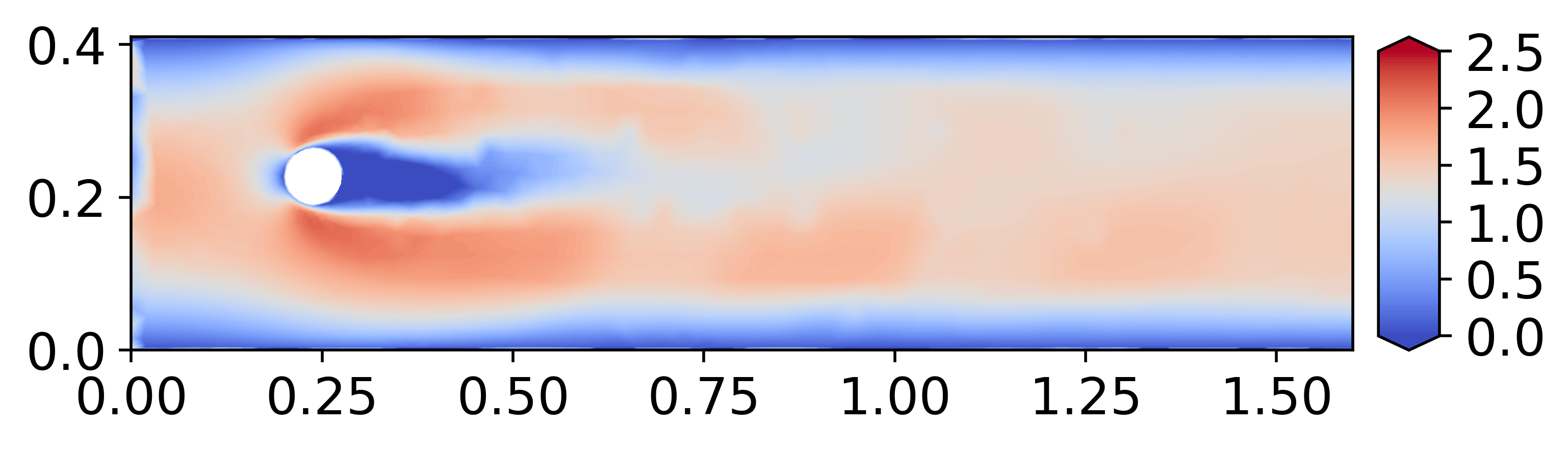}
        \caption{Prediction of the model without pooling operation}\label{fig:pooling_flow_d}
    \end{subfigure}
    \hfill
    \begin{subfigure}[h]{0.48\textwidth}
        \centering
        \includegraphics[width=\textwidth]{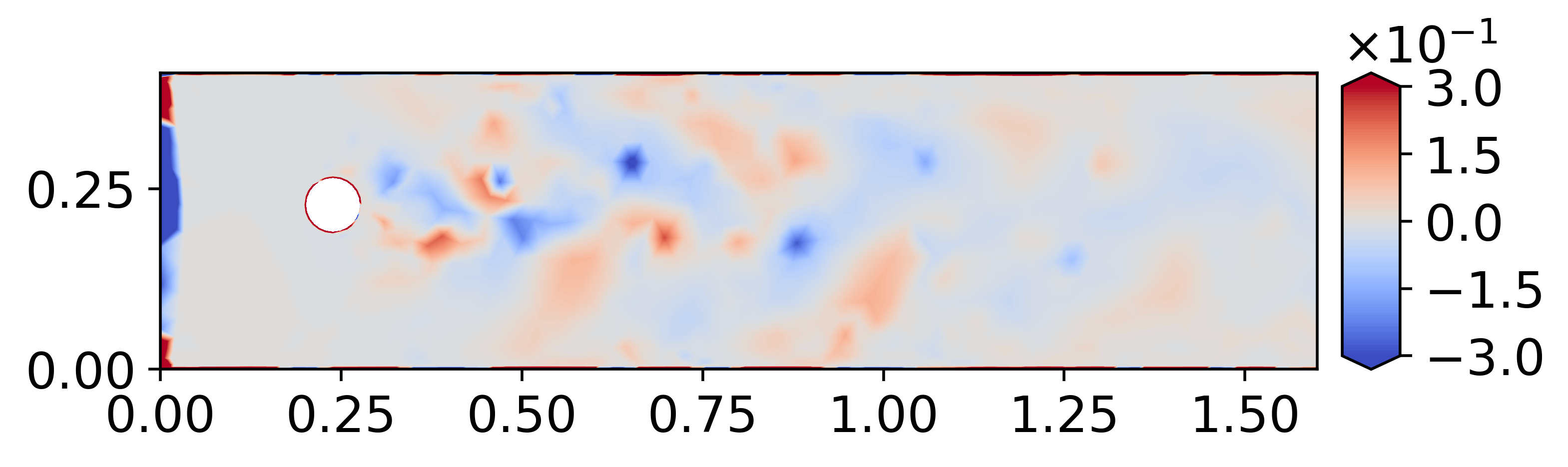}
        \caption{Error of the model without pooling operation}
    \end{subfigure}

    \caption{Visualization of the $x$-velocity fields 100 snapshots after the trained snapshot range: predicted by the model with pooling ratio of 0.6 and model without pooling/unpooling operation.}
    \label{fig:pooling_flow}
\end{figure*}

In terms of training time, there is a clear trend that as the pooling ratio increases, the training time increases incrementally due to the denser mesh that needs to be processed during encoding/decoding. It is worth noting that the model without pooling has the shortest training time: this is due to the absence of the gpool/gUnpool operation, which requires saving and transferring the pooling information. Finally, to demonstrate the effects of pooling, the predicted flow fields by the model with pooling ratio 0.6 and the model without pooling/unpooling are compared in Fig. \ref{fig:pooling_flow}. Again, it can be verified that while the model with pooling ratio 0.6 shows the successful results as in Fig. \ref{fig:pooling_flow_b}, the model without pooling shows a noisy flow field as in Fig. \ref{fig:pooling_flow_d}. Although the pooling/unpooling operation involves a higher computational cost, it should be noted that the computational burden of it is the price we pay for the significant increase in accuracy over model without pooling/unpooling. This trade-off between computational efficiency and prediction accuracy underscores the importance of selecting an appropriate pooling ratio, while also emphasizing the necessity of pooling operation in Graph U-Net architectures.

\begin{figure*}[htb!]
    \centering
    
    \begin{subfigure}[h]{0.48\textwidth}
        \centering
        \includegraphics[width=\textwidth]{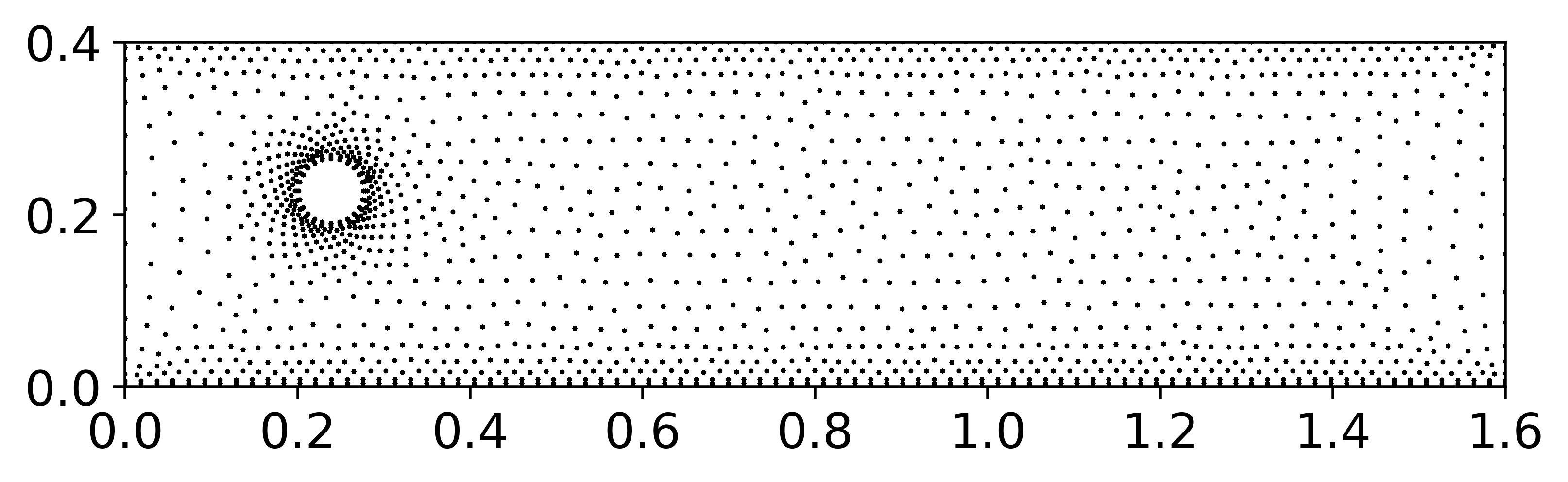}
        \caption{Nodes of original mesh}\label{fig:reduced_poolratio_a}
    \end{subfigure}
    
    \vfill
    
    \begin{subfigure}[h]{0.48\textwidth}
        \centering
        \includegraphics[width=\textwidth]{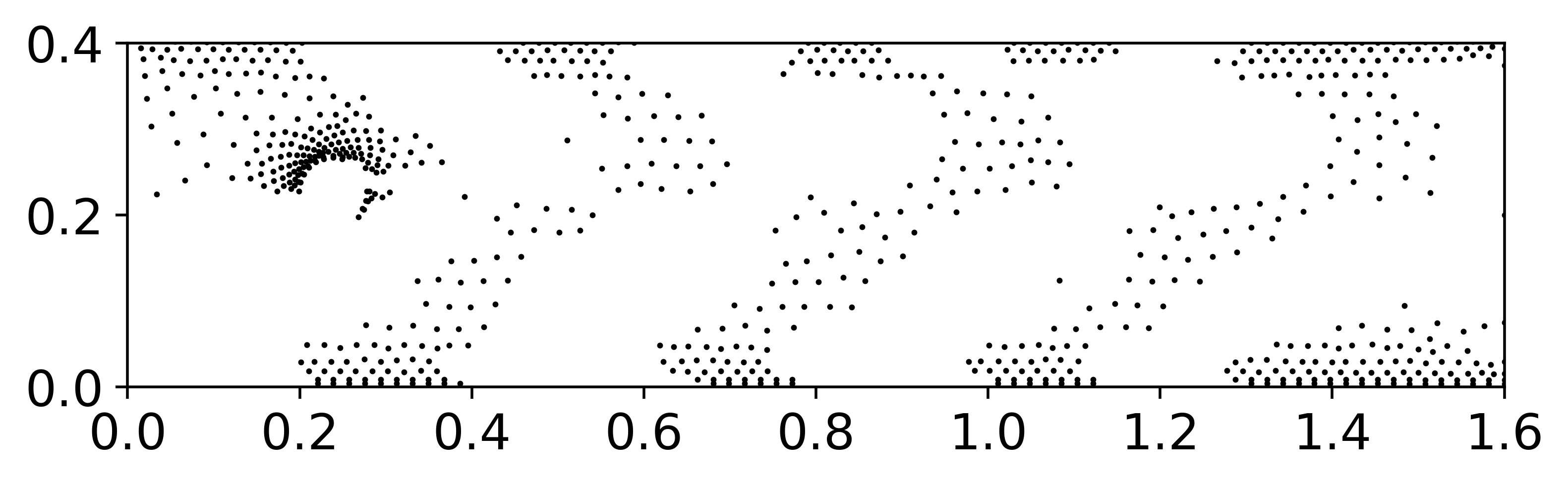}
        \caption{Pooled nodes after pooling once with the ratio of 0.4}
    \end{subfigure}
    \hfill
    \begin{subfigure}[h]{0.48\textwidth}
        \centering
        \includegraphics[width=\textwidth]{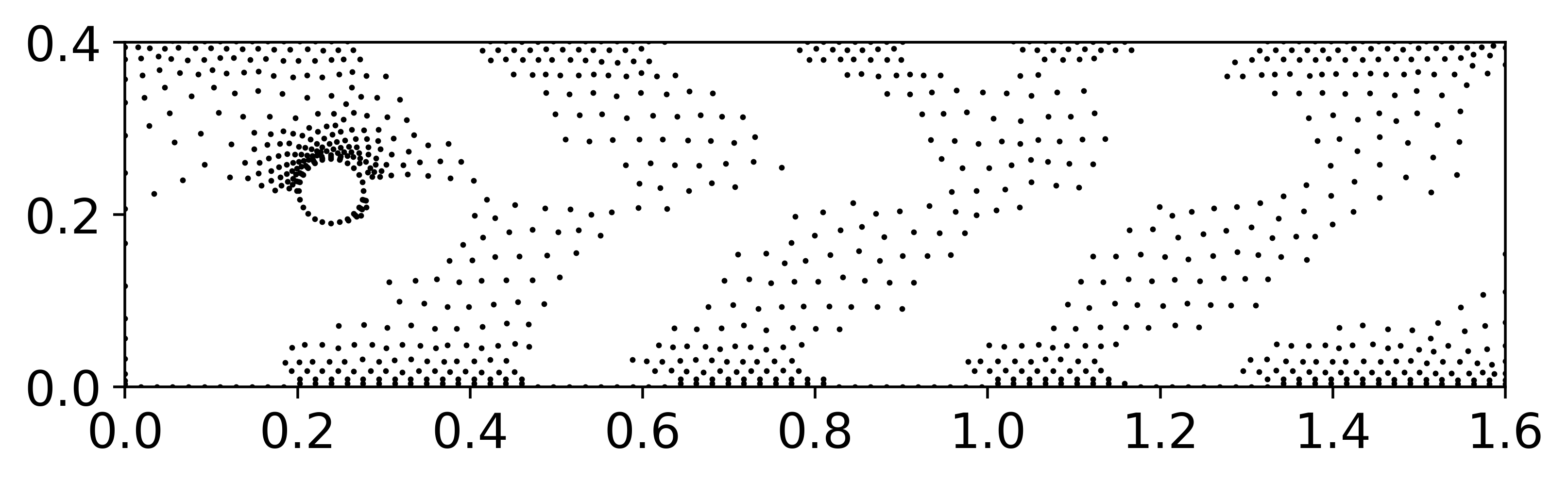}
        \caption{Pooled nodes after pooling once with the ratio of 0.6}\label{fig:reduced_poolratio_c}
    \end{subfigure}

    \vfill
    
    \begin{subfigure}[h]{0.48\textwidth}
        \centering
        \includegraphics[width=\textwidth]{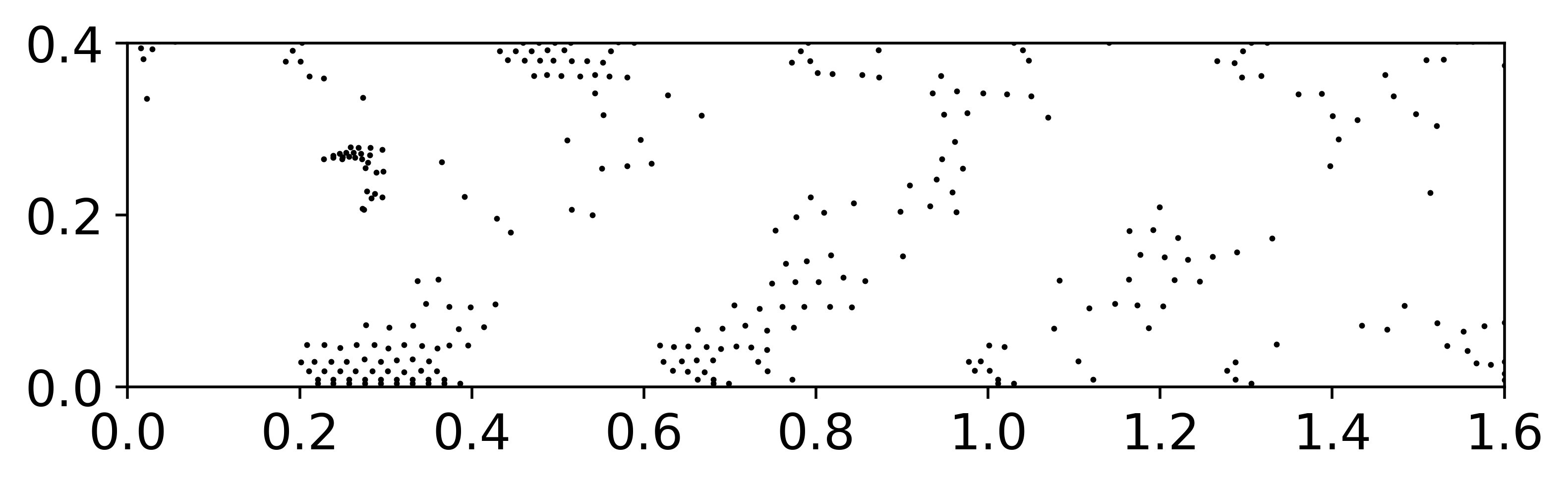}
        \caption{Pooled nodes after pooling twice with the ratio of 0.4}
    \end{subfigure}
    \hfill
    \begin{subfigure}[h]{0.48\textwidth}
        \centering
        \includegraphics[width=\textwidth]{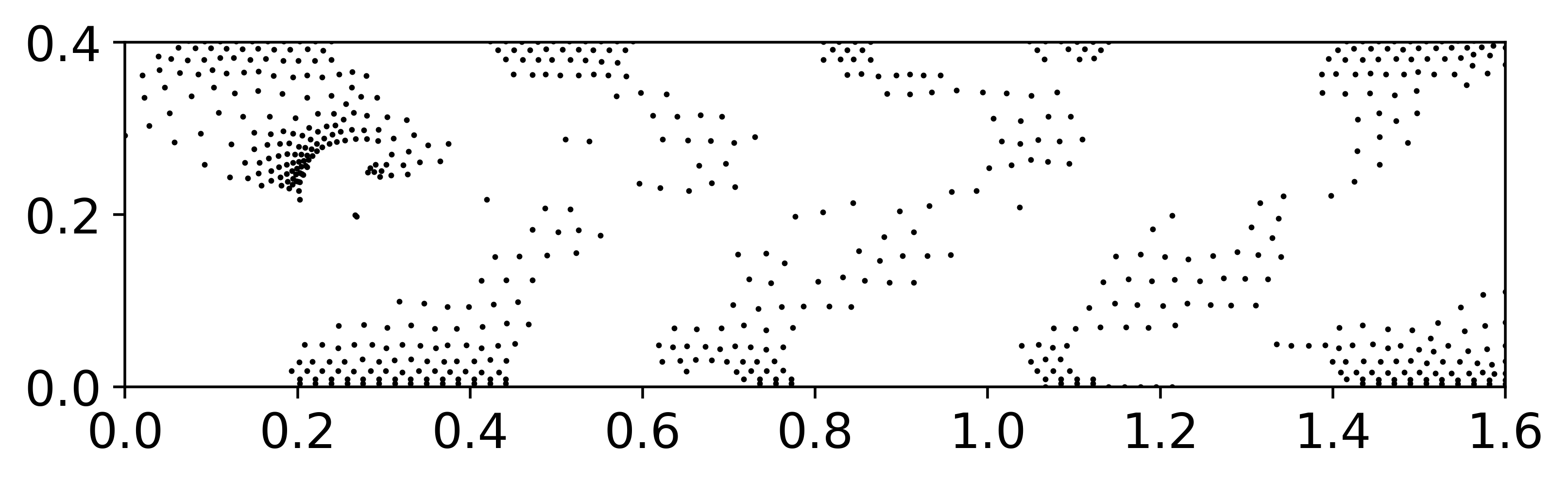}
        \caption{Pooled nodes after pooling twice with the ratio of 0.6}
    \end{subfigure}

    \vfill
    
    \begin{subfigure}[h]{0.48\textwidth}
        \centering
        \includegraphics[width=\textwidth]{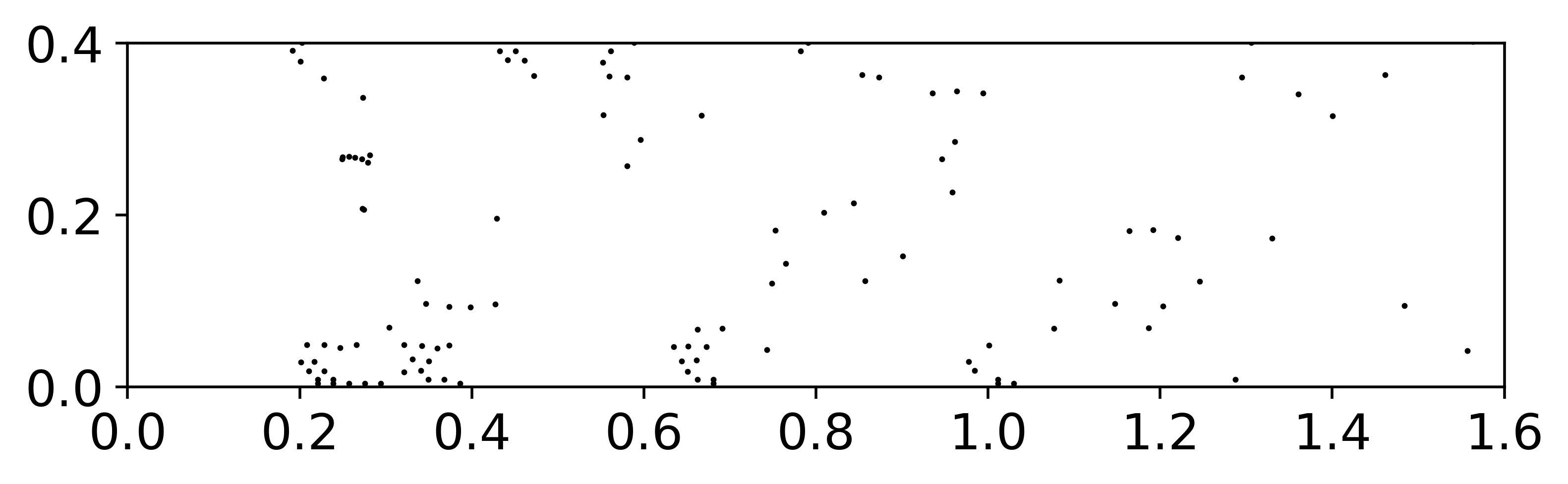}
        \caption{Pooled nodes after pooling three times with the ratio of 0.4}
    \end{subfigure}
    \hfill
    \begin{subfigure}[h]{0.48\textwidth}
        \centering
        \includegraphics[width=\textwidth]{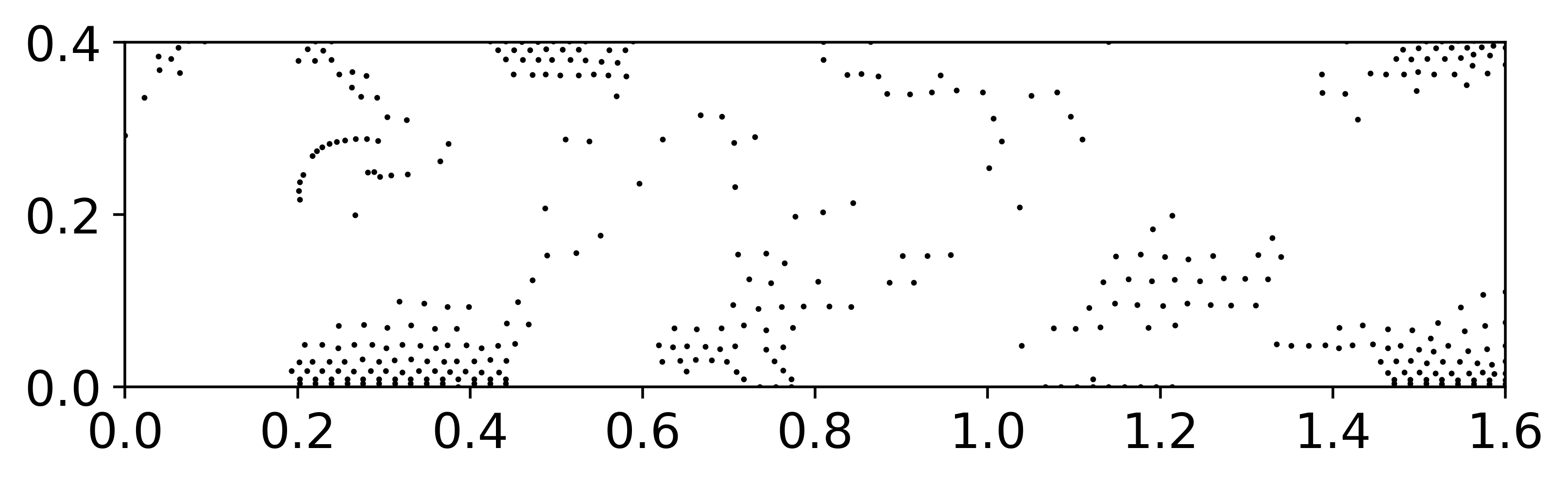}
        \caption{Pooled nodes after pooling three times with the ratio of 0.6}
    \end{subfigure}

    \vfill 

    \begin{subfigure}[h]{0.48\textwidth}
        \centering
        \includegraphics[width=\textwidth]{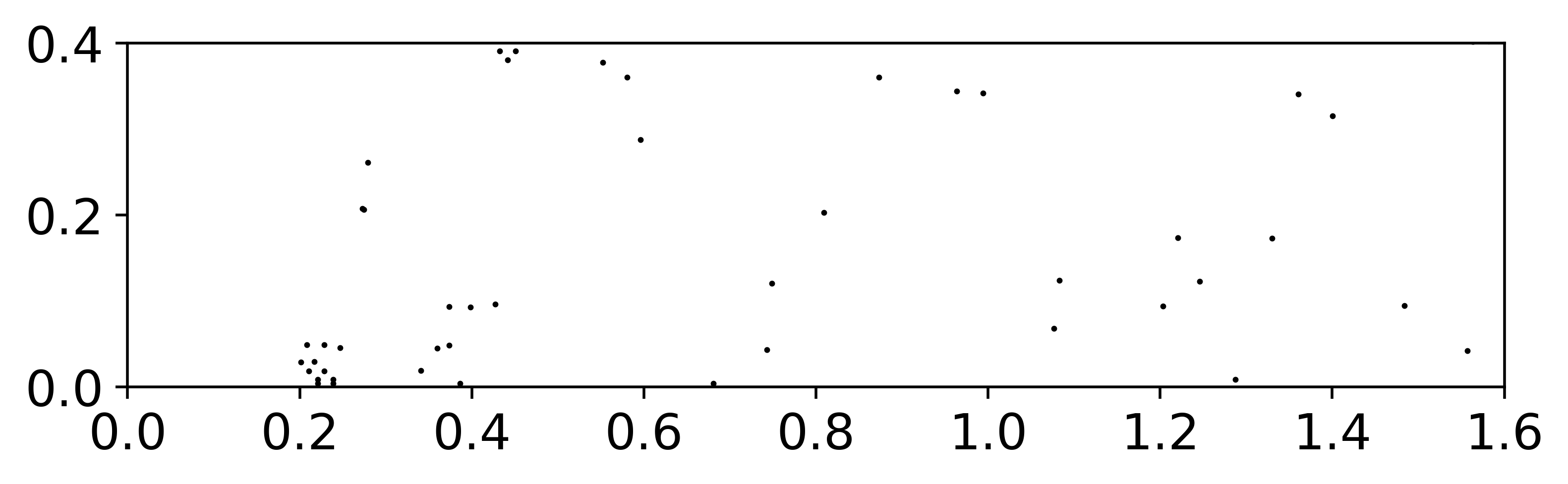}
        \caption{Pooled nodes after pooling four times with the ratio of 0.4}
    \end{subfigure}
    \hfill
    \begin{subfigure}[h]{0.48\textwidth}
        \centering
        \includegraphics[width=\textwidth]{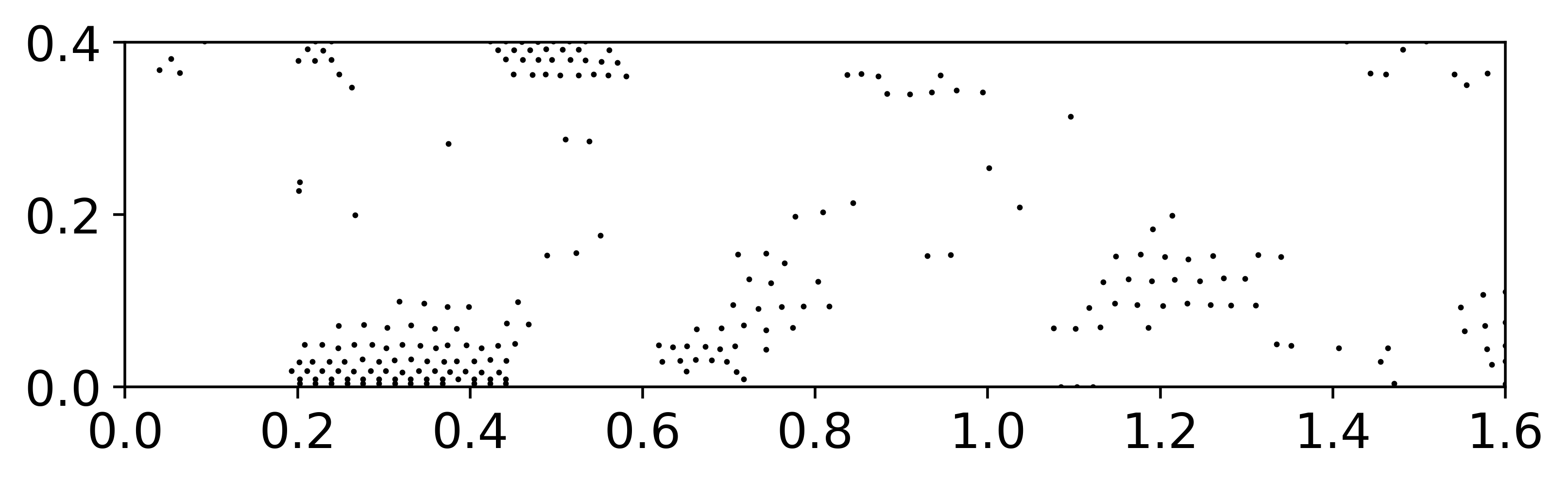}
        \caption{Pooled nodes after pooling four times with the ratio of 0.6}\label{fig:reduced_poolratio_i}
    \end{subfigure}


    \caption{Visualization of the pooled nodes at each level of the encoder for pooling ratios 0.4 (left column) and 0.6 (right column). The nodes of the original mesh are shown at the top.}
    \label{fig:reduced_poolratio}
\end{figure*}

To further illustrate how the pooling operation works during the node-selection process, we visualize the nodes retained after all pooling operations (four times in this study) in the trained encoders for two different pooling ratios: 0.4 and 0.6. Fig. \ref{fig:reduced_poolratio} shows the pooled nodes after each pooling operation of the encoder for these two pooling ratios. It can be observed that the pooling ratio significantly influences the number of nodes retained at each level of the encoder. With a pooling ratio of 0.4, the number of nodes is rapidly reduced, leading to a coarser graph representation as we move deeper into the encoder. This aggressive coarsening may result in the loss of important spatial information, especially in regions with complex flow patterns or fine-grained details. The interesting point is that with the pooling ratio of 0.6, although the last pooled nodes in Fig. \ref{fig:reduced_poolratio_i} are significantly sparse compared to the original nodes in Fig. \ref{fig:reduced_poolratio_a}, the decoder part of the Graph U-Net successfully restores the original full dimension with satisfactory flow fields, as can be found in Fig. \ref{fig:pooling_flow_b}. Based on the findings above, we select a pooling ratio of 0.6 as the optimal value for subsequent experiments, as it provides a good balance between graph coarsening and preservation of spatial information.

The visualization of the pooled nodes (Fig. \ref{fig:reduced_poolratio}) at different levels of the encoder reveals an intriguing pattern in the node-selection process. While the authors anticipated that nodes inside the vortex would be preserved, enhancing the Graph U-Net's ability to predict vortex shedding, the resulting pooled nodes do not exhibit a clear trend related to vortex shedding. Instead, they show an ambiguous pattern of alternating clustered nodes in the upper and lower regions along the x-axis (Fig. \ref{fig:reduced_poolratio_c}). Despite the lack of explicit physical meaning in the retained nodes as pooling progresses, it can be inferred that the pooling process captures some abstract physics within the vortex shedding phenomenon. To improve the interpretability of the gPool approach, future work could explore extending the pooling method to directly incorporate flow quantities, such as velocity gradient or vorticity magnitude, instead of relying on a trainable projection vector $p$.

\subsection{Effects of noise injection}
\label{sec:noise}

\subsubsection{Motivation for noise injection}
\label{sec:mot}

During the inference phase of the Graph U-Net used in this study, the model generates rollouts by feeding its own noisy predictions as input for predicting subsequent snapshot (Fig. \ref{fig:rollout}). However, during the training phase, the Graph U-Net was trained using clean, ground-truth snapshot data without any noise. This discrepancy of input snapshots during the training (without noise/error) and inference (with noise/error) phases can lead to substantial errors because the trained model has never been exposed to noisy snapshot data during training, but encounters noisy input snapshots during inference. This leads to errors in predicting the next snapshot, and these errors rapidly accumulate as the rollout process continues, causing the model's inputs to gradually deviate from the training distribution \cite{sanchez2020learning}. As a result, Graph U-Nets become susceptible to auto-regressive rollout when faced with noisy input data.

In this context, we propose to use the noise injection for the long-term prediction stability of the Graph U-Net model in spatio-temporal flow-field prediction tasks. Inspired by the work of \citet{sanchez2020learning}, who introduced random-walk noise into the inputs of the training data to intentionally corrupt the data and thus mitigate error accumulation over long rollouts, we adopt and extend this technique to improve the robustness of our model. We expect that by introducing noise into the training data, the training distribution of inputs will be closer to the input distribution encountered during rollouts in the inference phase, making the model more robust to noisy input snapshots and reducing error accumulation.

\subsubsection{Noise injection approaches}
\label{sec:NI_method}

\citet{sanchez2020learning}, who first proposed using random walk noise to train GNNs, injected the noise only into the input snapshots of the training dataset. Going one step further, the present study also investigates the scenario where the noise is injected on both the input and output of the training data. Therefore, two different approaches are investigated:

\begin{enumerate}

\item Input noise injection (I-noise): in this approach, we corrupt only the input node feature ($x$-velocity in our case) with Gaussian noise. The noise is added during the training, simulating the scenario where the model receives noisy snapshots as input data during rollout in the inference phase.
\item Input-Output noise injection (I/O-noise): in addition to corrupting the input node feature, we also inject noise into the output node feature during training. This approach aims to prevent the model from overfitting to the exact ground truth snapshot, potentially encouraging it to learn more generalized patterns and improving its stability during long-term predictions.

\end{enumerate}

To investigate the impact of noise injection on the model's performance, we compare three variants of the Graph U-Net model: (1) without noise injection (conventional approach), (2) with I-noise, and (3) with I/O-noise. Furthermore, we recognize that the magnitude of the injected noise, $\sigma$ in Gaussian noise $\mathcal{N}(0, \sigma^2)$, will be a critical hyperparameter that can significantly influence the model's behavior. To find the optimal noise magnitude, we explore a range of noise sizes ($\sigma$), specifically 0.001, 0.01, 0.02, 0.04, 0.08, 0.16, 0.32, and 0.64, which are selected considering the scale of our dataset.

\begin{figure*}[htb!]
    \centering    
    \begin{subfigure}[h]{0.48\textwidth}
        \centering
        \includegraphics[width=\textwidth]{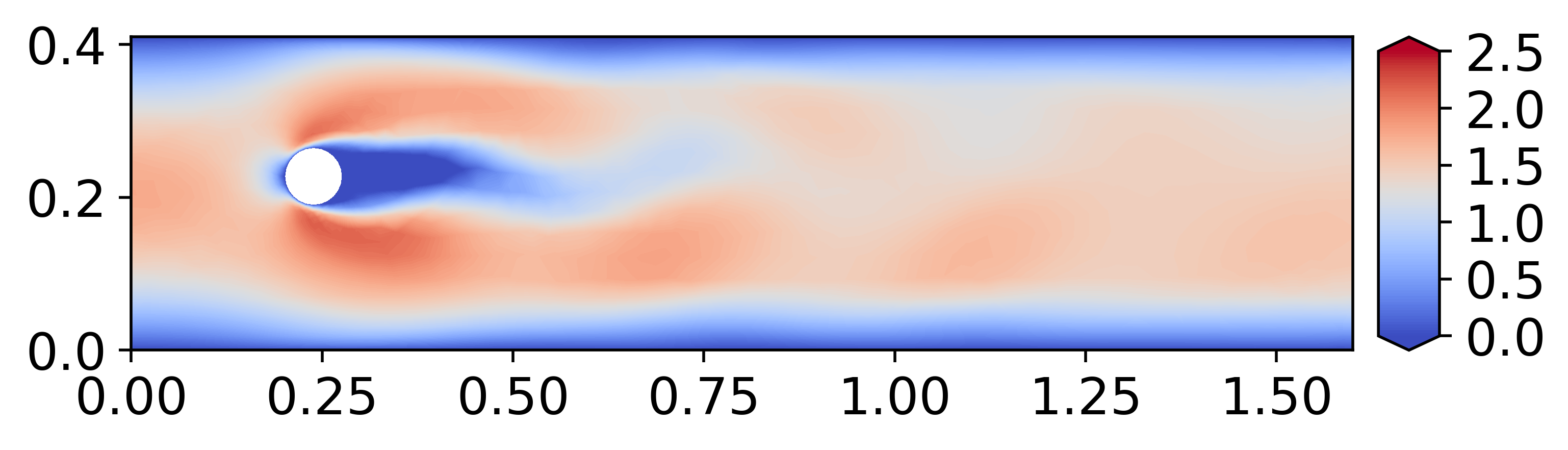}
        \caption{Original flow field before noise injection}
    \end{subfigure}
    \hfill
    \begin{subfigure}[h]{0.48\textwidth}
        \centering
        \includegraphics[width=\textwidth]{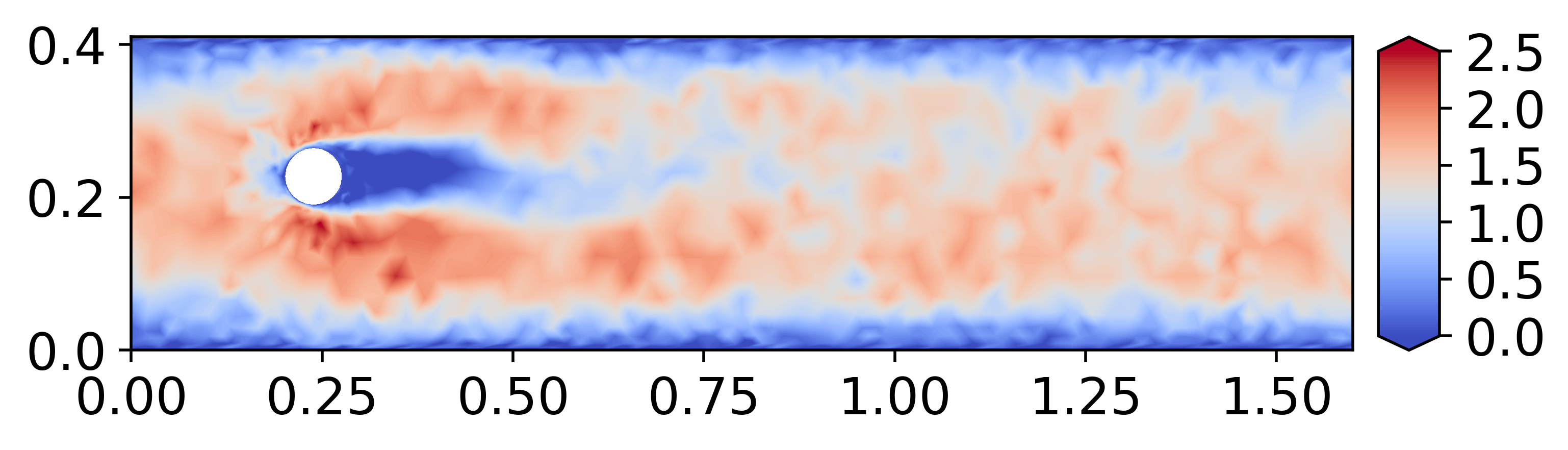}
        \caption{Corrupted flow field after noise injection with noise size 0.16}\label{fig:noised_flow_b}
    \end{subfigure}

    \caption{Visual comparison of the flow fields before and after the noise injection: the best noise case with the size of 0.16 is shown as an example.}
    \label{fig:noised_flow}
\end{figure*}

Table \ref{tab:NI_rollout200} presents the results, showing the MSE metric for each model variant and noise size over 200 rollout inference (up to this section, the MSE was computed within 100 rollouts after the trained snapshot range but it increases to 200 rollouts since more robust rollouts are enabled with the noise injection approaches). The results demonstrate that both I-noise and I/O-noise lead to remarkable improvements in the long-term prediction stability of the Graph U-Net model compared to the conventional model without noise injection when appropriate noise size is selected. This finding highlights that injecting noise during training significantly helps the model become more robust to error accumulation during rollouts. Regarding the optimal noise size, a noise size of 0.16 yields the best performance for both I-noise and I/O-noise approaches, while I-noise shows slightly better performance than I/O-noise. With this noise level, the I-noise and I/O-noise show 85\% and 86\% improvement, respectively, over the model without noise injection that has an MSE of $6.26\times10^{-3}$. To visually inspect how the original flow field is corrupted by the noise size of 0.16, Fig. \ref{fig:noised_flow} compares the flow field before and after the noise injection. It is worth noting that even with this corrupted level in Fig. \ref{fig:noised_flow_b}, the Graph U-Net shows a much better performance than that without noise injection. In any case, for the dataset used in this study, this noise size can be considered to provide sufficient perturbations to simulate the inference distribution while still allowing the model to learn the essential features of the training data --- when excessive noise is injected, such as the cases of 0.32 and 0.64, the error metrics skyrocket, highlighting the importance of choosing the right level of noise considering the scale of the trained data. Based on these results, we select the noise size of 0.16 as the optimal value for subsequent experiments and the effects of the noise injection will be highlighted again in Section \ref{sec:generalization}.

\begin{table}[htb!]
\centering
\begin{tabularx}{0.8\columnwidth}{@{\extracolsep{\fill}}c|cccccccc}
\cline{1-9}
\multirow{2}{*}{Noise type} & \multicolumn{8}{c}{Noise size} \\ \cline{2-9}
& 0.001 & 0.01 & 0.02 & 0.04 & 0.08 & 0.16 & 0.32 & 0.64 \\ \cline{1-9}
Without noise & 6.26 & - & - & - & - & - & - & - \\
I-noise & 7.05 & 6.43 & 6.73 & 2.84 & 13.10 & \textbf{0.85} & 20.90 & 55.08 \\
I/O-noise & 9.10 & 4.65 & 5.10 & 2.24  & 16.47 & \textbf{0.92} & 11.75 & 48.02 \\ \cline{1-9}
\end{tabularx}
\caption{MSE ($\times 10^{-3}$) over 200 rollout after the trained snapshot range as noise injection approach varies. Note that the result without noise reports a single value since it does not depend on the noise size.}
\label{tab:NI_rollout200}
\end{table}

Our findings offer insights into the application of noise injection techniques in graph-based models for fluid-dynamics simulations. Both the I-noise and I/O-noise approaches significantly enhance the model's robustness to error accumulation during rollouts. This study highlights the importance of selecting an appropriate noise injection approach and carefully tuning the noise size based on the specific problem and data scale to optimize the performance of Graph U-Nets in unsteady flow prediction tasks.

\section{Transductive and inductive performance of enhanced Graph U-Nets}
\label{sec:generalization}

This section focuses on exploring the generalization capabilities of Graph U-Nets in unsteady flow-field prediction, extending the Section \ref{sec:Improve} where the temporal prediction performance was discussed only on the trained mesh scenario. This investigation is divided into two parts: transductive learning and inductive learning. In Section \ref{sec:trans}, we examine the transductive-learning performance, which aims to predict quantities for unseen node regions within the same Graph Used during training. Subsequently, Section \ref{sec:induct} delves into the inductive-learning performance, assessing the model's ability to generalize to completely new (unseen/untrained) graphs based on the learned representations from the trained graphs. This evaluation is conducted using different graphs that were not included in the training process of the Graph U-Nets: this inductive-learning performance is crucial for real-world applications, as the model may frequently encounter novel geometries and flow conditions that differ from those seen during training. By assessing both transductive and inductive-learning performance, we aim to provide a comprehensive understanding of the generalization capabilities of Graph U-Nets in handling diverse and unseen scenarios in spatio-temporal flow-field prediction.

\subsection{Transductive-learning settings}
\label{sec:trans}

\begin{figure*}[htb!]
    \centering

    \begin{subfigure}[h]{0.48\textwidth}
        \centering
        \includegraphics[width=0.9\textwidth]{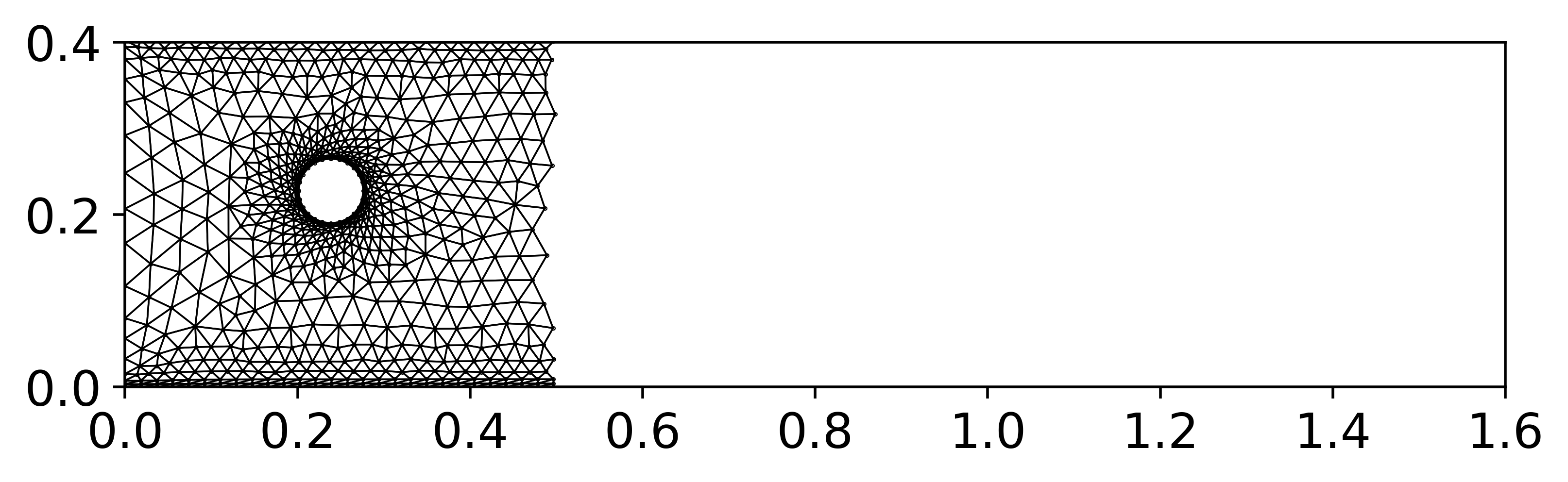}
        \caption{Mesh for Trans 1 case: contains nodes in $x<0.5$ }\label{fig:trans_mesh_a}
    \end{subfigure}
    \hfill
    \begin{subfigure}[h]{0.48\textwidth}
        \centering
        \includegraphics[width=0.9\textwidth]{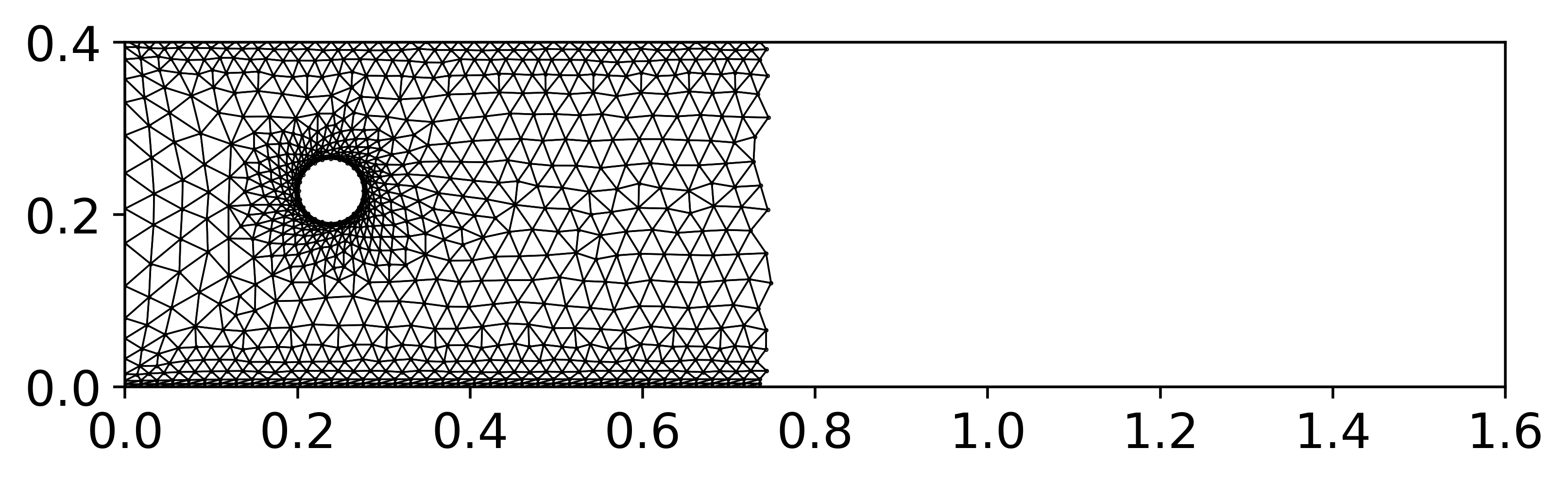}
        \caption{Mesh for Trans 2 case: contains nodes in $x<0.75$}\label{fig:trans_mesh_b}
    \end{subfigure}

    \vfill
    
    \begin{subfigure}[h]{0.48\textwidth}
        \centering
        \includegraphics[width=0.9\textwidth]{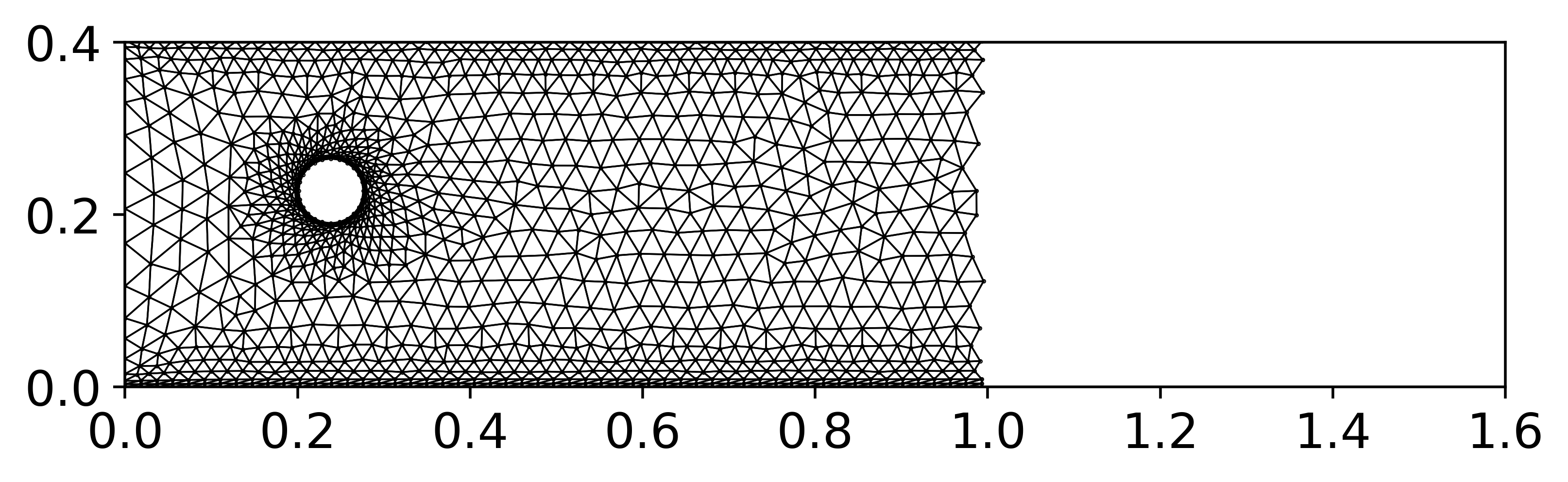}
        \caption{Mesh for Trans 3 case: contains nodes in  $x<1$}\label{fig:trans_mesh_c}
    \end{subfigure}
    \hfill
    \begin{subfigure}[h]{0.48\textwidth}
        \centering
        \includegraphics[width=0.9\textwidth]{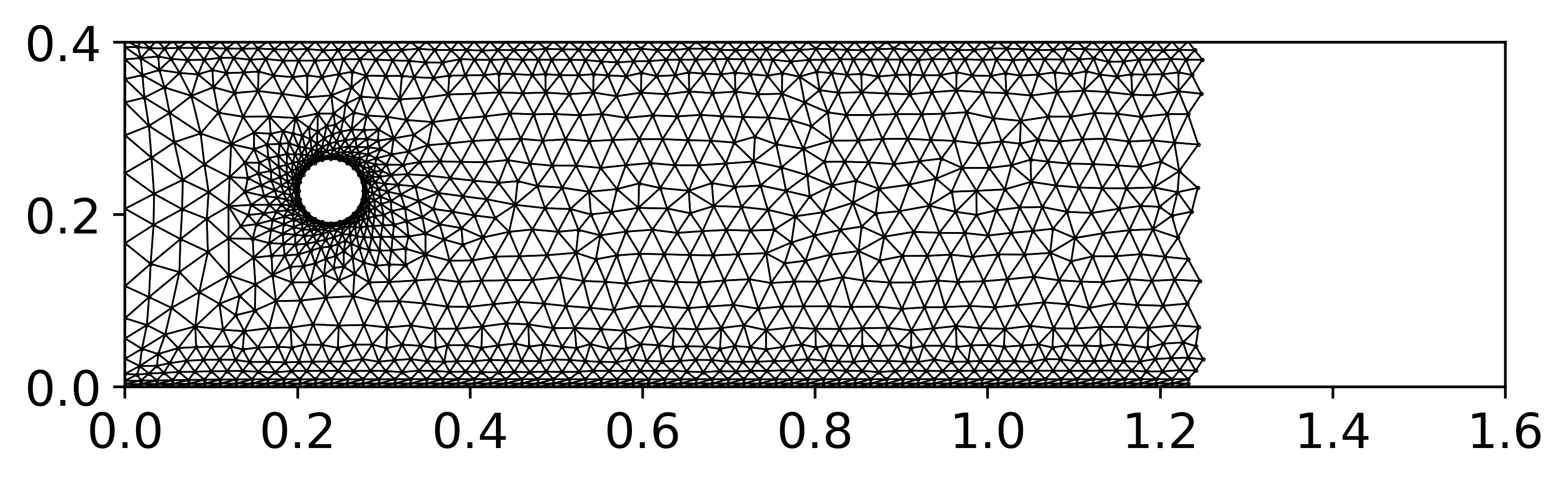}
        \caption{Mesh for Trans 4 case: contains nodes in $x<1.25$}\label{fig:trans_mesh_d}
    \end{subfigure}

    \caption{Visualization of the meshes in four transductive cases: as the numbering of the cases increases, a larger portion of the $x$-domain is trained.}
    \label{fig:trans_mesh}
\end{figure*}

This section investigates the performance of the transductive learning in Graph U-Nets, which aims to predict flow quantities of unseen nodes within the same Graph Used for training. To evaluate the transductive performance, four mesh scenarios are adopted: Trans 1, Trans 2, Trans 3, and Trans 4. Each scenario is derived from the original mesh case used throughout this study (mesh in Fig. \ref{fig:mesh_trained}), also with the same flow fields, but with different mesh domains. Trans 1 has the smallest domain, containing only nodes with $x<0.5$ from the original graph (Fig. \ref{fig:trans_mesh_a}). Trans 2 includes nodes with $x<0.75$ (Fig. \ref{fig:trans_mesh_b}), Trans 3 has nodes with $x<1$ (Fig. \ref{fig:trans_mesh_c}), and Trans 4 comprises nodes with $x<1.25$ (Fig. \ref{fig:trans_mesh_d}). These artificially truncated meshes are used to evaluate the transductive performance of Graph U-Nets. More specifically, Graph U-Nets are trained with these mesh scenarios and then tested to confirm how well they can predict the remaining flow fields (for example, the $x>0.5$ domain for the Trans 1 case) that are not seen during training. Note that I-noise approach with the noise size of 0.16 is adopted in this section.

\begin{figure*}[htb!]
    \centering

        \includegraphics[width=0.6\textwidth]{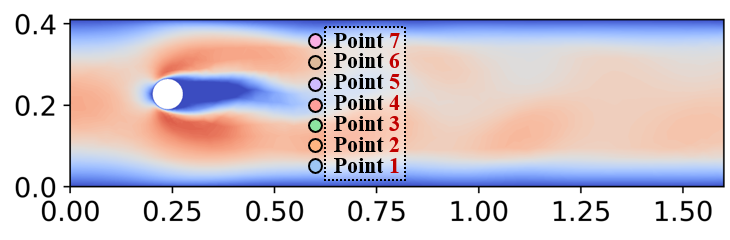}

    \caption{The locations of seven probe points to be examined for a more intuitive understanding of how vortex shedding is predicted in terms of $x$-velocity.}
    \label{fig:trans_zeroshot_a}
\end{figure*}

To visually present the results of the transductive-learning cases, the velocity profiles at the seven points behind the cylinder (predicted up to 200 snapshots after the trained snapshot range) will be plotted. Fig. \ref{fig:trans_zeroshot_a} shows the locations of the seven points that are used to investigate how well the Graph U-Net captures the wake characteristics (with coordinates [0.6, 0.05], [0.6, 0.1], [0.6, 0.15], [0.6, 0.2], [0.6, 0.25], [0.6, 0.3], and [0.6, 0.35]). 

\begin{figure*}[htb!]
    \centering
    \begin{subfigure}[h]{0.7\textwidth}
        \centering
        \includegraphics[width=\textwidth]{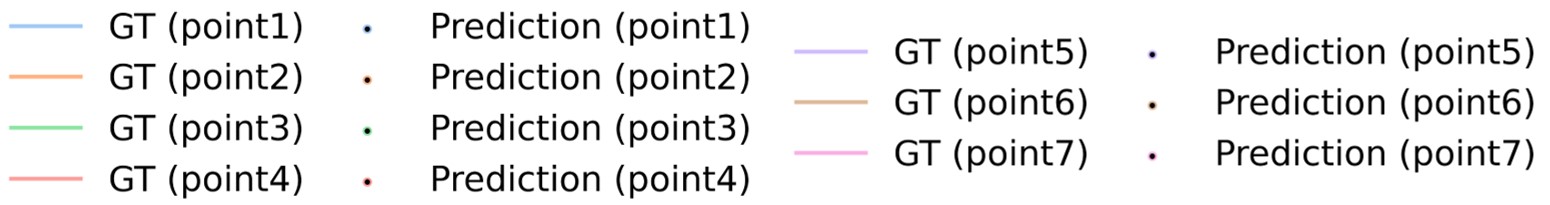}
    \end{subfigure}
    
    \begin{subfigure}[h]{0.35\textwidth}
        \centering
        \includegraphics[width=\textwidth,trim={0 0 0 7cm},clip]{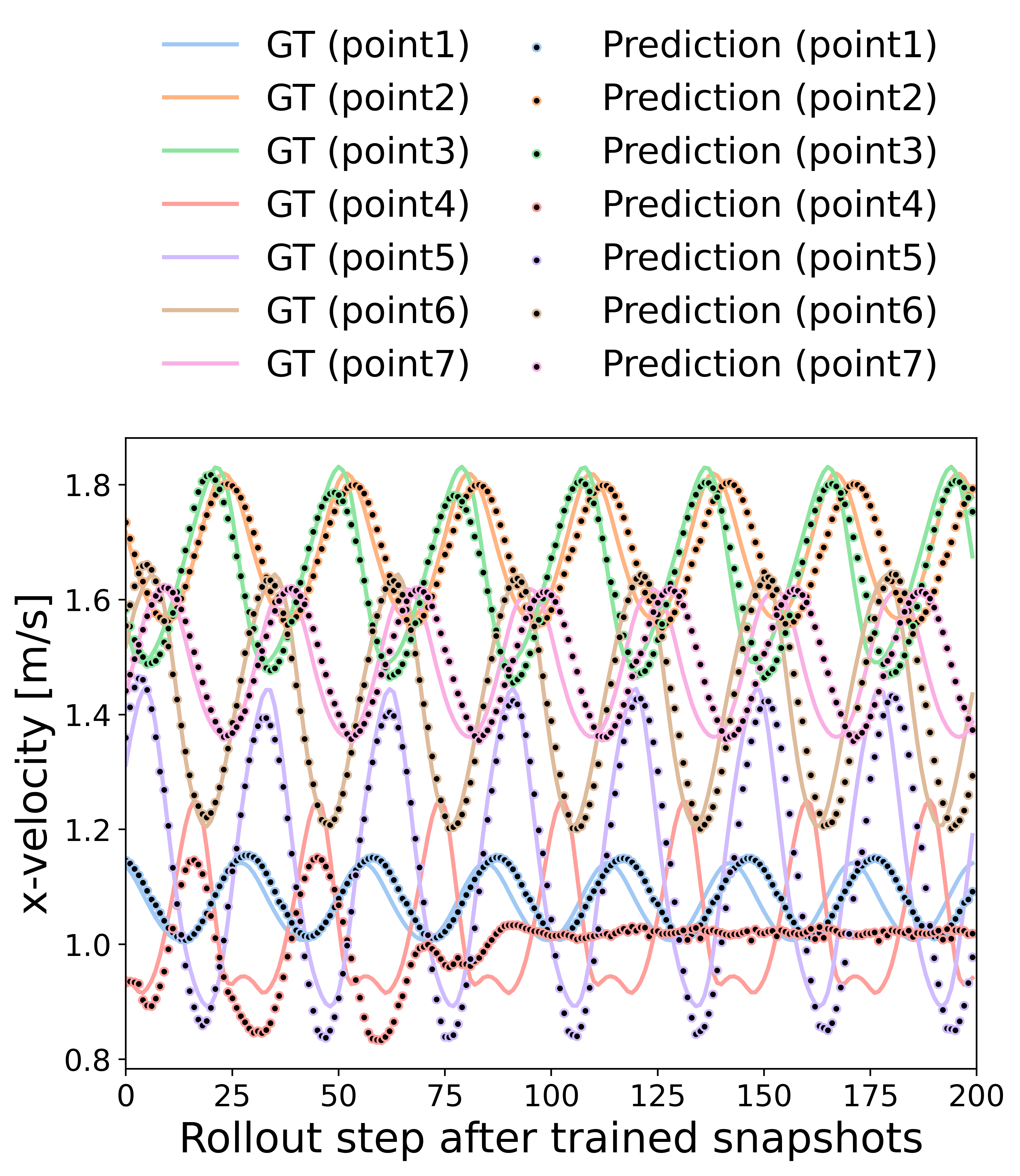}
        \caption{Trans 1 case}\label{fig:trans_zeroshot_b}
    \end{subfigure}
    \hspace{0.05\textwidth}
    \begin{subfigure}[h]{0.35\textwidth}
        \centering
        \includegraphics[width=\textwidth,trim={0 0 0 7cm},clip]{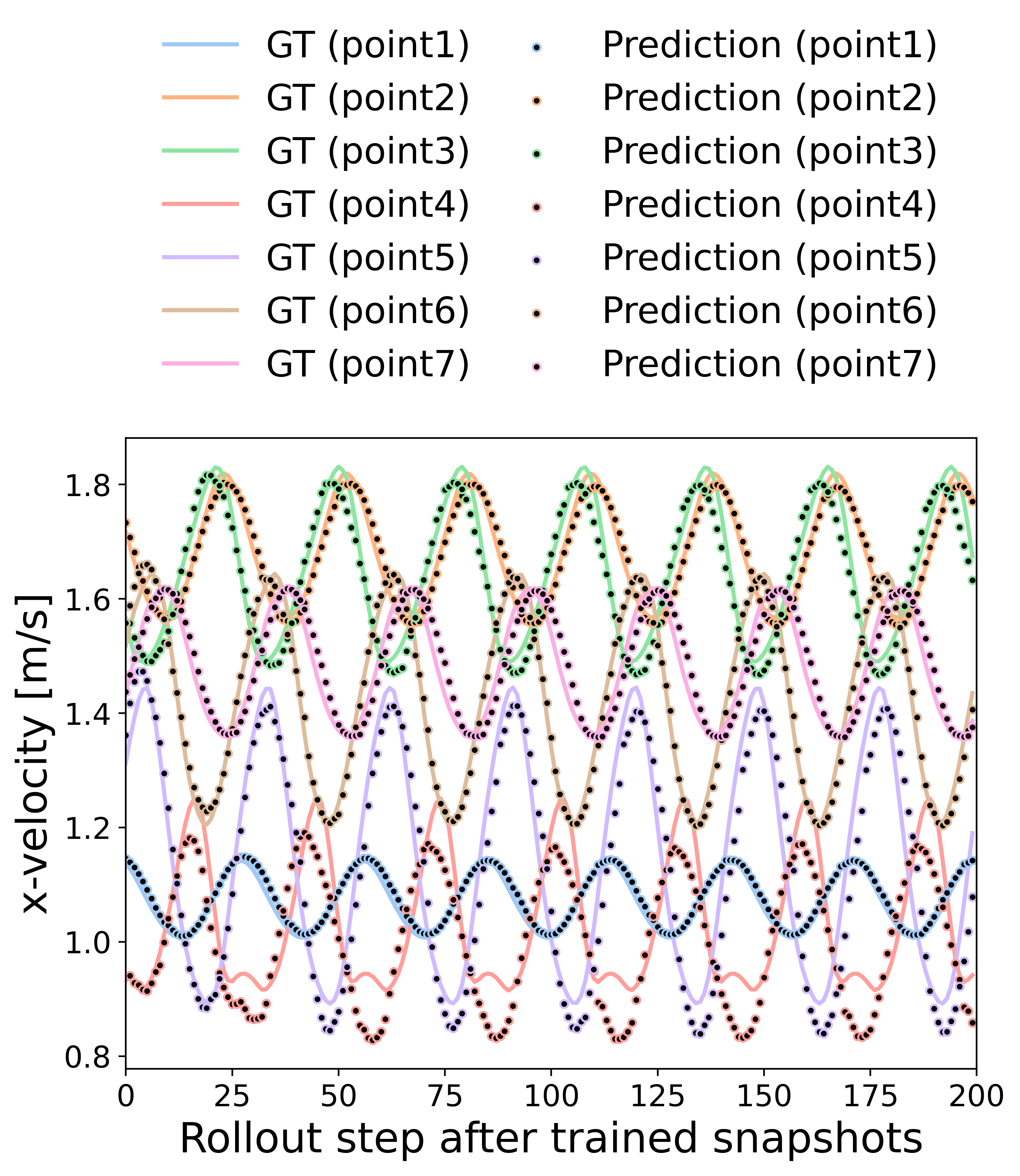}
        \caption{Trans 2 case}\label{fig:trans_zeroshot_c}
    \end{subfigure}
    
    \vfill
    
    \begin{subfigure}[h]{0.35\textwidth}
        \centering
        \includegraphics[width=\textwidth,trim={0 0 0 7cm},clip]{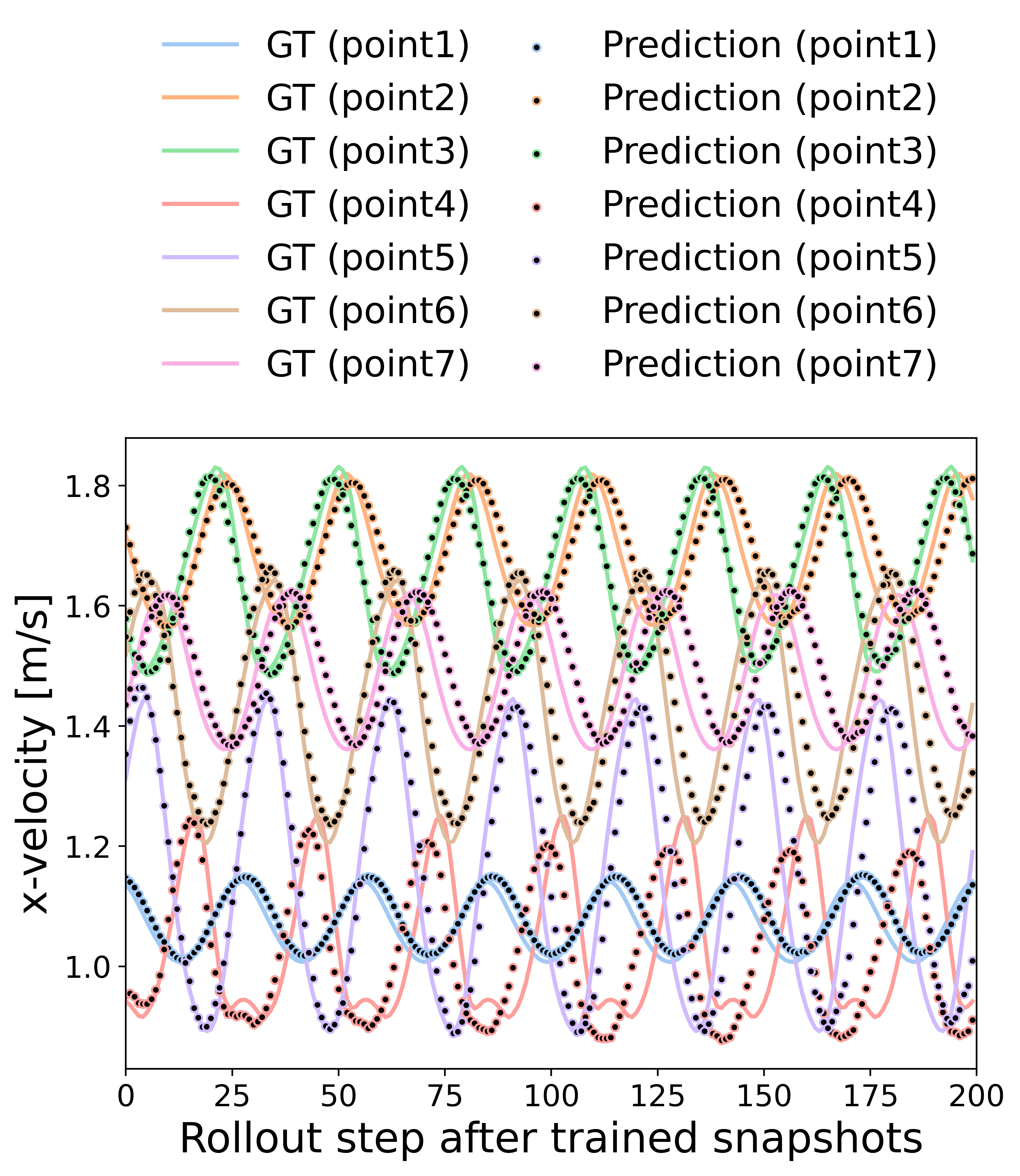}
        \caption{Trans 3 case}\label{fig:trans_zeroshot_d}
    \end{subfigure}
    \hspace{0.05\textwidth}
    \begin{subfigure}[h]{0.35\textwidth}
        \centering
        \includegraphics[width=\textwidth,trim={0 0 0 7cm},clip]{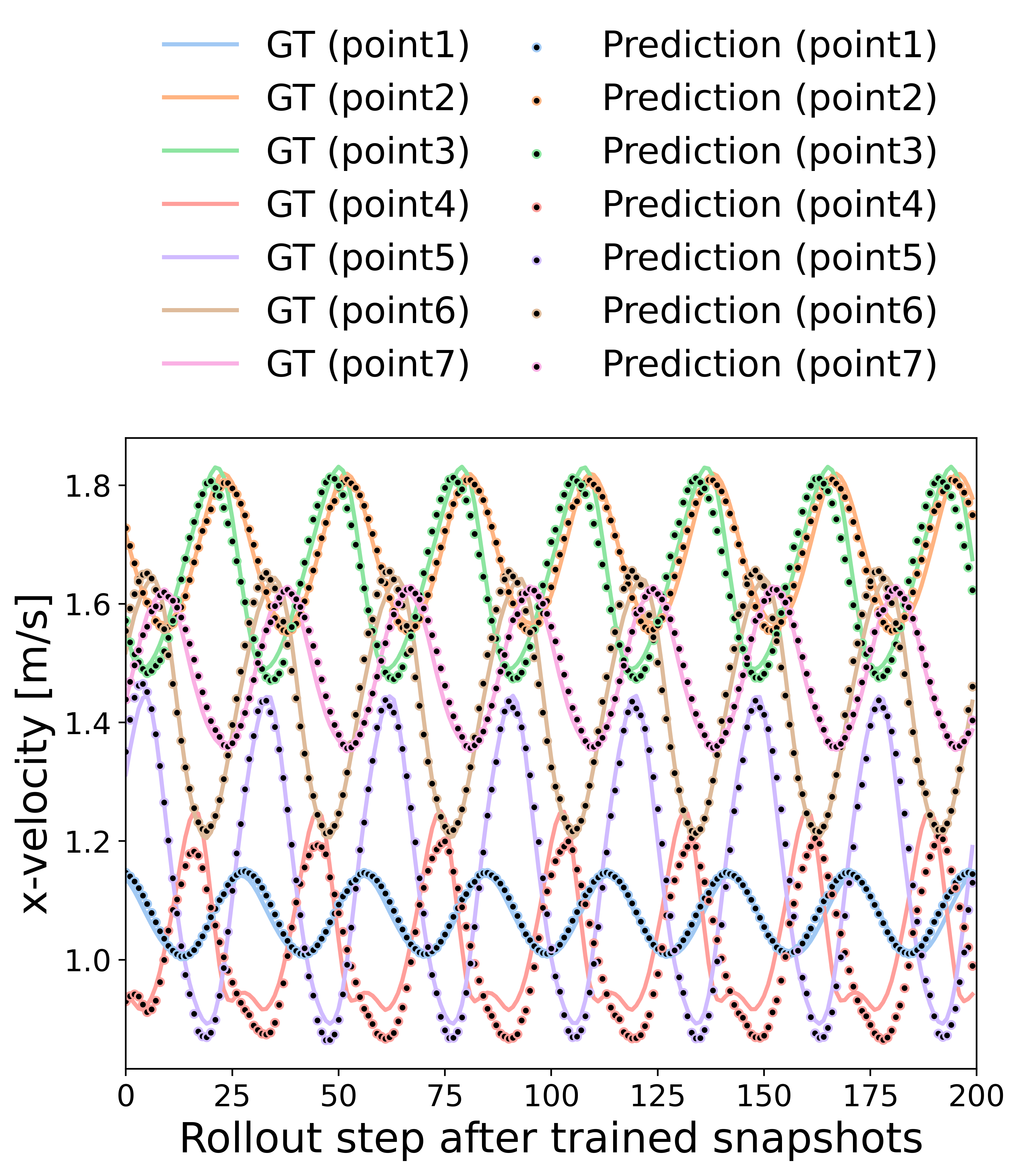}
        \caption{Trans 4 case}\label{fig:trans_zeroshot_e}
    \end{subfigure}


    \caption{Comparison of the predicted $x$-velocity at seven points (Fig. \ref{fig:trans_zeroshot_a}) in four transductive cases with I-noise approach. The velocity is predicted up to 200 snapshots ($x$-axis) after the trained snapshot range: (a) to (d) show the results from Trans 1 to Trans 4 cases.}
    \label{fig:trans_zeroshot}
\end{figure*}

\begin{figure*}[htb!]
    \centering
    \vspace{0.05\textwidth}
    
    \begin{subfigure}[h]{0.3\textwidth}
        \centering
        \includegraphics[width=\textwidth,trim={0 0 0 0},clip]{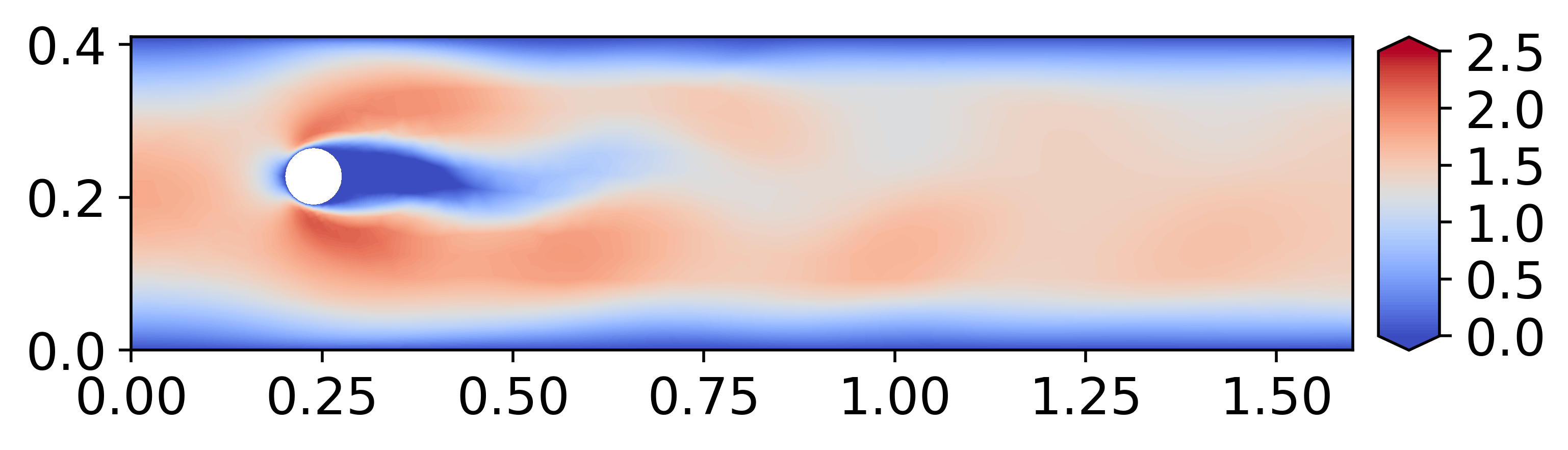}
        \caption{Ground truth after 50 rollouts}
    \end{subfigure}
    \hfill
    \begin{subfigure}[h]{0.3\textwidth}
        \centering
        \includegraphics[width=\textwidth,trim={0 0 0 0},clip]{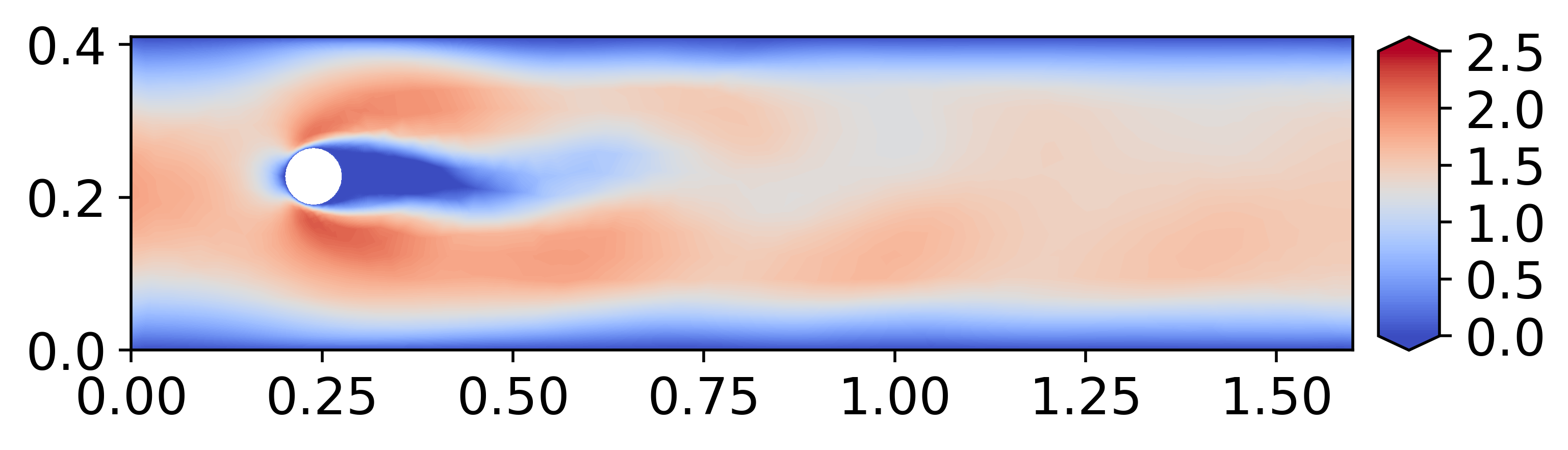}
        \caption{Prediction after 50 rollouts}
    \end{subfigure}
    \hfill
    \begin{subfigure}[h]{0.3\textwidth}
        \centering
        \includegraphics[width=\textwidth,trim={0 0 0 0},clip]{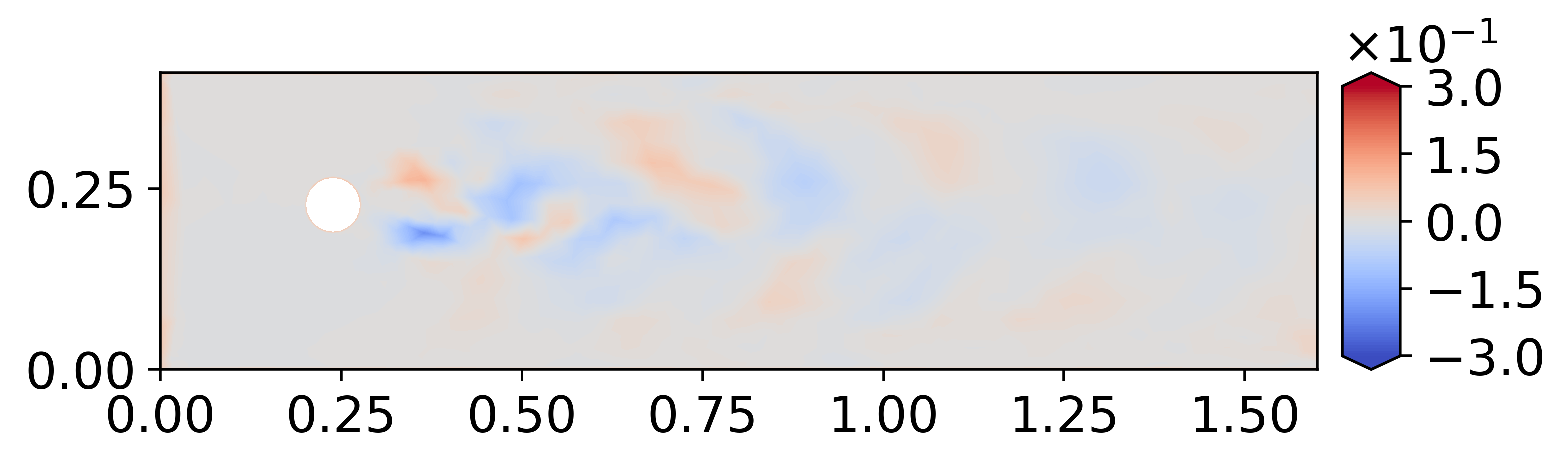}
        \caption{Error after 50 rollouts}
    \end{subfigure}

    \vfill

    \begin{subfigure}[h]{0.3\textwidth}
        \centering
        \includegraphics[width=\textwidth,trim={0 0 0 0},clip]{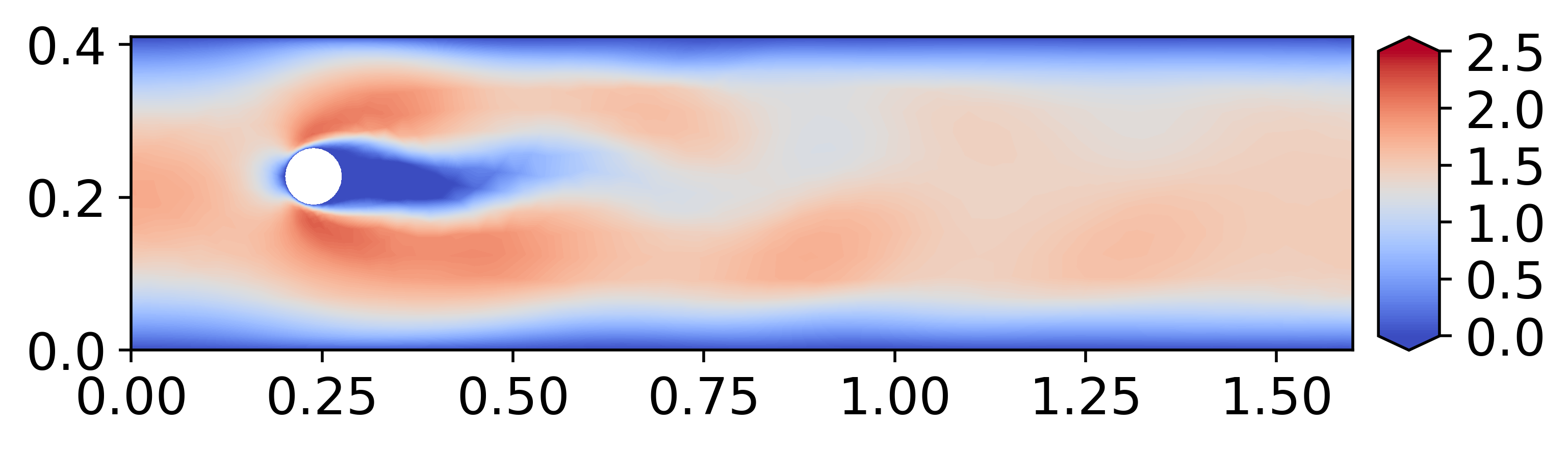}
        \caption{Ground truth after 100 rollouts}
    \end{subfigure}
    \hfill
    \begin{subfigure}[h]{0.3\textwidth}
        \centering
        \includegraphics[width=\textwidth,trim={0 0 0 0},clip]{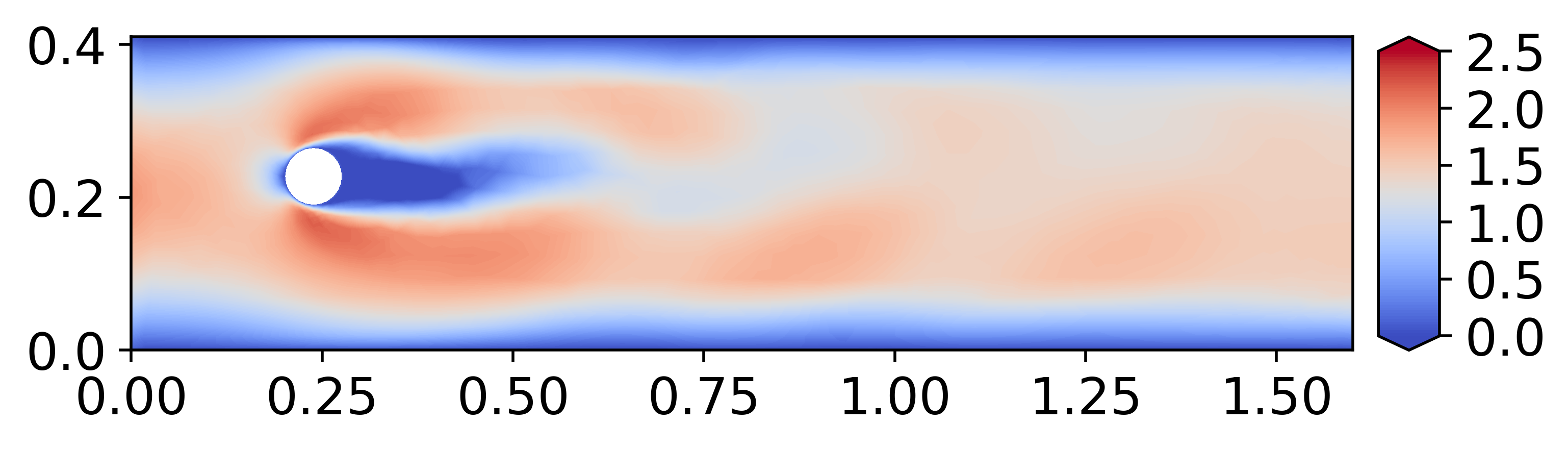}
        \caption{Prediction after 100 rollouts}
    \end{subfigure}
    \hfill
    \begin{subfigure}[h]{0.3\textwidth}
        \centering
        \includegraphics[width=\textwidth,trim={0 0 0 0},clip]{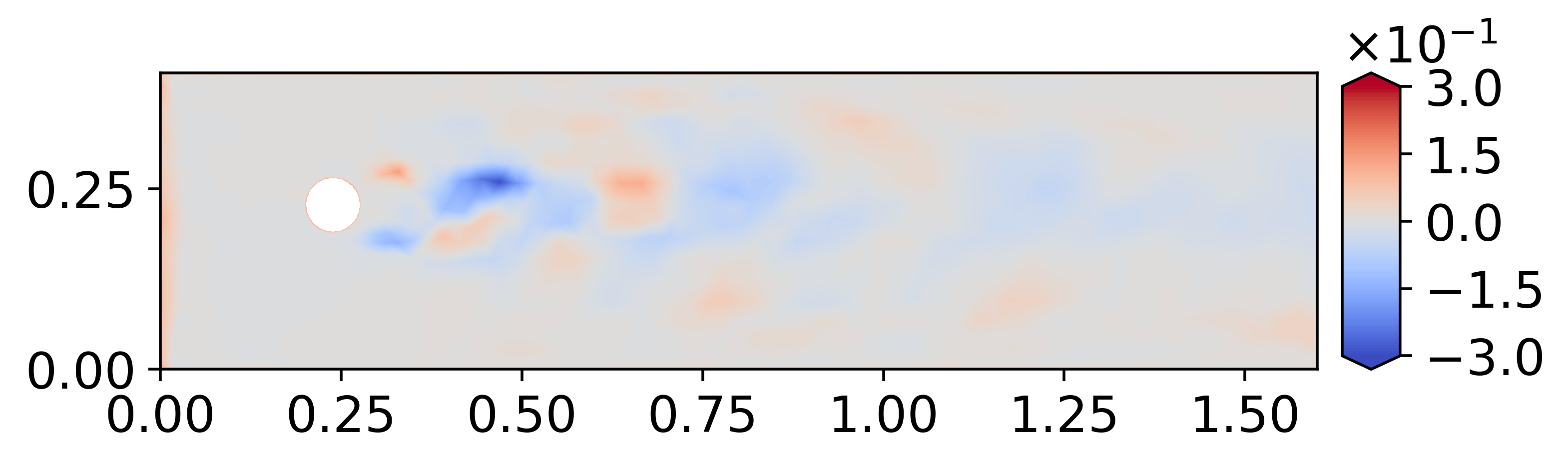}
        \caption{Error after 100 rollouts}
    \end{subfigure}

    \vfill

    \begin{subfigure}[h]{0.3\textwidth}
        \centering
        \includegraphics[width=\textwidth,trim={0 0 0 0},clip]{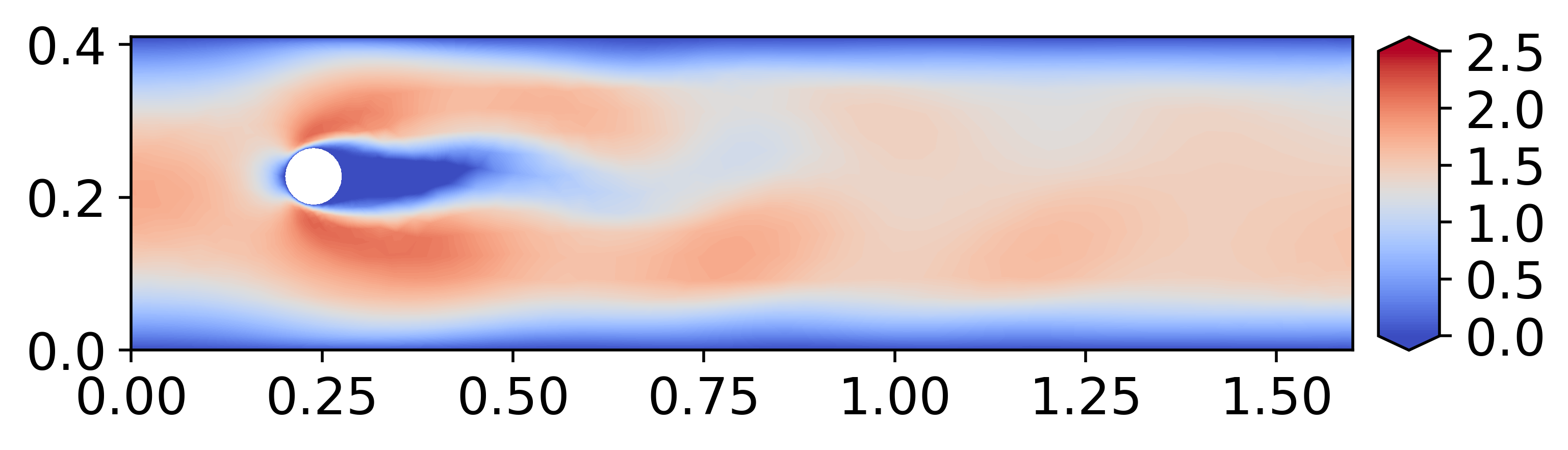}
        \caption{Ground truth after 150 rollouts}
    \end{subfigure}
    \hfill
    \begin{subfigure}[h]{0.3\textwidth}
        \centering
        \includegraphics[width=\textwidth,trim={0 0 0 0},clip]{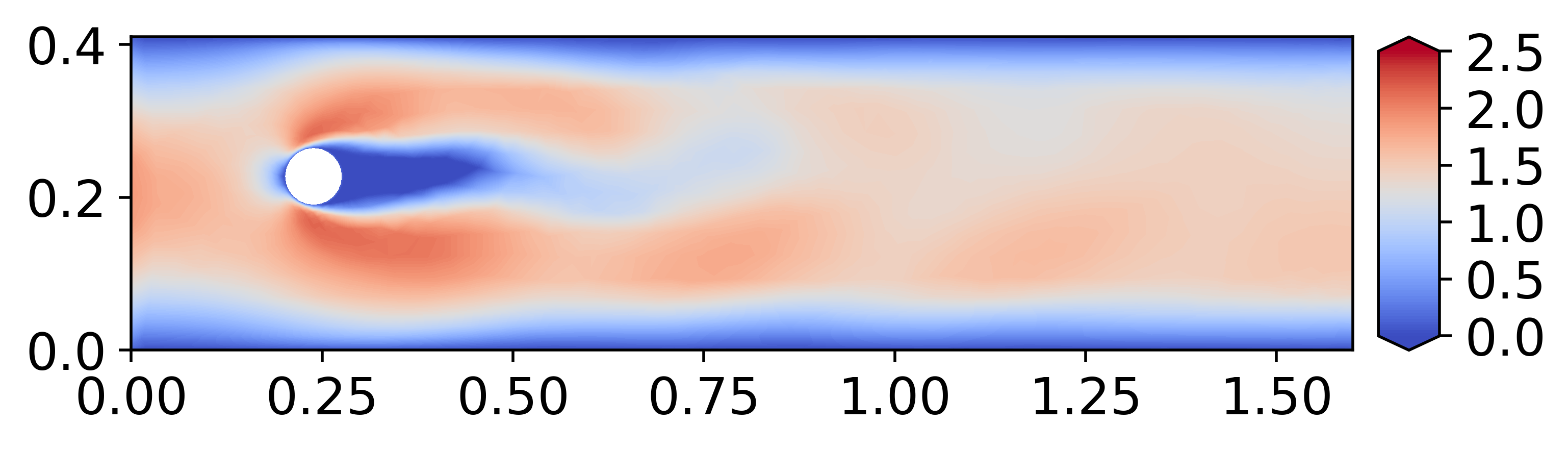}
        \caption{Prediction after 150 rollouts}
    \end{subfigure}
    \hfill
    \begin{subfigure}[h]{0.3\textwidth}
        \centering
        \includegraphics[width=\textwidth,trim={0 0 0 0},clip]{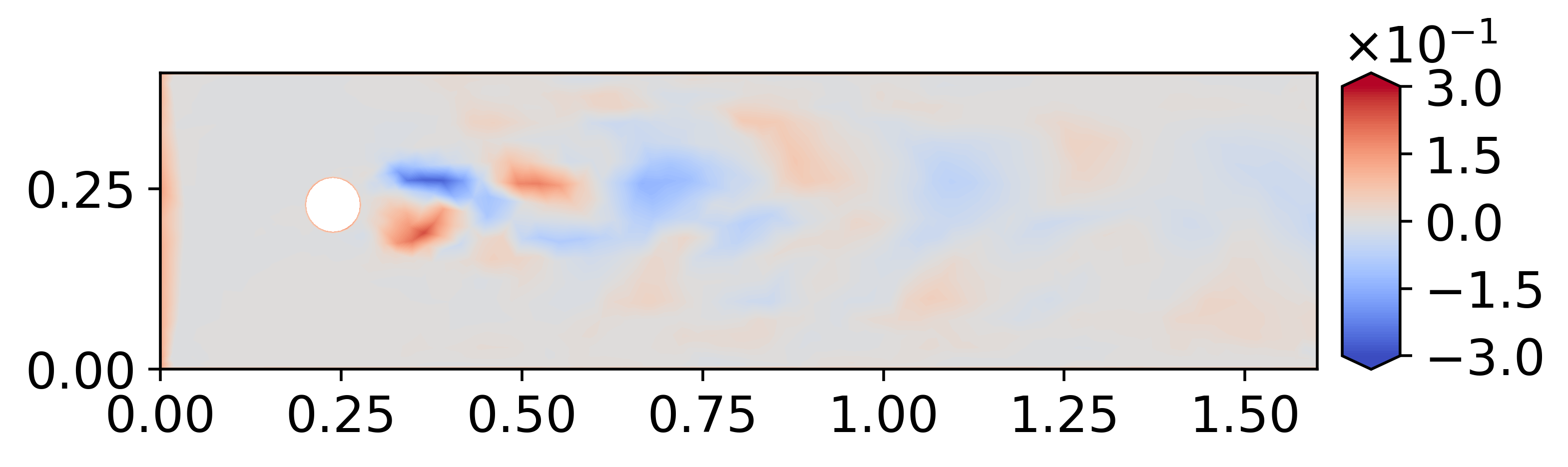}
        \caption{Error after 150 rollouts}
    \end{subfigure}

    \caption{Visualization of the $x$-velocity flow fields predicted by the model trained in Trans 2 case, where only the $x<0.75$ region is trained. From top to bottom, each row shows the results of 50, 100, and 150 rollouts after the trained snapshot range. From left to right, each column shows the ground truth, prediction, and error.}
    \label{fig:trans_ff}
\end{figure*}

The results in Trans 1 to Trans 4 cases are shown from Fig. \ref{fig:trans_zeroshot_b} to Fig. \ref{fig:trans_zeroshot_e}, respectively. As can be observed, Trans 2, 3, and 4 demonstrate reasonable results, indicating that the Graph U-Nets learn the appropriate vortex-shedding properties. A closer look at how well the velocity trend matches the ground truth shows that the accuracy of vortex shedding detection in Trans 4 is much higher than in Trans 2 and 3. On the other hand, in Trans 1, point4 totally fails to capture the periodic characteristics of the vortex shedding as the rollout progresses (after about 75 rollouts). In fact, with a domain only containing the region $x<0.5$, the Graph U-Net cannot learn the full shape of the primary vortex right behind the cylinder, whereas in Trans 2, the $x<0.75$ domain includes the whole extent of the primary vortex; thus, the results of Trans 2 show a dramatic improvement compared to Trans 1. To show a more intuitive performance of the transductive learning, the $x$-velocity flow fields of the Trans 2 case are visualized in Fig. \ref{fig:trans_ff}: snapshots of 50, 100, and 150 rollouts after the trained snapshot range are shown. Although the U-Net graph only learns the physics in the $x<0.75$ region in the Trans 2 case, it accurately captures the temporal dynamics of vortex shedding even in the 150 rollout case. Furthermore, the vortex dissipation dynamics behind $x=0.75$ are also successfully predicted, despite the fact that this region is not trained in Trans 2, highlighting the generalized predictive power of Graph U-Nets in terms of transductive learning.

From the various aspects investigated in this section, it can be concluded that the Graph U-Nets with the proposed settings in this study (with GMM operator, with the selected best pooling ratio, and with the noise injection approach) can perform successfully in transductive learning for unsteady flow-field prediction. The trained models effectively predict quantities for unseen nodes within the same Graph Used during training by leveraging the trained information obtained from the truncated graph structure. This highlights the generalizability of Graph U-Nets in capturing the dynamics of transient flow fields, even when dealing with unseen nodes in familiar meshes.

\subsection{Inductive-learning settings}
\label{sec:induct}

To evaluate the inductive-learning performance of Graph U-Nets, we consider three additional mesh scenarios, named Induct1, Induct2, and Induct 3, as shown in \ref{app:5mesh} (Fig. \ref{fig:5meshes}). These scenarios feature different inlet velocities, cylinder diameters (also the location of cylinder), and number of nodes and edges, as summarized in Table \ref{tab:induct}. Note that the baseline mesh used until this section (Fig. \ref{fig:mesh_trained}) is also presented for the comparison. Here, the Induct 1 and Induct 2 cases denote the fast-vortex shedding cases (shorter shedding periods) similar to the baseline mesh scenario, while Induct 3 is the slow-vortex shedding case compared to other cases. This diversity in mesh characteristics allows for a comprehensive evaluation of the model's generalization capabilities, especially in terms of inductive learning. It is important to emphasize that the different meshes in each scenario are due to different physics --- to model flow fields with different flow conditions and different cylinder sizes and positions, different meshes are used for each scenario type as in Fig. \ref{fig:5meshes}.

\begin{table}[htb!]
\centering
\begin{threeparttable}
\centering

\begin{tabular*}{0.6\columnwidth}{@{\extracolsep{\fill}}c|cccc}
\cline{1-5}
\multirow{2}{*}{} & \multicolumn{4}{c}{Scenario type} \\ \cline{2-5}
& Baseline & Induct 1 & Induct 2 & Induct 3 \\ \cline{1-5}
Shedding period & 29 & 28 & 28 & 40 \\ 
Inlet $x$-velocity\tnote{*} [m/s] & 1.78 & 2.21 & 2.02 & 1.68 \\
Cylinder diameter [m] & 0.074 & 0.116 & 0.089 & 0.158 \\
Number of nodes & 1,946 & 1,852 & 1,925 & 1,756 \\ 
Number of edges & 11,208 & 10,644 & 11,082 & 10,060 \\ \cline{1-5}
MSE (I-noise) & 0.85 & 3.75 & 3.53 & \textbf{39.1}  \\ 
MSE (I/O-noise) & 0.92 & 4.14 & 3.53 & \textbf{44.28} \\ \cline{1-5}
\end{tabular*}

\begin{tablenotes}
        \footnotesize
        \item[*] Since the $x$-velocity profile of the inlet is assumed to be parabolic, only the midpoint of the profile, which is the maximum value of the inlet profile, is presented.
\end{tablenotes}

\caption{Characteristics of the new mesh scenarios used to evaluate the inductive-learning performance (including the baseline scenario used in previous experiments). Also, MSE ($\times10^{-3}$) over 200 rollout zero-shot predictions using the Graph U-Net trained only with the baseline mesh is presented: there is a clear trend that similar vortex-shedding cases (Induct 1 and Induct 2) have much lower accuracy than the Induct 3 case.}
\label{tab:induct}

\end{threeparttable}
\end{table}

\subsubsection{Zero-shot prediction}
\label{sec:zero}

To investigate the inductive-learning performance of Graph U-Nets, we first attempt to analyze the zero-shot prediction performance of the Graph U-Net trained only with the baseline mesh scenario: zero-shot prediction here means that the rollout inference is performed from the past snapshots of the Induct 1$\sim$3 mesh scenarios that were never seen during training. The results in the lower part of Table \ref{tab:induct} reveal a clear trend: both the I-noise and I/O-noise models demonstrate reasonable accuracy for the fast-vortex shedding cases (Induct 1 and Induct 2) but exhibit significantly degraded performance for the slow-vortex shedding case (Induct 3). This discrepancy can be attributed to the fact that the baseline mesh used for training exhibits fast vortex-shedding characteristics. As a result, the trained models have primarily learned to predict fast-shedding dynamics, leading to inaccurate predictions when applied to mesh scenario with slower vortex shedding, such as Induct 3.

\begin{figure*}[htb!]
    \centering

    \begin{subfigure}[b]{0.35\textwidth}
        \centering
        \includegraphics[width=\textwidth,trim={0 0 0 7cm},clip]{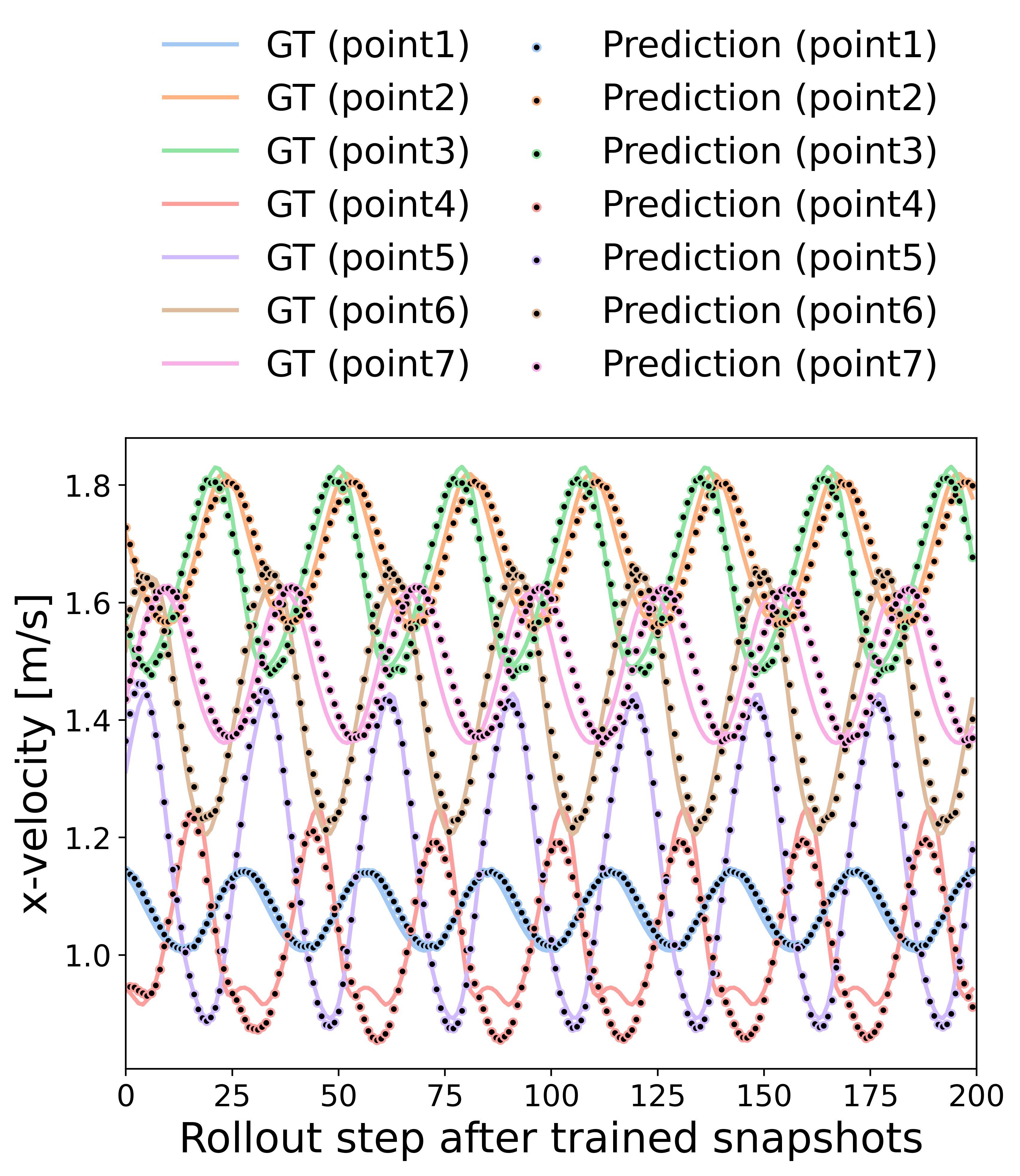}
        \caption{Baseline case (trained)}\label{fig:induct_zeroshot_a}
    \end{subfigure}
    \hspace{0.05\textwidth}
    \begin{subfigure}[b]{0.35\textwidth}
        \centering
        \includegraphics[width=\textwidth,trim={0 0 0 7cm},clip]{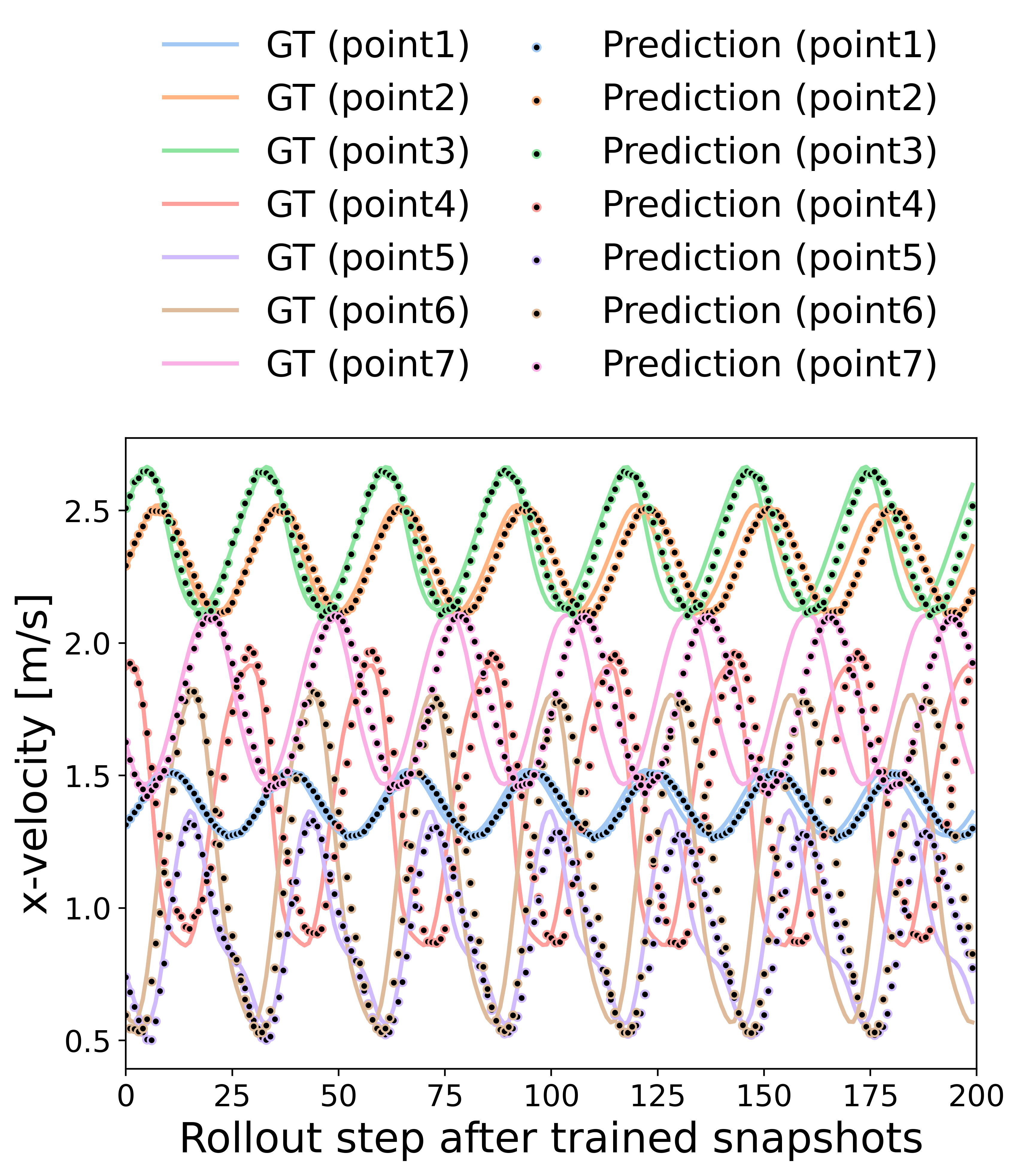}
        \caption{Induct 1 case (untrained, zero-shot)}\label{fig:induct_zeroshot_b}
    \end{subfigure}
    
    \vfill
    
    \begin{subfigure}[b]{0.35\textwidth}
        \centering
        \includegraphics[width=\textwidth,trim={0 0 0 7cm},clip]{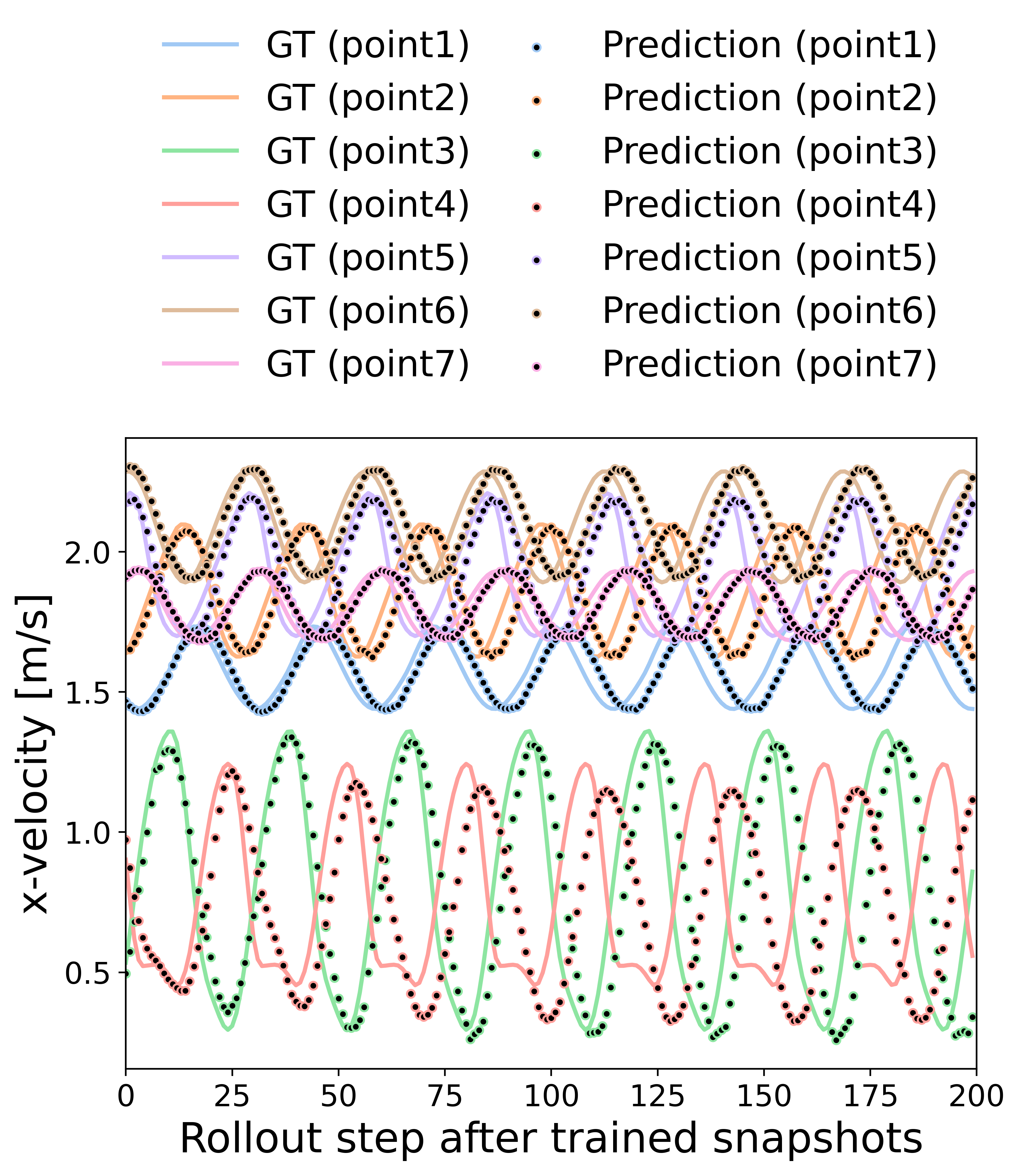}
        \caption{Induct 2 case (untrained, zero-shot)}\label{fig:induct_zeroshot_c}
    \end{subfigure}
    \hspace{0.05\textwidth}
    \begin{subfigure}[b]{0.35\textwidth}
        \centering
        \includegraphics[width=\textwidth,trim={0 0 0 7cm},clip]{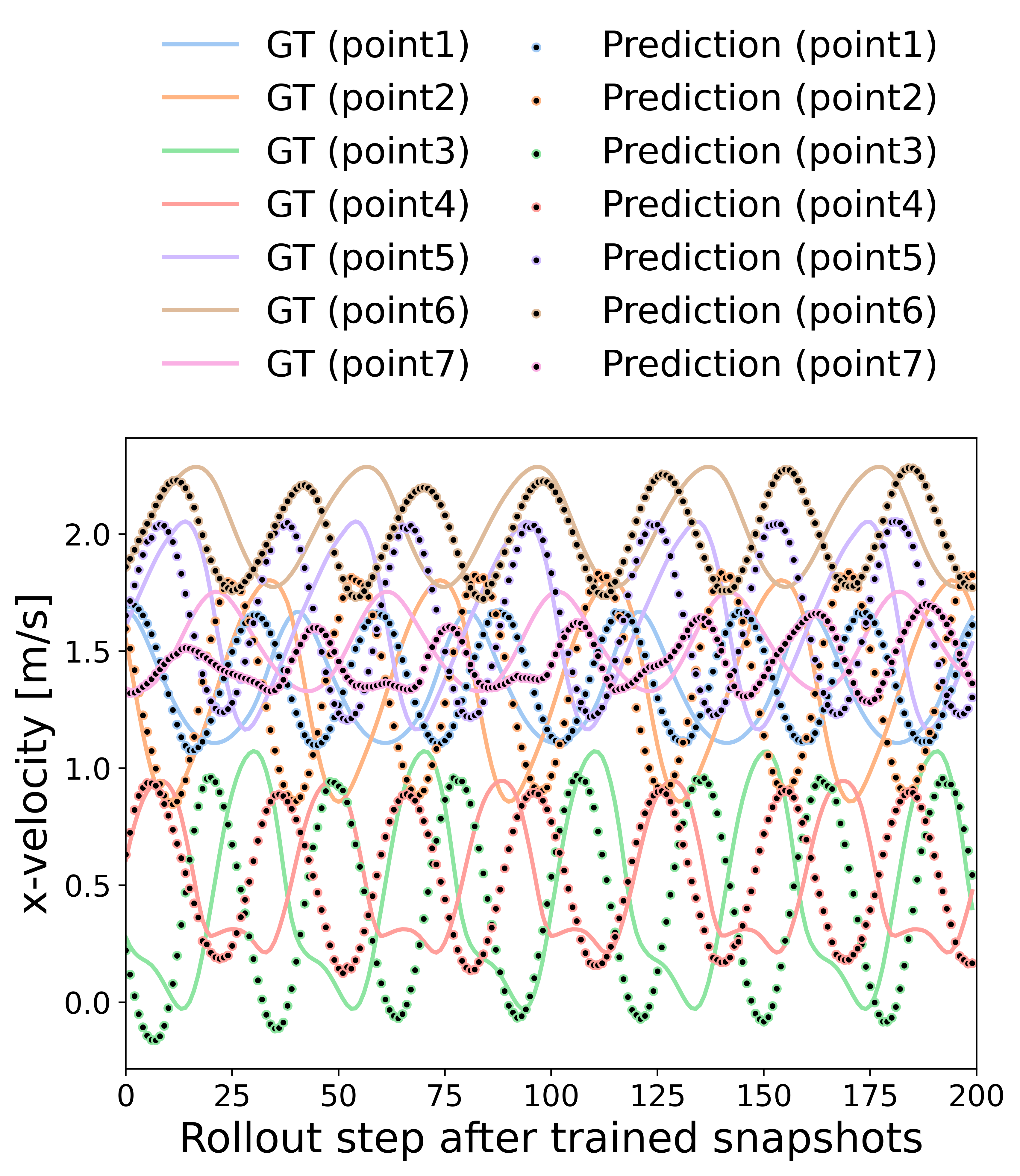}
        \caption{Induct 3 case (untrained, zero-shot)}\label{fig:induct_zeroshot_d}
    \end{subfigure}
    

    \caption{Comparison of the predicted $x$-velocity at seven points (Fig. \ref{fig:trans_zeroshot_a}) when only the baseline mesh scenario is used for the training: (a) shows the prediction results of the trained baseline case while (b), (c), and (d) show the zero-shot results of the three inductive cases tested in this section. Induct 3 case shows significantly different periodic $x$-velocities compared with the other untrained cases, Induct 1 and Induct 2.}
    \label{fig:induct_zeroshot}
\end{figure*}

To visually examine the long-term zero-shot rollout performance, Fig. \ref{fig:induct_zeroshot} compares the predicted $x$-velocity at seven points behind the cylinder during the prediction of 200 future snapshots using the model trained only with the baseline mesh scenario (and also with the I-noise approach). As expected, the model shows the best performance in predicting the future snapshots of the baseline mesh (Fig. \ref{fig:induct_zeroshot_a}), which was used for the training. For the fast vortex shedding cases (Induct 1 and Induct 2), the model also demonstrates reasonable performance, as shown in Figs. \ref{fig:induct_zeroshot_b} and \ref{fig:induct_zeroshot_c}. Although there is a tendency to predict slower shedding periods compared to the ground truth as the rollout progresses, the overall accuracies remain satisfactory. However, the model fails to accurately capture the phase of the vortex shedding in the slower-shedding case (Induct 3), as evident in Fig. \ref{fig:induct_zeroshot_d}. The clear trend of the predicted fluctuations exhibiting shorter periods than the ground truth indicates that the model has only learned fast vortex shedding dynamics of the baseline mesh scenario during training. As a result, it incorrectly applies this learned fast-shedding tendency even when inferring the slow-shedding case, Induct 3.

The results in this section demonstrate the Graph U-Net's ability to generalize to unseen meshes with similar physical characteristics, while also revealing its limitations when faced with different dynamics. Its inductive-learning performance is significantly influenced by the characteristics of the trained mesh scenarios, emphasizing the importance of training on mesh scenarios with varying flow characteristics to improve robustness and generalization capabilities, and to effectively handle novel scenarios.

\subsubsection{Towards inductive learning}
\label{sec:induct_train}

To overcome the limitations of the Graph U-Net models in generalizing to unseen meshes with different flow characteristics, we propose to use a simple but effective approach: training the model on a combination of meshes exhibiting both fast and slow vortex-shedding phenomena. By exposing the model to a wider range of flow dynamics during training, we aim to improve its inductive-learning performance and thus enable accurate predictions on mesh scenarios with different shedding characteristics.

In this study we accomplish this by adopting only two mesh scenarios for training: the previous mesh with fast-vortex shedding characteristics (baseline mesh used for training throughout this study) and the Induct 3 mesh, which exhibits slow vortex shedding behavior. By incorporating the Induct 3 mesh into the training set, we provide the model with explicit examples of slower vortex shedding dynamics, allowing it to learn and capture the intricacies of both fast and slow shedding phenomena. And note that since the training data is twice as large as before, the training time is also twice as long as the single-mesh training scenario. For the inductive-learning setting, the overall architecture of the Graph U-Net remains the same as in previous sections. However, the number of snapshots used as input for predicting the very next snapshot increases to 40 (compared to 20 in previous experiments), and the number of kernels increases to 3 (compared to 1 in previous experiments). Since the input channel increases to 40, each channel in the encoder's GMM convolutional operator changes to 40, 20, 10, 5, and 1 (which was 20, 15, 10, 5, and 1 in previous experiments). To maintain consistency with the best-performing model in Section \ref{sec:noise}, we employ both I-noise and I/O-noise injection approaches with a noise size of 0.16 --- to demonstrate the benefits of noise injection techniques in the inductive setting, experiment without noise injection is also carried out for the comparison. Finally, the results of the inductive-learning setting are summarized in Table \ref{tab:induct_summary}: various pooling ratios are also explored, including the case without pooling/unpooling. Considering that the results from Table \ref{tab:induct} indicate that the Induct 3 case was predicted with MSE of $39.1\times10^{-3}$ in I-noise approach, the MSE of $1.78\times10^{-3}$ in I-noise case without pooling in Table \ref{tab:induct_summary} seems remarkable. However, without noise, the lowest MSE for the Induct 3 case is $18.34\times10^{-3}$ (when pooling is not applied), again highlighting the effectiveness of noise injection techniques on the robustness of rollout performance. For the additional experiments that investigate the effects of the number of input snapshots and the number of GMM kernels, see \ref{app:induct_hyper}.

\renewcommand{\arraystretch}{1.15}
\begin{table}[htb!]
\centering
\begin{threeparttable}
\centering

\begin{tabular}{c|c|cccc}
\cline{1-6} 
\multirow{2}{*}{\shortstack{Noise\\type}} &  \multirow{2}{*}{Pooling Ratio} & \multicolumn{4}{c}{Scenario type} \\ \cline{3-6} 
 & & Baseline\tnote{*} & Induct 1 & Induct 2 & Induct 3\tnote{*} \\ \cline{1-6} 
\multirow{5}{*}{I-noise}     & 0.2 & 110.49 & 128.49 & 112.89 & 142.33 \\
                             & 0.4 & 95.40 & 78.38 & 101.10 & 71.52 \\
                             & 0.6 & 32.40 & 26.65 & 32.95 & 91.34 \\
                             & 0.8 & \textbf{2.74} & 7.41 & 6.33 & 41.05 \\ \cline{2-6} 
                             & Without pooling & 4.97 & \textbf{6.32} & \textbf{6.30} & \textbf{1.78} \\ \cline{1-6} 
\multirow{5}{*}{I/O-noise}     & 0.2 & 103.68 & 117.97 & 114.30 & 142.94 \\
                             & 0.4 & 113.08 & 73.84 & 74.20 & 80.97 \\
                             & 0.6 & 4.68 & 17.33 & 12.76 & 94.16 \\
                             & 0.8 & \textbf{2.60} & \textbf{5.44} & \textbf{5.37} & 14.67 \\ \cline{2-6} 
                             & Without pooling & 5.32 & 6.31 & 6.30 & \textbf{1.86} \\ \cline{1-6} 
\multirow{5}{*}{\shortstack{Without\\noise}}   & 0.2 & 73.60 & 99.93 & 90.90 & 96.04 \\
                             & 0.4 & 70.83 & 85.62 & 89.04 & 146.64 \\
                             & 0.6 & \textbf{22.99} & \textbf{46.17} & \textbf{39.52} & 88.26 \\
                             & 0.8 & 66.48 & 82.24 & 106.76 & 149.36 \\ \cline{2-6} 
                             & Without pooling & 44.25 & 127.34 & 97.45 & \textbf{18.34} \\ \cline{1-6} 
\end{tabular}

\begin{tablenotes}
        \footnotesize
        \item[*] Both Baseline and Induct 3 graphs are trained for this inductive-learning setting.
\end{tablenotes}

\caption{MSE ($\times10^{-3}$) over 200 rollout predictions in the inductive-learning setting is presented: different noise injection approaches (including without noise injection) and pooling ratios (including without pooling operation) are examined. Note that the Induct 1 and Induct 2 cases are zero-shot predictions.}
\label{tab:induct_summary}

\end{threeparttable}
\end{table}

The most interesting point in the results in Table \ref{tab:induct_summary} is that for inductive learning discussed in this section, where more than one mesh scenario is trained, pooling no longer improves the performance of Graph U-Nets. Rather, when focusing on the MSE of the Induct 3 in Table \ref{tab:induct_summary}, the model without pooling always exhibits dramatically better performance than the models with pooling. For example, in the I-noise cases, the MSE of the Induct 3 without pooling is $1.78\times10^{-3}$, while the MSE with pooling at a pooling ratio of 0.6 is $91.34\times10^{-3}$. Furthermore, not using pooling is also beneficial in terms of training time (as already noted in Table \ref{tab:pool_ratio}): the model with a pooling ratio of 0.6 requires 7,007 seconds, while the model without pooling requires only 5,581 seconds (NVIDIA 3080 GPU is used). Considering that the best pooling ratio was found to be 0.6 in Table \ref{tab:pool_ratio} when only the single-mesh scenario was used for training, these conflicting results in inductive learning where two different mesh scenarios are trained are noteworthy. The reasons for this can be attributed to the following factors:

\begin{enumerate}
    \item Variability in graph structures: different graphs have different structures, making it challenging for pooling operations to consistently coarsen or abstract the graphs in a way that preserves relevant features across all of them.
    \item Improved generalization and flexibility: Graph U-Nets without pooling can better adapt to the unique characteristics of each graph because they learn individually from the raw, fully detailed structure of the graphs. By preserving every detail at every GMM block, they can more accurately capture the complex physics and varying dynamics of different mesh scenarios. This flexibility and ability to learn from detailed graph structures supports generalization across different graphs with varying physical properties, making Graph U-Nets without pooling more suitable for inductive-learning scenarios.
\end{enumerate}

\begin{figure*}[htb!]
    \centering

    \begin{subfigure}[b]{0.35\textwidth}
        \centering
        \includegraphics[width=\textwidth,trim={0 0 0 7cm},clip]{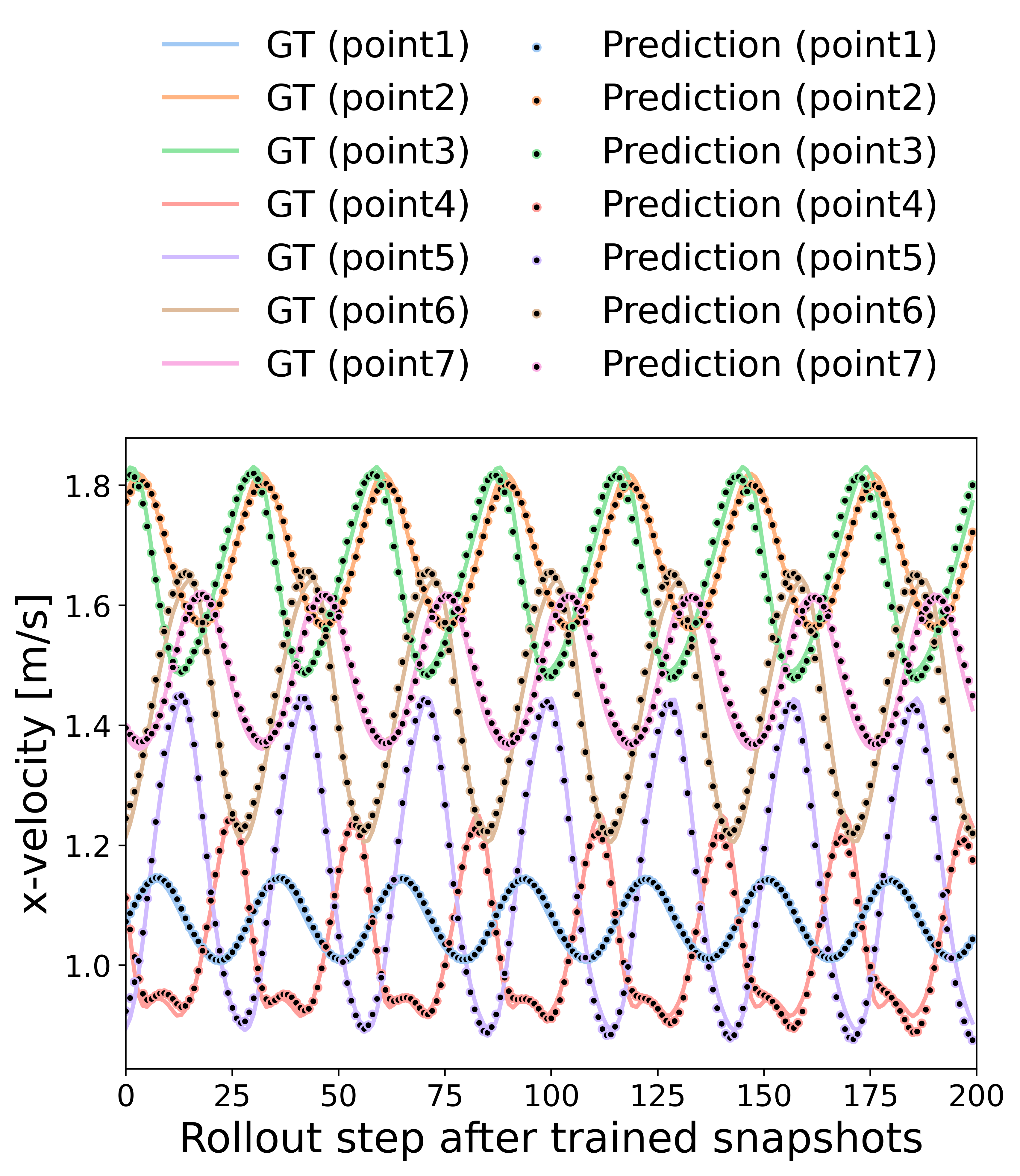}
        \caption{Baseline case (trained)}\label{fig:induct_1}
    \end{subfigure}
    \hspace{0.05\textwidth}
    \begin{subfigure}[b]{0.35\textwidth}
        \centering
        \includegraphics[width=\textwidth,trim={0 0 0 7cm},clip]{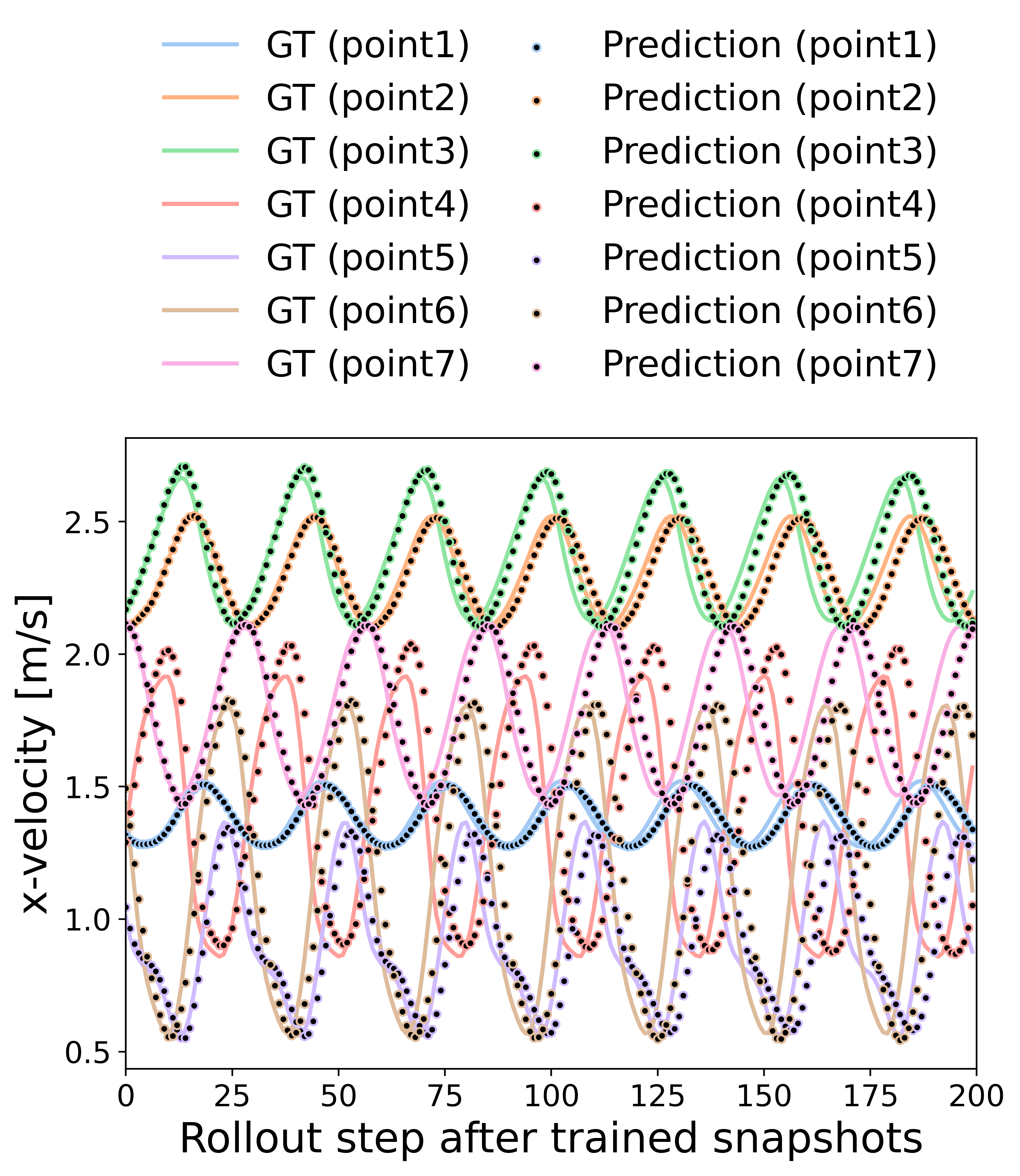}
        \caption{Induct 1 case (untrained, zero-shot)}\label{fig:induct_2}
    \end{subfigure}
    
    \vfill
    
    \begin{subfigure}[b]{0.35\textwidth}
        \centering
        \includegraphics[width=\textwidth,trim={0 0 0 7cm},clip]{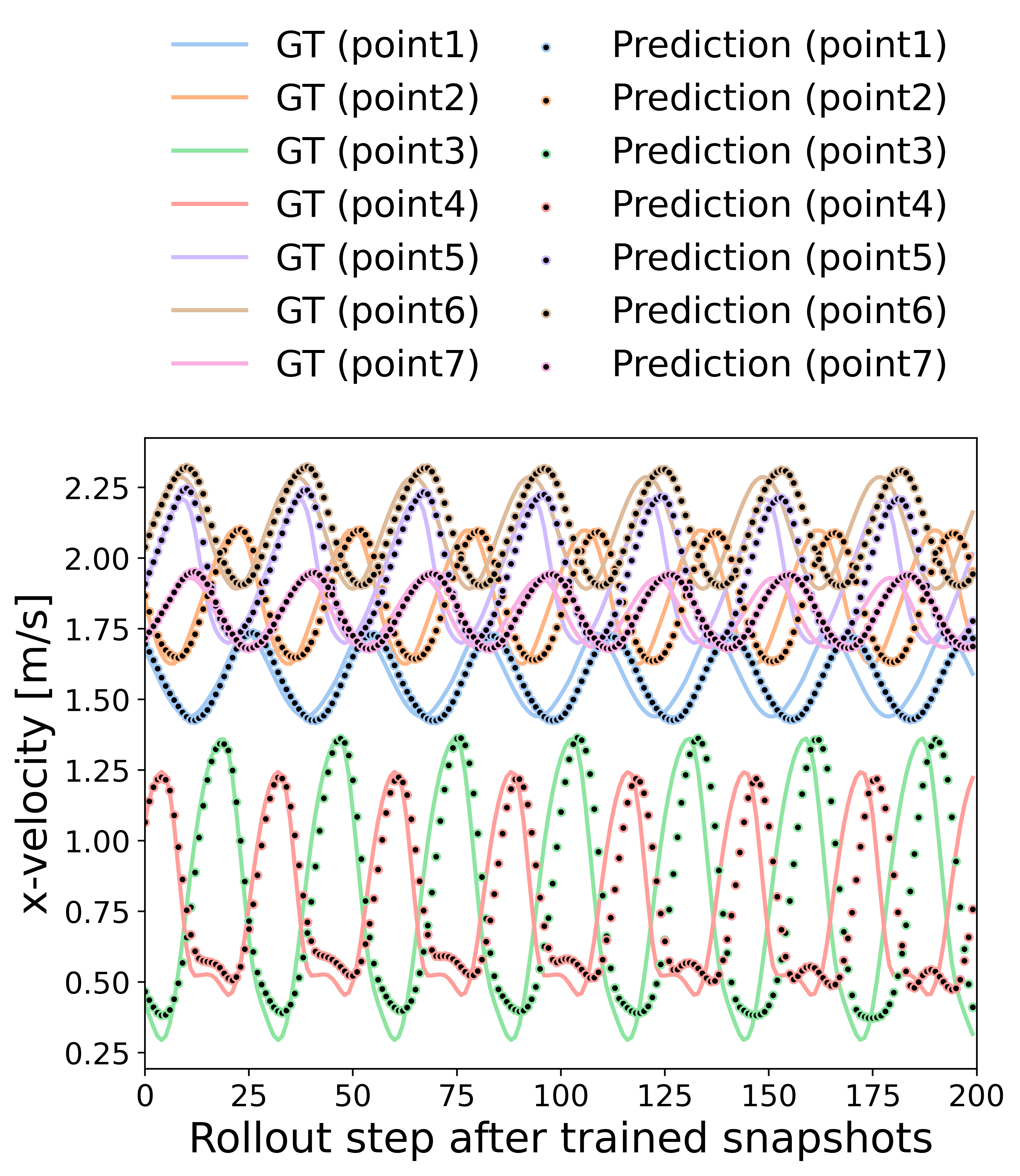}
        \caption{Induct 2 case (untrained, zero-shot)}\label{fig:induct_3}
    \end{subfigure}
    \hspace{0.05\textwidth}
    \begin{subfigure}[b]{0.35\textwidth}
        \centering
        \includegraphics[width=\textwidth,trim={0 0 0 7cm},clip]{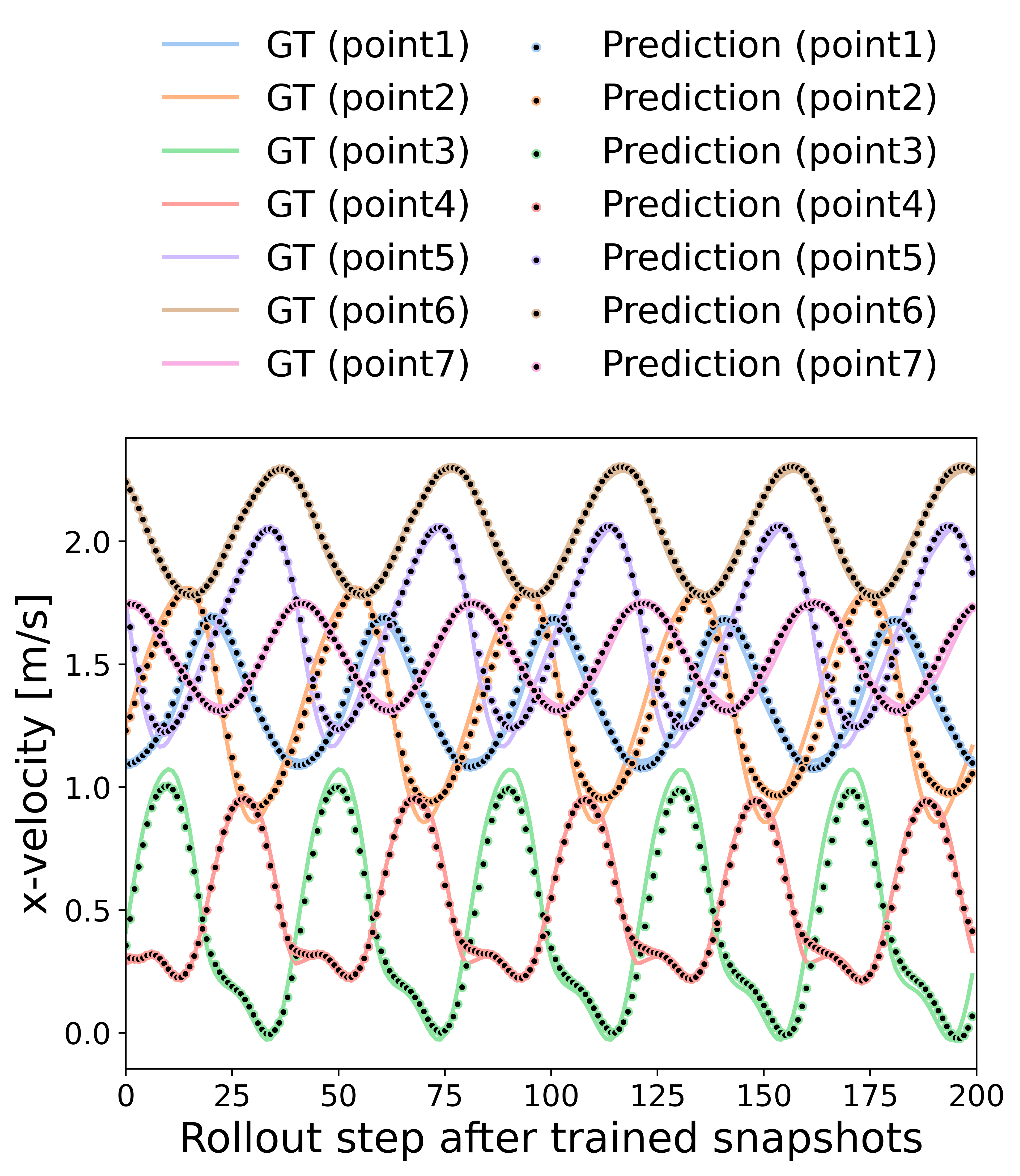}
        \caption{Induct 3 case (trained)}\label{fig:induct_4}
    \end{subfigure}
    

    \caption{Comparison of the predicted $x$-velocity at seven points (Fig. \ref{fig:trans_zeroshot_a}) when both the baseline and Induct 3 graphs are used for inductive learning. The results are from the model with I-noise and without pooling: (a) and (d) show the results of the trained case while (b) and (c) show the zero-shot results of the two inductive cases tested in this section.}
    \label{fig:induct}
\end{figure*}

\begin{figure*}[htb!]
    \centering
    
    \begin{subfigure}[h]{0.48\textwidth}
        \centering
        \includegraphics[width=\textwidth]{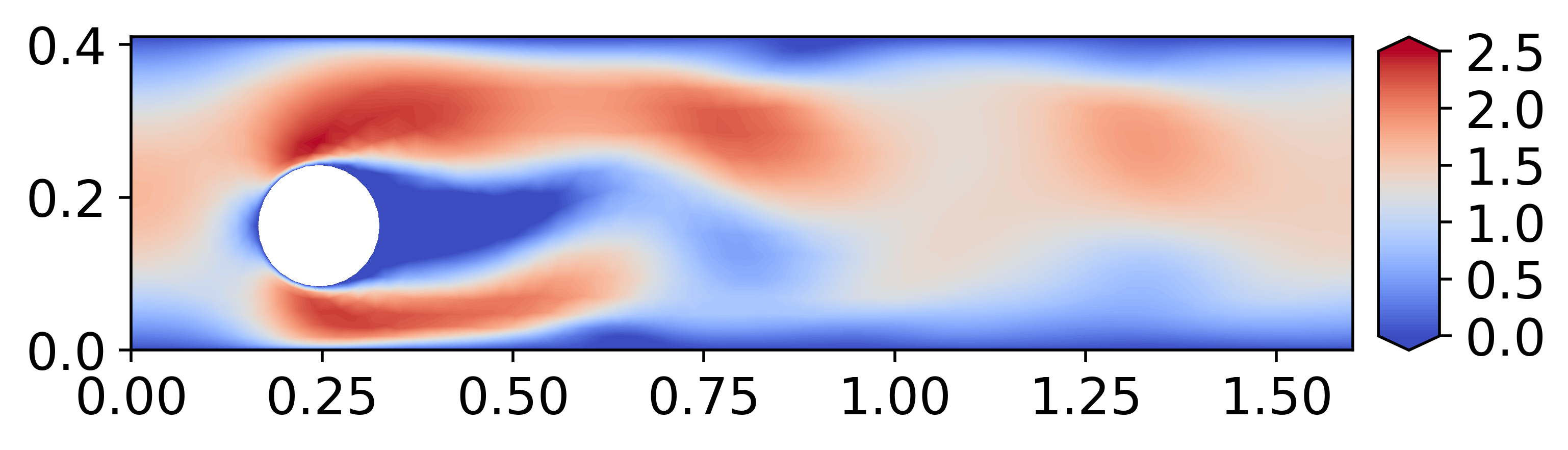}
        \caption{Ground truth}\label{fig:induct_ff_a}
    \end{subfigure}

    \vfill
    
    \begin{subfigure}[h]{0.48\textwidth}
        \centering
        \includegraphics[width=\textwidth]{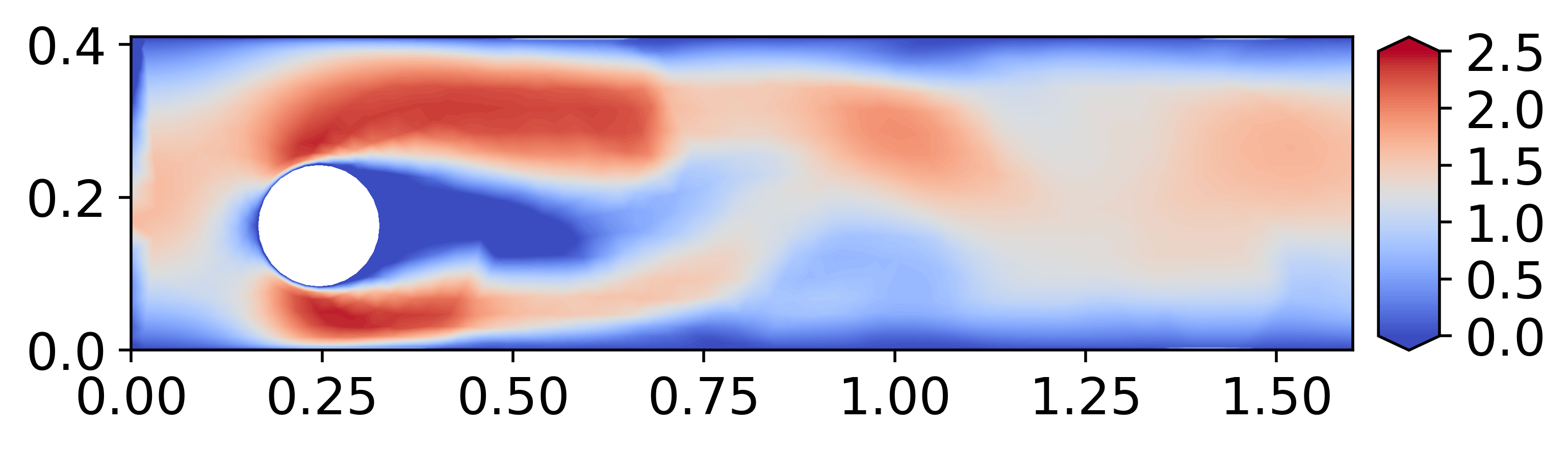}
        \caption{Prediction from the model without inductive learning}\label{fig:induct_ff_b}
    \end{subfigure}
    \hfill
    \begin{subfigure}[h]{0.48\textwidth}
        \centering
        \includegraphics[width=\textwidth]{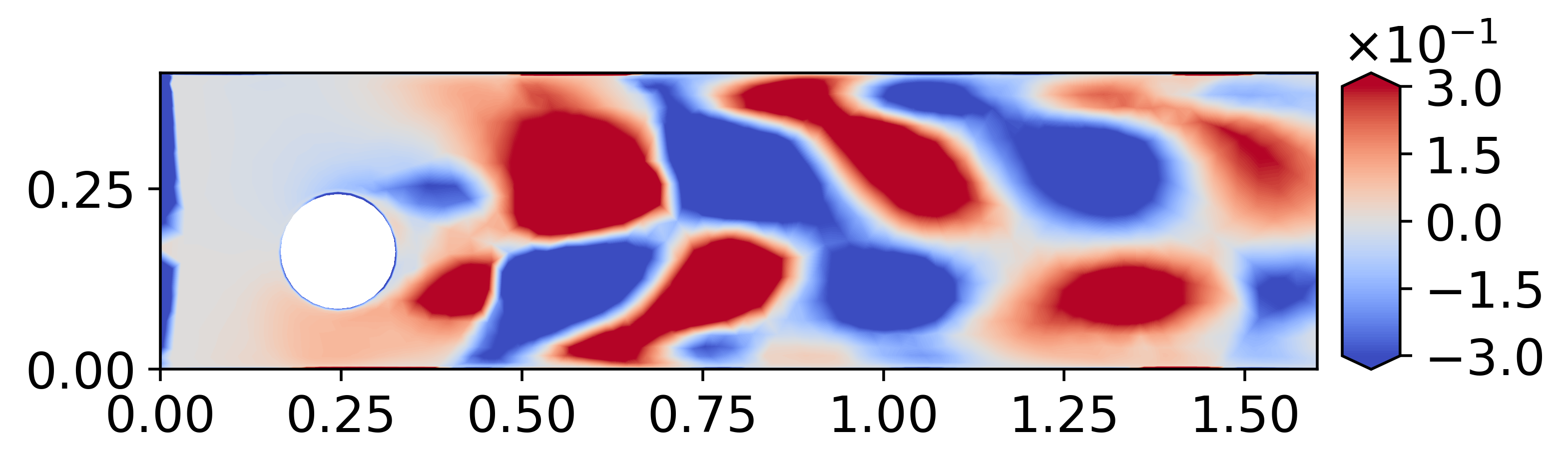}
        \caption{Error from the model without inductive learning}\label{fig:induct_ff_c}
    \end{subfigure}

    \vfill
    
    \begin{subfigure}[h]{0.48\textwidth}
        \centering
        \includegraphics[width=\textwidth]{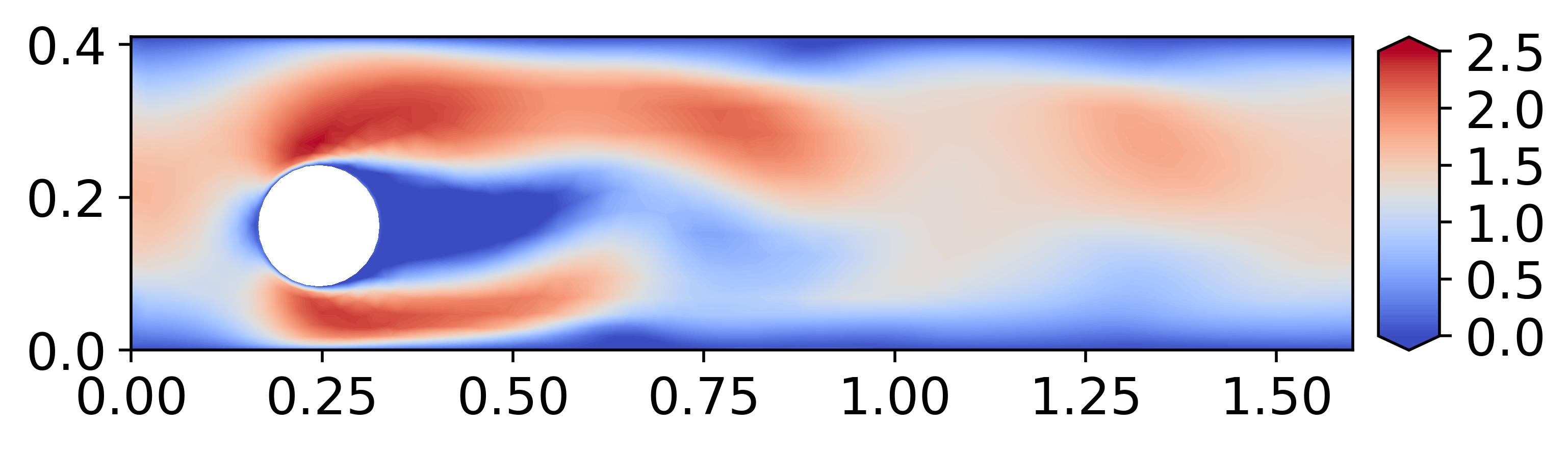}
        \caption{Prediction from the model with inductive learning}\label{fig:induct_ff_d}
    \end{subfigure}
    \hfill
    \begin{subfigure}[h]{0.48\textwidth}
        \centering
        \includegraphics[width=\textwidth]{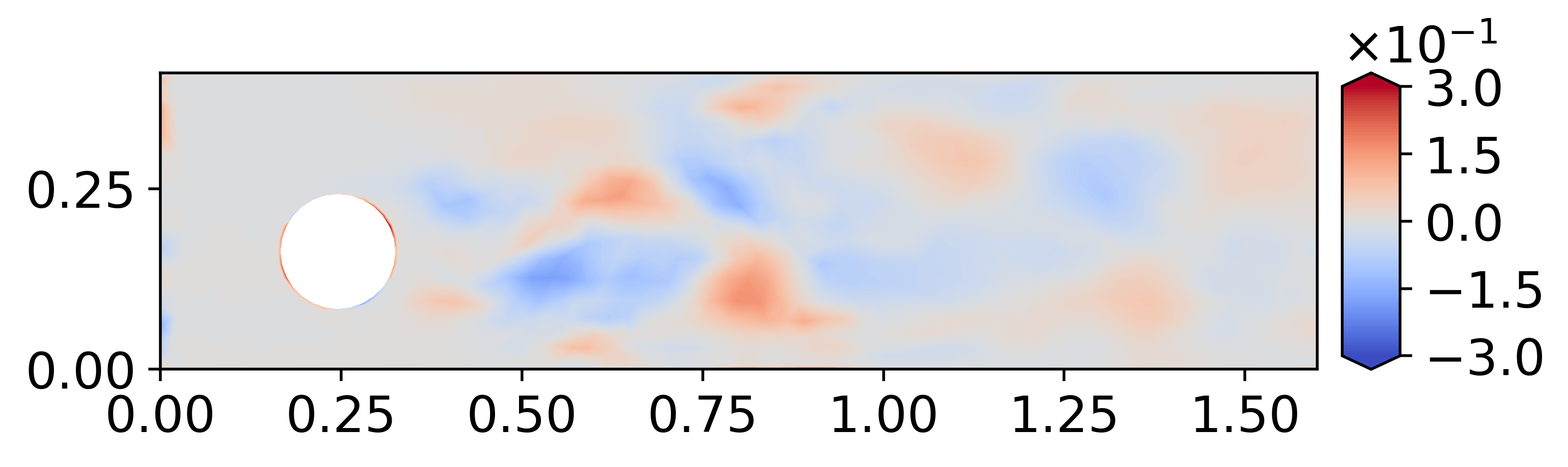}
        \caption{Error from the model with inductive learning}\label{fig:induct_ff_e}
    \end{subfigure}
    

    \caption{For two models (without inductive setting and with inductive setting), $x$-velocity flow fields of 150 rollouts after the trained snapshot range in Induct 3 graph are compared. Both models are trained with I-noise (noise size of 0.16) and without pooling: Figs. \ref{fig:induct_ff_b} and \ref{fig:induct_ff_c} show the results without inductive learning and Figs. \ref{fig:induct_ff_d} and \ref{fig:induct_ff_e} show the results with inductive learning.}
    \label{fig:induct_ff}
\end{figure*}

Then, Fig. \ref{fig:induct} shows the velocity profiles at seven points behind the cylinder. In Figs. \ref{fig:induct_1} and \ref{fig:induct_4}, the baseline mesh scenario and Induct 3 case are shown, respectively. Since they are used for model training, they show satisfactory accuracy. Figs. \ref{fig:induct_2} and \ref{fig:induct_3} show the Induct 1 and Induct 2 cases, respectively, and still demonstrate reasonable accuracy, highlighting that the embedding of the Induct 3 case for inductive learning does not decrease the inductive performance with respect to the Induct 1 and Induct 2 cases, while achieving the capability of the Graph U-Net to train the slow vortex shedding case as in Induct 3 results. In Fig. \ref{fig:induct_ff}, the predicted $x$-velocity flow fields for Induct 3 case are also compared between models without inductive learning (only baseline graph is trained) and with inductive learning (both baseline and Induct 3 graphs are trained). Fig. \ref{fig:induct_ff_b} (model without inductive learning) shows a flow field that is very different from the ground truth in Fig. \ref{fig:induct_ff_a}: the flow field is quite discontinuous and the phase of the vortex shedding does not match the ground truth. However, Fig. \ref{fig:induct_ff_d} (model with inductive learning) shows a flow field similar to that of the ground truth, although the predicted flow field is the result of 150 rollouts after the training-snapshot range. When the MSE over 200 rollouts is computed from these two models, the model without inductive learning shows an MSE of $73.76\times10^{-3}$ in the Induct 3 case, while the model with inductive learning shows $1.78\times10^{-3}$ --- i.e., improvement of 98\% can be obtained when inductive learning is adopted for the Induct 3 scenario.

It might be argued that successfully predicting slow-vortex shedding by including a slow vortex-shedding mesh scenario in the training data is an obvious outcome. However, it is crucial to emphasize that this study focuses on predicting future snapshots by learning solely from past snapshots. The truly remarkable finding in this section is that a single Graph U-Net model, when trained simultaneously with mesh scenarios exhibiting different physical properties, can effectively learn and adapt to both properties concurrently. This means that during the inference phase, even when presented with input snapshots corresponding to different vortex-shedding speeds, the model can generate appropriate predictions specific to each shedding speed, rather than merely outputting an average of the trained speeds. Graph U-Nets demonstrate this capability to accurately predict future snapshots of both fast and slow vortex-shedding cases (when pooling is not applied) by simply training the model with past snapshots of the two vortex-shedding scenarios. While this can be considered an interpolation in terms of vortex-shedding speed, it also showcases the model's ability to extrapolate successfully in the time domain, given that only past snapshots are used for training. This highlights the robustness and adaptability of Graph U-Nets in capturing and predicting complex flow dynamics across different scales and physical characteristics.

\subsubsection{Effects of normalization}
\label{sec:induct_norm}

As noted in Section \ref{sec:LN}, all previously trained Graph U-Net models are trained with the same normalization technique, LN. In this section, we investigate the impact of the following different normalization techniques on the inductive-learning performance of Graph U-Nets: without normalization, with LN (layer normalization - Section \ref{sec:LN}), and with GN (graph normalization - Section \ref{sec:GN}). Note again that LN normalizes node features across the channel dimension (in this study, snapshot dimension) for each individual node, while GN normalizes node features across the node dimension for each feature. To evaluate the effects of normalization, we utilize the models without pooling. Again, to investigate the effects of noise injection, we compare I-noise and I/O-noise approaches, and also compare the three normalization approaches to train both baseline and Induct 3 cases. The results are summarized in Table \ref{tab:induct_norm}, which presents the MSE over 200 rollout predictions for each combination of noise type and normalization approach.

\renewcommand{\arraystretch}{1.15}
\begin{table}[htb!]
\centering
\begin{threeparttable}
\centering

\begin{tabular}{c|c|cccc}
\cline{1-6} 
\multirow{2}{*}{Description} &  \multirow{2}{*}{Norm type} & \multicolumn{4}{c}{Scenario type} \\ \cline{3-6} 
 & & Baseline\tnote{*} & Induct 1 & Induct 2 & Induct 3\tnote{*} \\ \cline{1-6} 
\multirow{3}{*}{\shortstack{I-noise\\\&\\Without pooling}}     & Without & 4.24 & 9.47 & 7.73 & 5.95 \\
                             & LN & 4.97 & \textbf{6.32} & 6.30 & \textbf{1.78} \\
                             & GN & \textbf{1.82} & 6.55 & \textbf{6.02} & 8.88 \\ \cline{1-6} 
\multirow{3}{*}{\shortstack{I/O-noise\\\&\\Without pooling}}     & Without & 4.77 & 9.73 & 8.07 & 5.62 \\
                             & LN & 5.32 & \textbf{6.31} & \textbf{6.30} & \textbf{1.86} \\
                             & GN & \textbf{2.20} & 6.76 & 6.35 & 13.24 \\ \cline{1-6} 
\end{tabular}

\begin{tablenotes}
        \footnotesize
        \item[*] Both Baseline and Induct 3 graphs are trained for this inductive-learning setting.
\end{tablenotes}

\caption{MSE ($\times10^{-3}$) over 200 rollout predictions is presented with varying normalization types in inductive setting: without normalization, LN (layer normalization), and GN (graph normalization). Results are shown for both I-noise and I/O-noise approaches, without pooling operation.}
\label{tab:induct_norm}

\end{threeparttable}
\end{table}

Interestingly, while GN outperforms LN for the baseline graph across both noise injection types, LN exhibits superior accuracy in the Induct 3 graph, which represents a slow vortex-shedding scenario. With the I-noise approach, the model with LN achieves an MSE of $1.78\times10^{-3}$ for the Induct 3 case, compared to $8.88\times10^{-3}$ for the model with GN, and with I/O-noise approach, LN reduces the MSE to $1.86\times10^{-3}$ for the Induct 3 case, while GN yields an MSE of $13.24\times10^{-3}$. These GN results are worse than the results without normalization, which indicates that the ability to learn different physical behaviors in inductive setting heavily depends on the selection of normalization type. Also note that LN and GN show similar performance in Induct 1 and 2 graphs, which feature fast vortex-shedding characteristics similar to the baseline graph. The results in Table \ref{tab:induct_norm} suggest that for the inductive setting, where two graphs (baseline and Induct 3) with different physical characteristics (vortex shedding period) are used for training, the LN approach is the most effective when considering the overall accuracy for all graph scenarios.

The superior performance of LN can be attributed to its ability to focus on effectively capturing the temporal dynamics by normalizing the node features ($x$-velocity in this study) across time. This ensures that the model can learn and adapt to the unique temporal patterns and dependencies present at each node in the unsteady flow field, which is particularly important in our problem where the temporal evolution of the flow fields is a critical aspect of the prediction task.

In contrast, GN normalizes node features across all nodes for each feature dimension, focusing on capturing the spatial dynamics between nodes. While GN's ability to consider the overall structure and relationships within the nodes for each feature dimension can be advantageous in certain scenarios, it may not be as crucial in our specific problem. This is because the nodes of each graph type remain constant over time (e.g., every snapshot in baseline graph type has the same mesh, which is also true for Induct 3) in our case, where the primary focus is on predicting the temporal evolution of the flow field within the same mesh. By normalizing node features across node dimensions, GN may unintentionally suppress the unique temporal patterns and variations at each individual node because it considers the global statistics of the entire graph for each feature dimension rather than focusing on individual node dynamics. 

The comparative analysis of LN and GN within the inductive-learning setting in this section highlights the importance of selecting an appropriate normalization technique based on the characteristics of the problem. In our study, where the goal is to predict the temporal evolution of the flow field using Graph U-Nets, LN proves to be superior. By normalizing features for each node independently over the channel dimension, LN effectively captures the unique temporal dynamics at each node. In contrast, GN normalizes features for each feature dimension and focuses on the spatial dynamics between nodes, which may not be as effective in our case where Graph U-Nets are used for the temporal prediction task. However, it should be noted that in scenarios where the meshes vary across the temporal dynamics, GN may outperform LN due to its ability to consider the spatial dynamics within the graph nodes. These findings underscore the importance of selecting appropriate normalization approaches that align with the specific characteristics and requirements of the problem at hand.

\section{Conclusions and future work}
\label{sec:conclusion}

This study presented a comprehensive investigation of Graph U-Nets specialized for unsteady flow-field prediction, focusing on their mesh-agnostic spatio-temporal forecasting capabilities. By introducing key enhancements to the traditional Graph U-Net architecture, such as the GMM convolutional operator and noise injection approaches, we have established a framework for adapting these models, originally developed primarily for classification tasks, to perform effectively in regression tasks for spatio-temporal flow prediction across diverse mesh configurations and flow conditions.

The key findings of our study can be summarized as follows:
\begin{enumerate}
    \item The GMM convolutional operator is proposed to be applied instead of the traditional GCN due to its greater flexibility in modeling the flow dynamics between neighboring nodes: GMM shows 95\% improvement over GCN operators in terms of MSE.
    \item Noise injection approaches to mitigate error accumulation over long rollouts are incorporated. We found that these techniques lead to around 86\% reduction in temporal-prediction error, provided that the appropriate noise size is chosen.
    \item We investigated both the transductive and inductive performance of Graph U-Nets, demonstrating their predictive capabilities on both trained and untrained meshes. For transductive learning, we discovered that the Graph U-Net can predict the flow quantities on the unseen nodes with satisfactory accuracy even when trained on a mesh with the rear half of the domain truncated. For inductive learning, we observed that 98\% improvement can be obtained in predicting future flow fields for a specific mesh scenario by applying the inductive-learning settings compared to a model trained without the inductive settings.
    \item During inductive learning, we found that Graph U-Nets without pooling operations performed better than those with pooling. The underlying reason for this is attributed to the ability of the Graph U-Net to learn from the raw, fully detailed structure of each graph scenario, and thus adapt more flexibly to the unseen characteristics of new graphs during the inference stage.
    \item We further discovered that LN (layer normalization) outperforms GN (graph normalization) and also the case without normalization. Since LN normalizes the node features for each node independently, it can effectively capture the unique temporal dynamics at each node. Therefore, LN shows much better results than GN in our problem, where the meshes remain constant over the temporal dynamics. This finding underscores the importance of selecting appropriate normalization approaches that match the characteristics of the problem.
\end{enumerate}

The proposed Graph U-Net framework has the potential to advance the field of digital twins through its application in real-time unsteady flow-field prediction based on unstructured grids, highlighting the versatility of Graph U-Nets beyond their conventional classification applications. By accurately and efficiently predicting unsteady flow fields across various mesh configurations and generalizing to different geometries and flow conditions as verified in this study, Graph U-Nets can significantly improve the fidelity, responsiveness, and adaptability of digital twins. Although this study demonstrates the effectiveness of Graph U-Nets using a benchmark vortex-shedding problem, the methodology's potential goes far beyond this single application. As the field of graph-based deep learning continues to evolve, we expect Graph U-Nets to play an increasingly crucial role in enabling accurate and efficient predictions of complex fluid-dynamics phenomena. 

Future work will include extending the Graph U-Net architecture to be robust for much longer rollout predictions by applying various temporal numerical schemes or deep-learning architectures specialized for temporal predictions \cite{vinuesa2023easy, solera2024beta}. For example, applying the forward Euler method with Graph U-Nets that predict $\delta u$ (which is the amount of change in velocity $u$) instead of predicting $u$ directly can be an option. Additionally, the training algorithm can be modified with multi-step rollout \cite{wu2022learning}, which uses the loss function computed from the next several snapshots all at once, instead of using only the immediately next snapshot. Finally, graph-based prediction will be coupled with the uncertainty-quantification approach \cite{yang2024towards, morimoto2022assessments} and then extended to the Bayesian-design-optimization applications, where optimal aerodynamic designs can be obtained through an adaptive-sampling (also known as active learning) procedure that takes into account the predictive uncertainty of the trained AI models \cite{yang2022design,yang2023inverse}.



\section*{CRediT authorship contribution statement}
\textbf{S. Yang}: Supervision, Conceptualization, Methodology, Software, Validation, Formal analysis, Investigation, Data Curation, Writing – Original Draft, Writing – Review \& Editing, Visualization.
\textbf{R. Vinuesa}: Supervision, Writing – Review \& Editing.
\textbf{N. Kang}: Supervision, Funding acquisition.

\section*{Declaration of competing interest}
The authors declare that they have no known competing financial interests or personal relationships that could have appeared to influence the work reported in this paper.

\section*{Data availability}
Data will be made available on request.

\section*{Acknowledgments}
This work was supported by the National Research Foundation of Korea (2018R1A5A7025409), and the Ministry of Science and ICT of Korea (No. 2022-0-00969 and No. RS-2024-00355857). Also, R.V. acknowledges financial support from ERC grant no.2021-CoG-101043998, DEEPCONTROL. Views and opinions expressed are however those of the author(s) only and do not necessarily reflect those of the European Union or the European Research Council. Neither the European Union nor the granting authority can be held responsible for them.

\clearpage
\appendix

\section{Visualization of the three meshes used for inductive learning (Section \ref{sec:induct})}
\label{app:5mesh}

\begin{figure*}[htb!]
    \centering

    \begin{subfigure}[h]{0.6\textwidth}
    \centering
        \includegraphics[width=\textwidth]{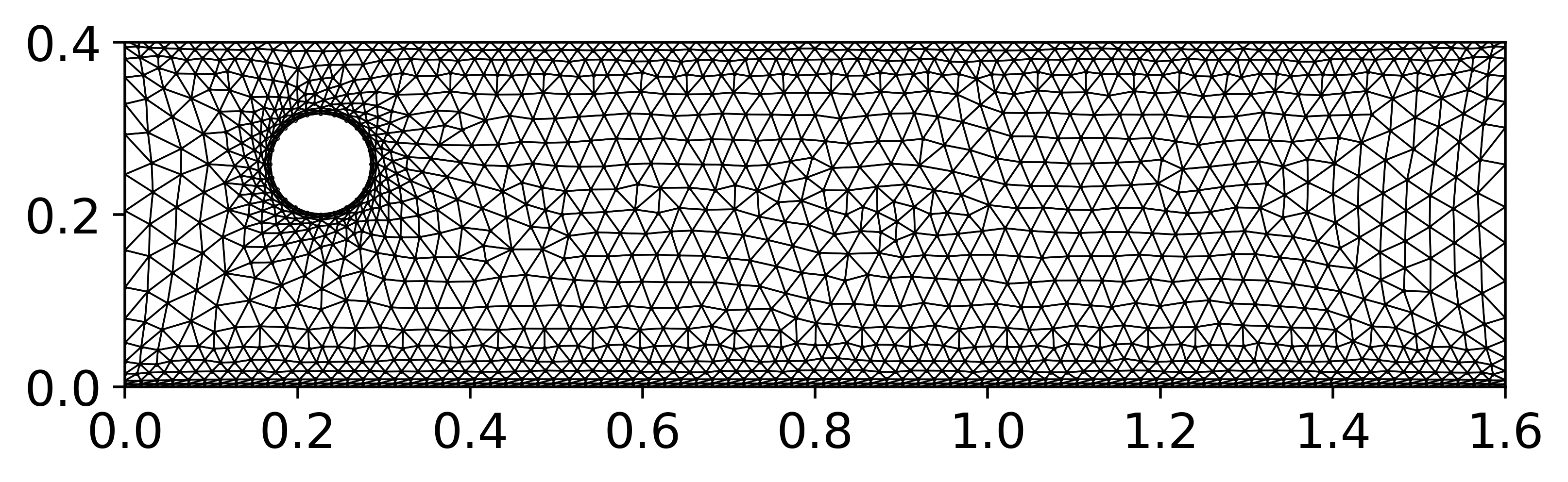}        
    \subcaption{Mesh for Induct 1 case}
    \label{fig:5meshes_1}
    \end{subfigure}

    \vfill

    \begin{subfigure}[h]{0.6\textwidth}
    \centering
        \includegraphics[width=\textwidth]{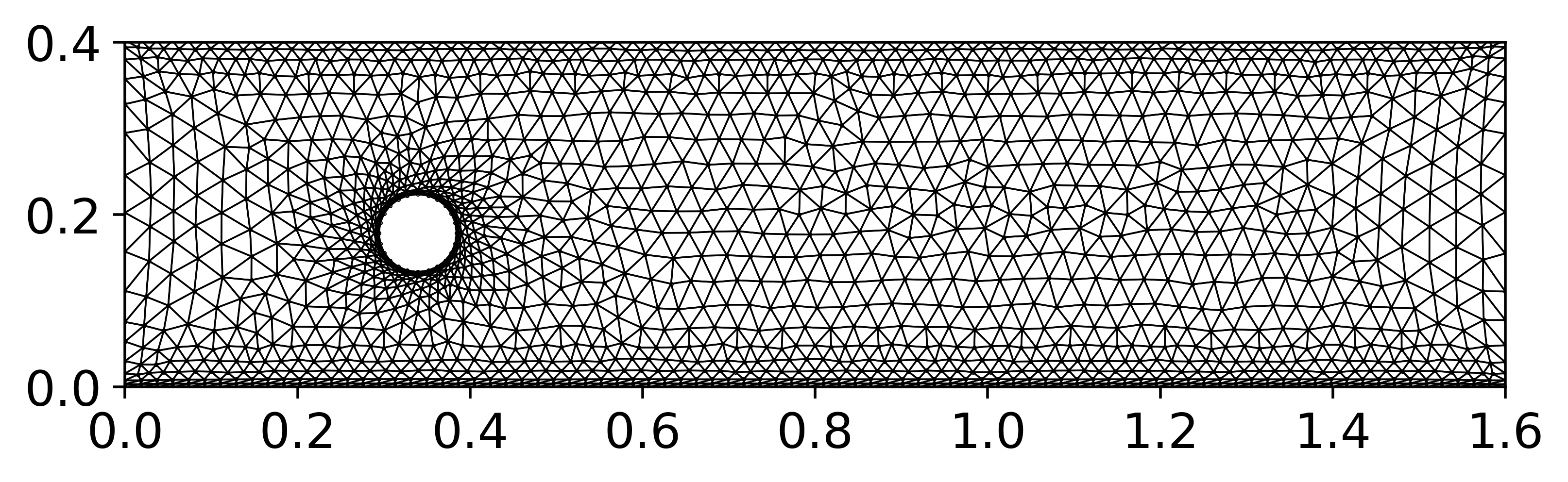}        
    \subcaption{Mesh for Induct 2 case}
    \label{fig:5meshes_2}
    \end{subfigure}

    \vfill

    \begin{subfigure}[h]{0.6\textwidth}
    \centering
        \includegraphics[width=\textwidth]{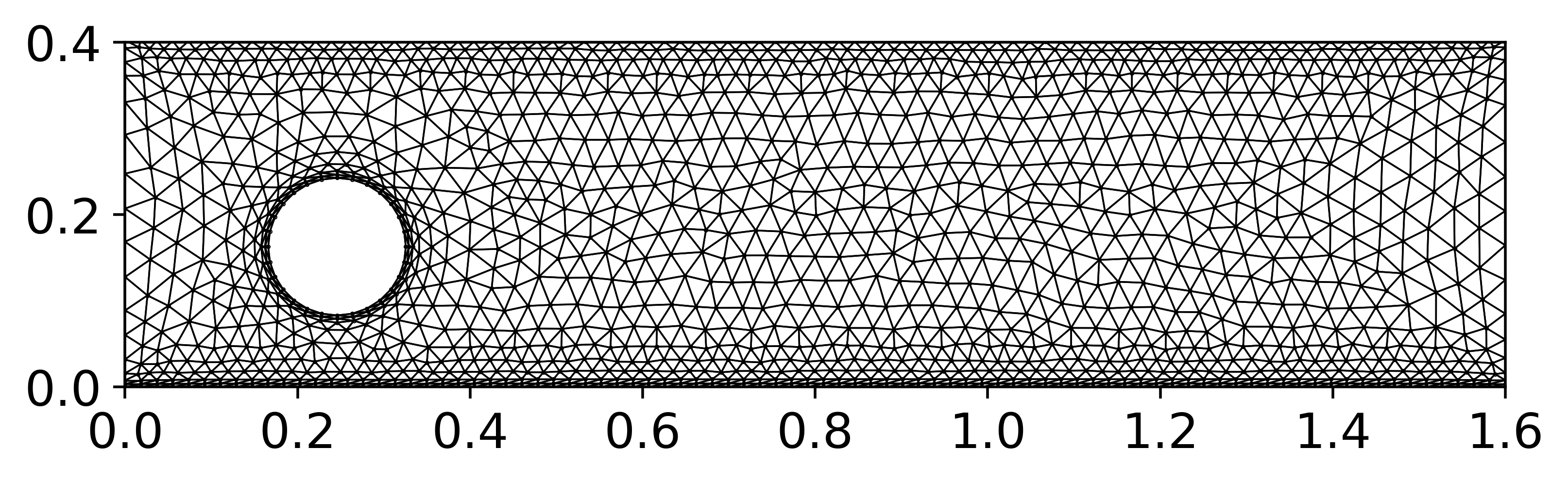}        
    \subcaption{Mesh for Induct 3 case}
    \label{fig:5meshes_3}
    \end{subfigure}

    \caption{Three meshes used for the zero-shot prediction in Section \ref{sec:zero} and inductive learning in Section \ref{sec:induct_train}.}\label{fig:5meshes}
\end{figure*} 

\clearpage
\section{Effects of the number of input snapshots and GMM kernels in inductive learning}\label{app:induct_hyper}

The experiments in this section are based on the I-noise approach without pooling in inductive-learning setting (Section \ref{sec:induct_train}). In Table \ref{tab:hyper_input}, MSE of five different Graph U-Nets with varying number of input snapshots are compared \cite{srinivasan2019predictions}. The results demonstrate that for all graph types (baseline, Induct 1$\sim$3), an input with 40 snapshots shows the best performance. For this reason, the number of input snapshots is set to 40 in Section \ref{sec:induct_train}. In Table \ref{tab:hyper_kernel}, MSE of different Graph U-Nets with different number of GMM kernels are compared. Although the GMM kernel with 1 was sufficient for the previous experiments before the inductive learning (as can be found in Table \ref{tab:GCN}), it is discovered that for the inductive-learning setting, more GMM kernels are required to ensure sufficient model capacity. In fact, the GMM with three kernels shows a dramatic improvement in the Induct 3 scenario compared with the single-kernel case (from $68.71\times10^{-3}$ to $1.78\times10^{-3}$).

\begin{table}[htb!]
\centering
\begin{threeparttable}
\begin{tabular*}{0.55\columnwidth}{@{\extracolsep{\fill}}c|cccc}
\cline{1-5} 
\multirow{2}{*}{\shortstack{Number of\\input snapshots}} & \multicolumn{4}{c}{Scenario type} \\ \cline{2-5} 
 & Baseline & Induct 1 & Induct 2 & Induct 3 \\ \cline{1-5} 
1 & 78.28 & 97.72 & 80.86 & 109.78 \\
5 & 123.83 & 146.40 & 117.56 & 163.05 \\
10 & 29.40 & 59.51 & 50.67 & 63.89 \\
20 & 70.19 & 86.88 & 80.45 & 108.19 \\
40 & \textbf{4.97} & \textbf{6.32} & \textbf{6.30} & \textbf{1.78}  \\ \cline{1-5}
\end{tabular*}

\caption{MSE ($\times10^{-3}$) over 200 rollout predictions in inductive learning with varying number of input snapshots: both baseline and Induct 3 meshes are trained for this inductive-learning setting.}
\label{tab:hyper_input}

\end{threeparttable}
\end{table}

\begin{table}[htb!]
\centering
\begin{threeparttable}
\begin{tabular*}{0.5\columnwidth}{@{\extracolsep{\fill}}c|cccc}
\cline{1-5} 
\multirow{2}{*}{\shortstack{Number of\\GMM kernels}} & \multicolumn{4}{c}{Scenario type} \\ \cline{2-5} 
 & Baseline & Induct 1 & Induct 2 & Induct 3 \\ \cline{1-5} 
1 & \textbf{3.15} & 7.00 & 6.47 & 68.71 \\
3 & 4.97 & \textbf{6.32} & \textbf{6.30} & \textbf{1.78} \\ \cline{1-5}
\end{tabular*}

\caption{MSE ($\times10^{-3}$) over 200 rollout predictions in inductive learning with two different numbers of GMM kernels: both baseline and Induct 3 meshes are trained for this inductive-learning setting.}
\label{tab:hyper_kernel}

\end{threeparttable}
\end{table}
\clearpage
\bibliographystyle{elsarticle-num-names} 
\bibliography{cas-refs}





\end{document}